\documentclass[runningheads]{llncs}

\usepackage{eccv}

\usepackage{eccvabbrv}

\usepackage{graphicx}
\usepackage{booktabs}

\usepackage[accsupp]{axessibility}  

\usepackage[normalem]{ulem}

\usepackage{hyperref}

\usepackage{orcidlink}

\begin{document}

\title{On the Potential of Open-Vocabulary Models for Object Detection in Unusual Street Scenes} 

\titlerunning{Object Detection in Unusual Street Scenes}

\author{Sadia Ilyas\inst{1,2} \and
Ido Freeman\inst{1}\orcidlink{0000-0003-4825-060X} \and
Matthias Rottmann\inst{2}\orcidlink{0000-0003-3840-0184}}

\authorrunning{S.~Ilyas et al.}

\institute{Aptiv Services Deutschland GmbH, Wuppertal, Germany \and
University of Wuppertal, Germany\\
\email{sadia.ilyas@aptiv.com}
}

\maketitle

\begin{abstract}

Out-of-distribution (OOD) object detection is a critical task focused on detecting objects that originate from a data distribution different from that of the training data. 
In this study, we investigate to what extent state-of-the-art open-vocabulary object detectors can detect unusual objects in street scenes, which are considered as OOD or rare scenarios with respect to common street scene datasets. Specifically, we evaluate their performance on the OoDIS Benchmark, which extends RoadAnomaly21 and RoadObstacle21 from SegmentMeIfYouCan, as well as LostAndFound, which was recently extended to object level annotations. 
The objective of our study is to uncover short-comings of contemporary object detectors in challenging real-world, and particularly in open-world scenarios. 
Our experiments reveal that open vocabulary models are promising for OOD object detection scenarios, however far from perfect. Substantial improvements are required before they can be reliably deployed in real-world applications.
We benchmark four state-of-the-art open-vocabulary object detection models on three different datasets. Noteworthily, Grounding DINO achieves the best results on RoadObstacle21 and LostAndFound in our study with an AP\textsubscript{50..95} of 48.3\% and 25.4\% respectively. YOLO-World excels on RoadAnomaly21 with an AP\textsubscript{50..95} of 21.2\%. 
\end{abstract}

\section{Introduction}
\label{sec:intro}

In safety-critical applications such as autonomous driving or medical diagnosis, it is crucial for an AI-based perception system to recognize that it may be ill-equipped to make accurate predictions. Knowing when it does not know allows the AI to flag such instances and avoid making unreliable decisions perhaps by opting to  defer control to a human operator or by taking conservative actions to mitigate the risk. 
Incorrect responses to OOD data can lead to potentially unfortunate outcomes. Therefore, addressing the challenges posed by OOD scenarios and proposing methods for OOD detection is vital for robust and trust-worthy AI systems \cite{li2023trustworthy}. 

\begin{figure}[tb]
\centering
  \begin{subfigure}{0.38\textwidth}
    \includegraphics[width=\linewidth]{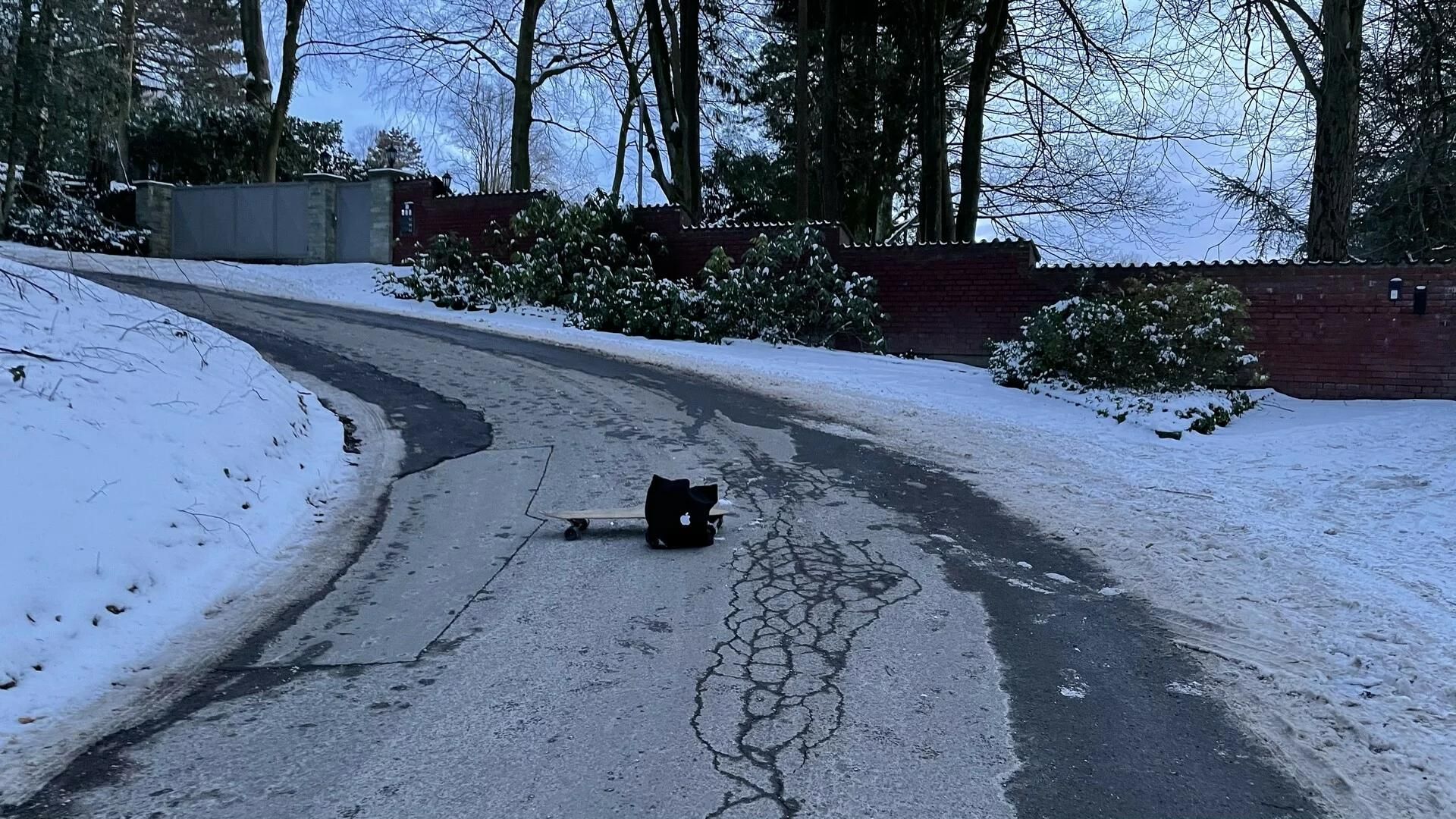}
  \end{subfigure}
  \begin{subfigure}{0.38\textwidth}
    \includegraphics[width=\linewidth]{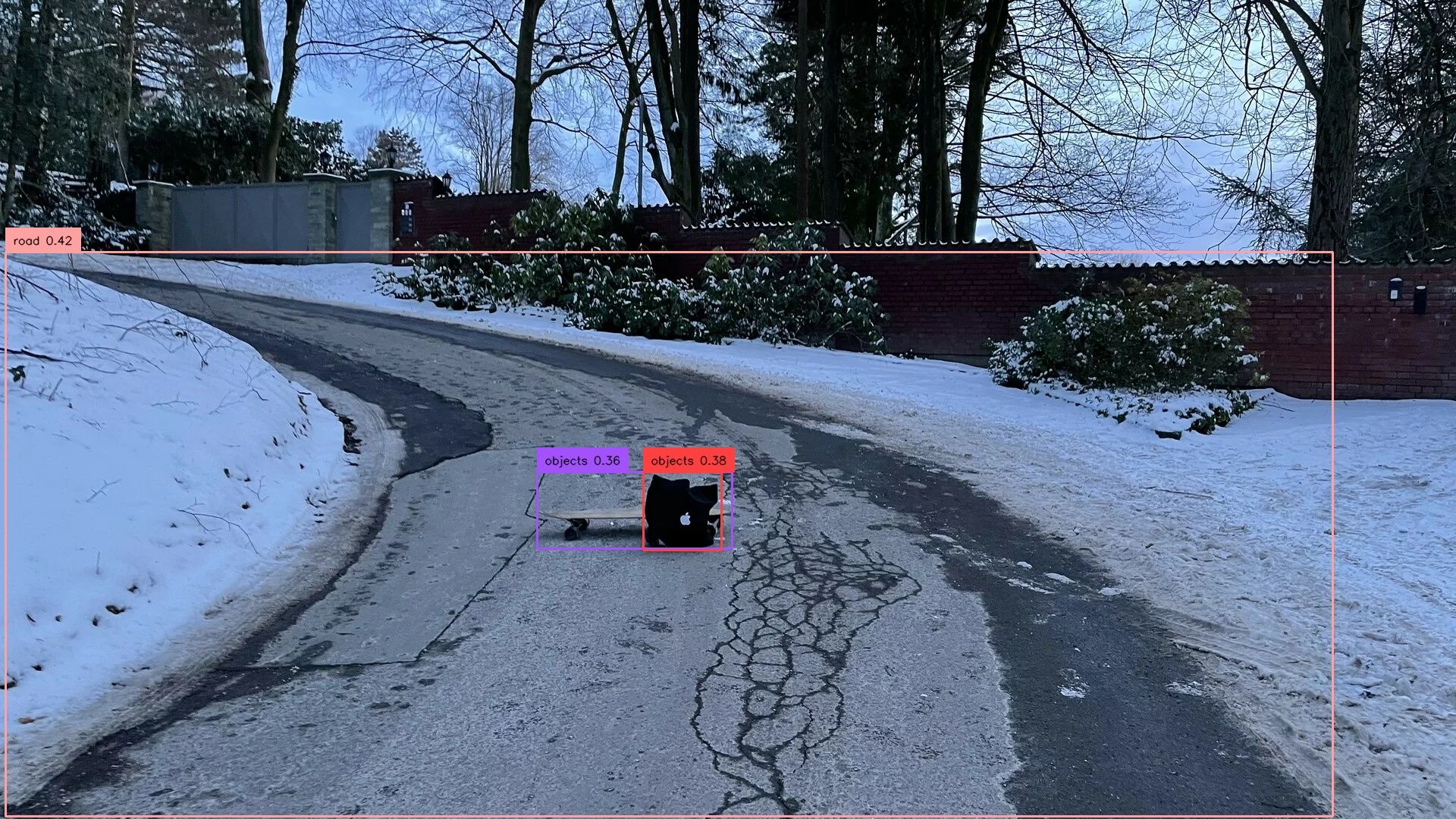}
  \end{subfigure}
\caption{Comparison of OOD object detection using two different models. The left image shows YOLOv8 incorrectly predicting everything as background and missing OOD objects, while the right image shows Grounding DINO detecting both road objects.}
\label{fig:figure1}
\end{figure}

Deep neural networks demonstrate exceptional performance when tested on data closely aligned with their training data distribution. However, they face a significant challenge in real-world when encountering unusual circumstances as shown in Figure \ref{fig:figure1}. 
In our study we observe that in such situations the network tends to make either over-confident predictions or elude the entire recognition of any unusual objects.

Previous methods aim to enhance the model's ability to discern between known (in-distribution, ID) and unknown (out-of-distribution, OOD) objects, with the central objective being to distinguish between ID and OOD objects. 
Notable strategies include employing methods originating from image classification \cite{liang2017enhancing}, Bayesian model uncertainty techniques \cite{mukhoti2018evaluating} and Post-hoc approaches \cite{wilson2023safe}.
These existing methods focus on correct classification of objects into ID and OOD. However, the challenge that has been often overlooked is addressing the objects that elude the recognition entirely. As one aspect, our study addresses this more complex aspect of OOD detection, emphasizing the importance of detecting all potential unusual objects and ensuring they are neither disregarded by the model nor classified as background.

Segmentation methods \cite{cheng2022masked, nekrasov2023ugains, gasperini2023segmenting} have been widely employed for anomaly detection, where the goal is to detect unusual objects on pixel level. 
While semantic segmentation has made substantial progress in anomaly detection (also under domain shift), the exploration of methods based on object detection remains relatively underdeveloped. 

With the advent of open-vocabulary foundation models like Grounding DINO et al.~\cite{bommasani2021opportunities} that have been trained on potentially everything, it is a legitimate question whether the problem of OOD object detection has been explained away by that amount of training data.
As of now, OOD object detection capabilities of foundation models remains largely unexplored. This is also due to their training data being large, diverse and hard to analyze. It can be expected that such models will be fine-tuned for many applications and, hence, their capabilities in unusual open-world scenarios is of clear interest.

\begin{figure}[tb]
\centering
  \begin{subfigure}{0.38\textwidth}
    \includegraphics[width=\linewidth]{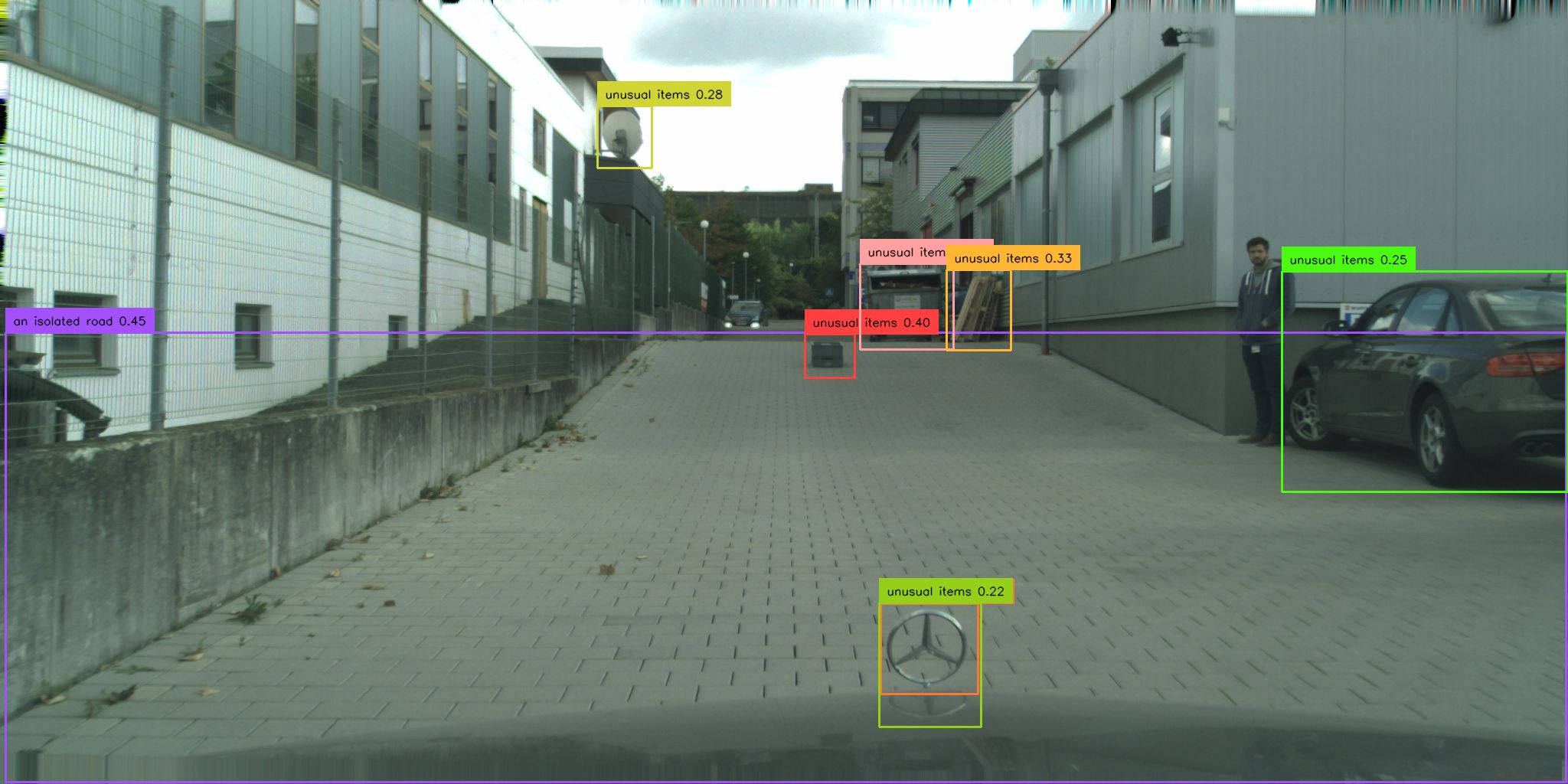}
      \caption{Prompt not fully understood}
    \label{fig:a}
  \end{subfigure}
  \begin{subfigure}{0.38\textwidth}
    \includegraphics[width=\linewidth]{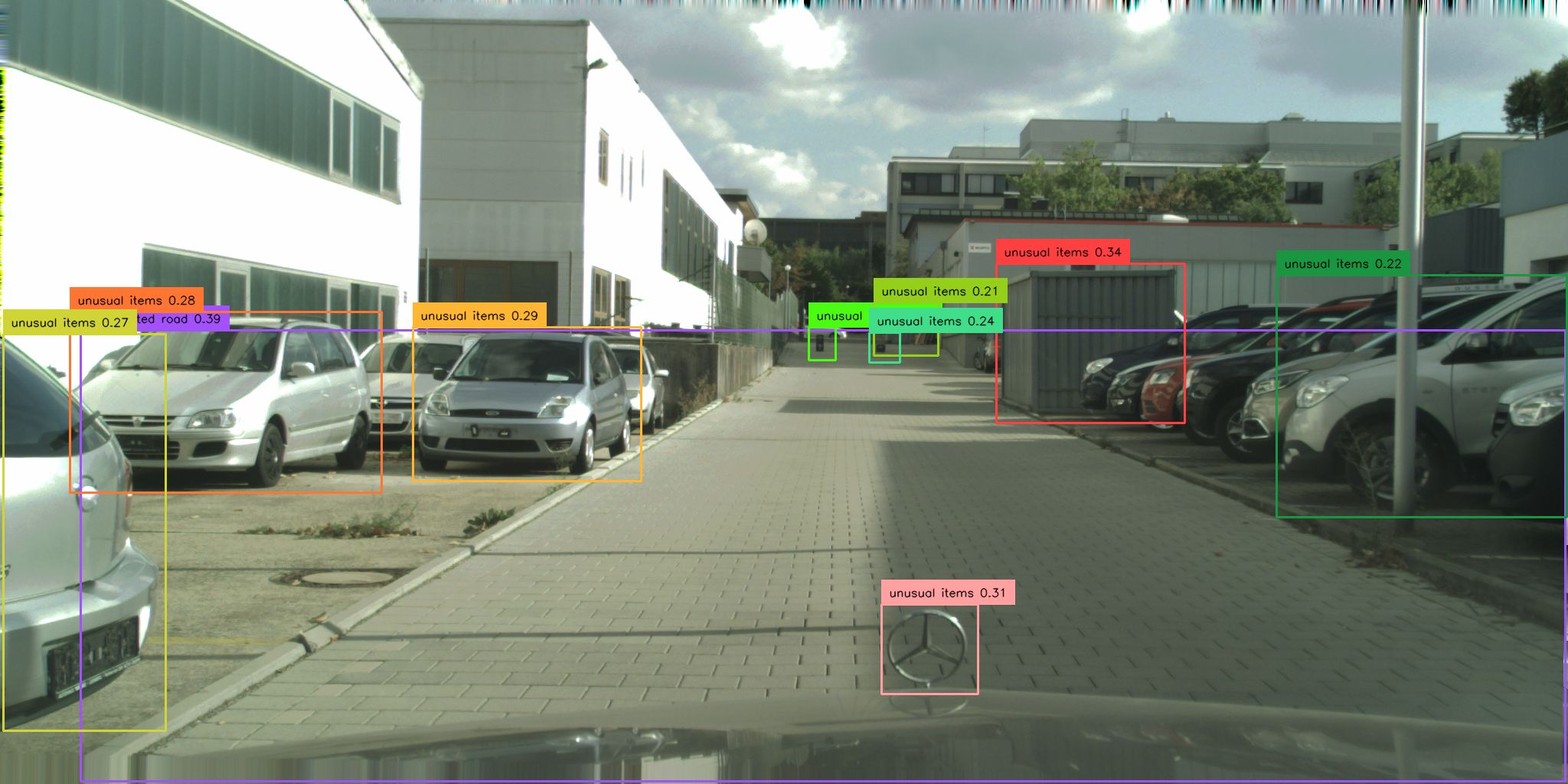}
      \caption{Cars predicted as 'unusual items'}
    \label{fig:b}
  \end{subfigure}
    \begin{subfigure}{0.38\textwidth}
    \includegraphics[width=\linewidth]{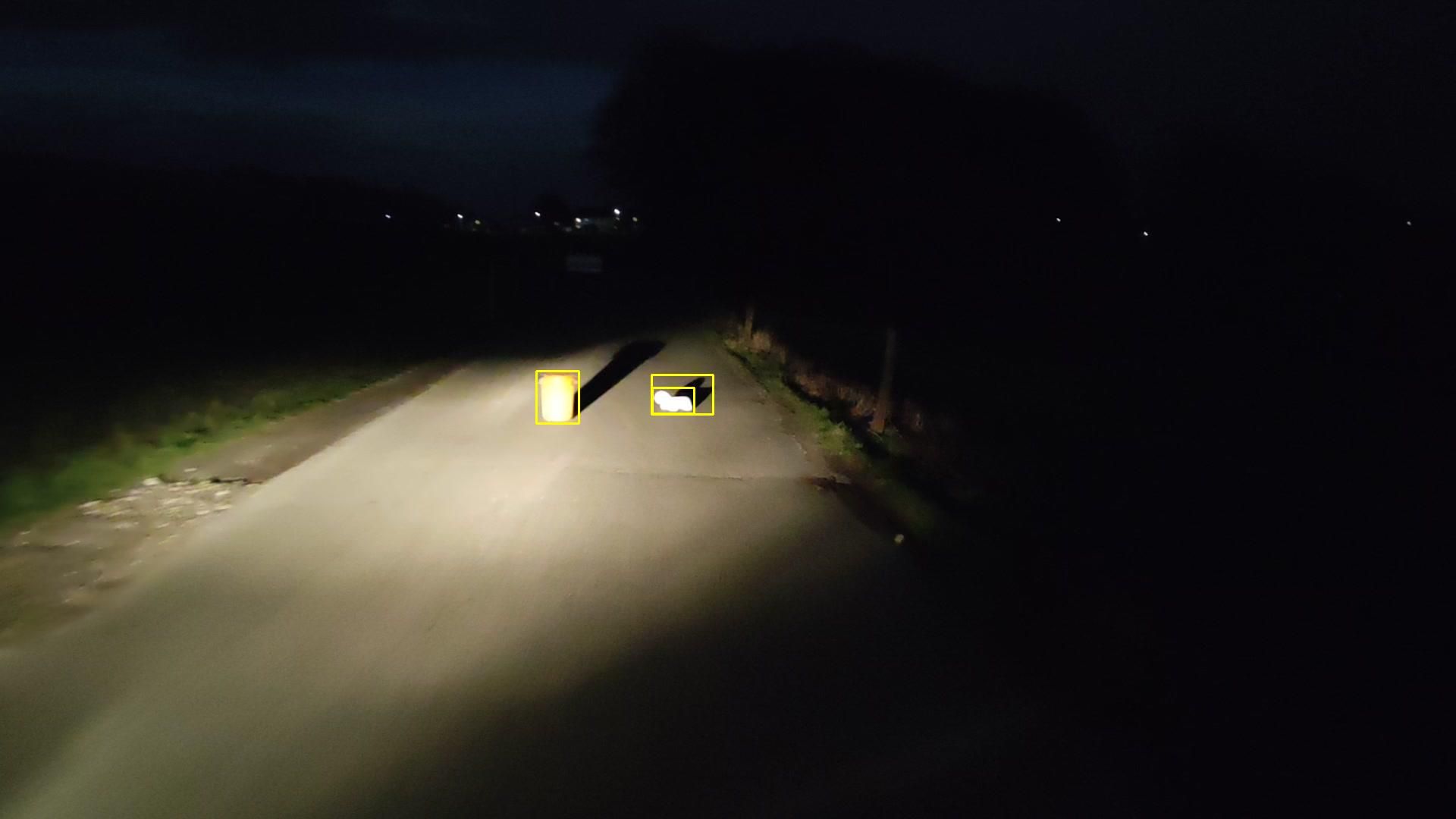}
    \caption{Challenging lighting condition}
    \label{fig:c}
  \end{subfigure}
  \begin{subfigure}{0.38\textwidth}
    \includegraphics[width=\linewidth]{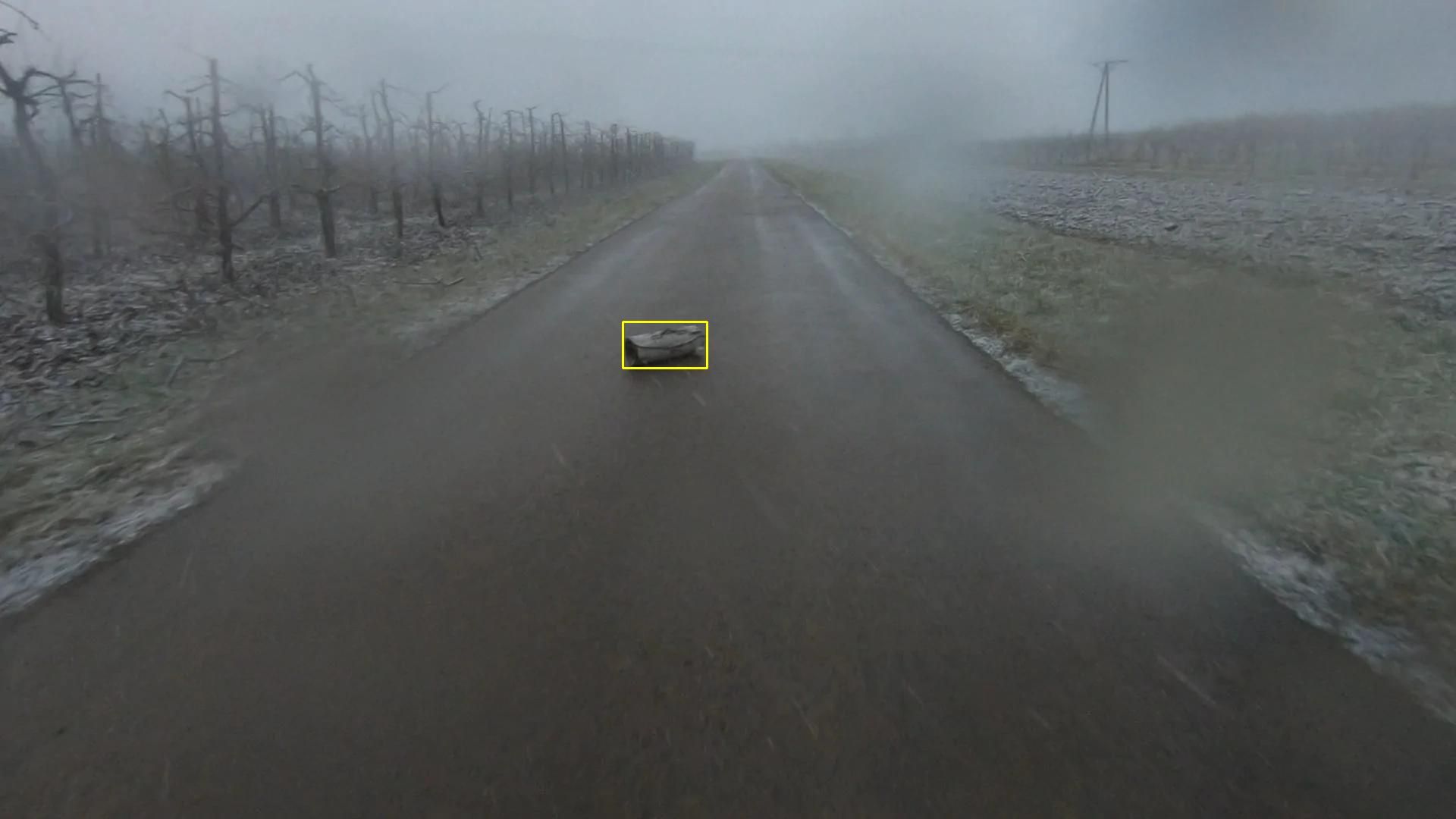}
    \caption{Severe weather condition}
    \label{fig:d}
  \end{subfigure}%
\caption{Panels (a) and (b) are prompted with text 'unusual items found on an isolated road'. In figure (a) the model fails to understand the prompt completely, thus predicts objects also outside the road region. In (b) the cars are mistaken to be 'unusual items' by the model in the street scene. In (c) and (d), the model preforms well by accurately detecting the OOD objects given the severe conditions.}
\label{fig:figure2}
\end{figure}

In this study, we extend the scope of object detection to investigate the challenges associated with OOD object detection using current foundation models.
We benchmark state-of-the-art open-vocabulary object detection models, including Grounding DINO \cite{liu2023grounding}, YOLO-World \cite{cheng2024yolo}, Omdet \cite{zhao2024real} and MDETR \cite{kamath2021mdetr}, for OOD object detection on three challenging anomaly detection datasets. Two of these datasets, RoadObstacle21 and RoadAnomaly21, are presented in OoDIS \cite{nekrasov2024oodis}. We also conduct extensive experiments on the publicly available LostAndFound \cite{pinggeralost} dataset, as well as Fishyscapes LostAndFound \cite{blum2019fishyscapes} which is a filtered subset of LostAndFound with extended annotations, included in OoDIS as Fishyscapes L\&F. We test the models by using prompt engineering and provide the models with generic text prompts for the complete dataset instead of explicit prompts referring to the objects present in each image. This approach tests the model's ability to detect unusual objects present in street scenes.
Our experiments reveal that these models are highly prompt dependent, often misunderstanding text prompts. They often lag in performance when confronted with smaller sized objects positioned far from the camera. However, in some cases they perform exceptional in detecting all objects even under severe weather conditions, as shown in Figure \ref{fig:figure2}.

Our contributions are summarized as follows:
\begin{itemize}
    \item We extensively study open-vocabulary models for OOD object detection in the context of street scenes, introducing prompt-based OOD object detection baselines.
    \item We identify that these models often lack confidence and struggle with accurate predictions, in particular in dense scenarios, revealing the limitations of current state-of-the-art models.
    \item  We identify suboptimal performance with multiple prompts and introduce a prompt ensemble for Grounding DINO that performs superior compared to multiple prompts used together in one model.
\end{itemize}

\section{Related Work}
This section gives a brief overview of previous methods pertaining to classification, segmentation and object detection in the context of OOD object detection.

\paragraph{\textbf{OOD detection in classification tasks:}}
One of the simplest tasks in which OOD detection plays a role is classification. Related methods mostly rely on analyzing the distribution of prediction scores to make decisions about the ID and OOD samples.
Classical methods include Maximum softmax probability \cite{hendrycks2016baseline}, Ensembling \cite{lakshminarayanan2017simple}, Energy-score method \cite{liu2020energy}, Mahalanobis distance \cite{lee2018simple}, GradNorm score \cite{huang2021importance}. Additionally, meta classifiers \cite{chan2021entropy} are introduced for entropy-based OOD object detection. Outlier exposure \cite{hendrycks2019oe} improves OOD detection by training the detection model on diverse sets of outlier data. UQGAN \cite{oberdiek2022uqgan} augments each class with out-of-class examples generated by a conditional GAN and uses this with a one-vs-all classifier to distinguish between ID and OOD.
With the development of internet scale pretrained models, there is an increasing interest in applying them to smaller tasks, thus to OOD detection. A zero-shot OOD detection method \cite{esmaeilpour2022zero} based on pre-trained CLIP \cite{radford2021learning} generates potential unknown class names for each test sample and calculates a confidence score using both known and candidate unknown class names.
CLIPN \cite{wang2023clipn} built on top of CLIP \cite{radford2021learning} equips it to distinguish between ID and OOD by using positive and negative semantic prompts. CLIP-OS \cite{sun2024clip} enhances ID/OOD separability by synthesizing reliable OOD samples from ID-relevant features and applying unknown-aware prompt learning. DOG \cite{jiang2024dog} generates synthetic outliers from ID data using a large scale pre-trained diffusion model, to train detection model for OOD detection.

\paragraph{\textbf{OOD detection in semantic segmentation:}}
Segmentation based methods for OOD detection have gained quite importance due to their ability to provide fine-grained analysis and localization of anomalous objects.
This motivated the development of benchmarks such as Fishyscapes \cite{blum2019fishyscapes} and SegmentMeIfYouCan \cite{chan2021segmentmeifyoucan} for anomaly detection based on semantic segmentation. Transformer based anomaly segmentation method, Mask2Anomaly \cite{rai2023unmasking}, which uses global mask attention to enhance its attention mechanisms, achieves state of the art on those benchmarks. 
 Integrating OOD detection with state-of-the-art semantic segmentation \cite{chan2021entropy} involves re-training with softmax entropy thresholding. Recent work includes generative models that resynthesize images for pixel-wise anomaly detection by comparing the original and resynthesized images \cite{lis2019detecting}. Combining uncertainty with these methods helps guide a network to detect anomalies at the pixel level from semantic maps \cite{di2021pixel}.

Existing methods often lack generalization to scenarios with significant domain shifts. In response, recent studies \cite{li2024anomaly, gu2024anomalygpt, elhafsi2023semantic, xu2024customizing, du2024uncovering} are focusing on customizing vision-language foundation models into anomaly detectors utilizing a multi modal prompting strategy \cite{xu2024customizing}. 
Large language models (LLMs) have recently also been explored for their potential to detect tabular anomalies without additional model training \cite{li2024anomaly}. AnomalyGPT \cite{gu2024anomalygpt}, based on LLM can detect anomalies in an image and provide their relative positions. Another study demonstrates that LLM based monitors can effectively identify semantic anomalies in vision-based policies for autonomous driving \cite{elhafsi2023semantic}. Additionally, LLMs have been used for Video Anomaly Understanding, to automatically detect unusual occurrences in videos \cite{du2024uncovering}. Unlike traditional object detection, these methods employ a question-answering approach using LLMs.

\paragraph{\textbf{OOD detection in object detection:}}

Object detection depends on spatial context to be able to localize the objects and further classify them. That makes it more challenging as compared to segmentation which operates on per pixel level. OOD object detection gets further complicated due to the scarcity of diverse datasets including wide range of OOD scenarios. Moreover, the absence of standardized benchmarks and evaluation protocols adds up on the challenges in the direction of object detection for OOD detection. However this recently has been changed due to the availability of OoDIS \cite{nekrasov2024oodis}. 

On the methodological side, VOS \cite{du2022vos} is one of the first works on object level for OOD detection. It is a framework that dynamically generates outliers during training by sampling virtual outliers from the low-likelihood region of the class-conditional Gaussian distributions. These virtual outliers, together with the ID objects are used to compute an uncertainty loss, which serves as a regularization term during training. The uncertainty estimation is trained together with the object detection loss, to help optimization of both aspects together. Furthermore the decision boundary is regularized via an energy score where OOD objects have a positive energy score.
SAFE \cite{wilson2023safe} utilizes sensitive layers from the backbone of an object detector that are effective at detecting OOD objects, particularly residual convolutional layers followed by batch normalization. These specific layers are further used to train an MLP to distinguish between ID samples and OOD samples.
While these methods are effective for OOD object detection within known categories, they might struggle to generalize well to entirely new classes or unexpected data distribution that was not encountered during the training. 
In our proposed study we build upon the idea of object detection by employing open-vocabulary detectors and explore the limitations of current foundation models in terms of OOD object detection. Current methods for OOD object detection fall behind the advent of foundation models. To the best of our knowledge these models are not yet explored for OOD object detection. Given the challenges associated with OOD object detection, we seek to uncover the limitations of existing foundation models concerning OOD object detection.

\section{Models and Evaluation Metrics}
Our emphasis in this study is primarily on investigating the open-vocabulary models in unusual scenarios.  We provide necessary information about the models utilized in Table \ref{tab:method}.
While datasets like Cityscapes \cite{Cordts2016Cityscapes} represent common street scenes with real world scenarios, in contrast, datasets from OoDIS, SegmentMeIfYouCan and LostAndFound are specifically designed to highlight rare and unusual situations.
In our experimental setup we focus on datasets namely those presented in OoDIS \cite{nekrasov2024oodis} benchmark and LostAndFound \cite{pinggeralost}, that cover a wide variety of unusual aspects in street scenes. The complementarity of these datasets help us obtain comprehensive evaluation of the open-vocabulary models and helps us determine how effectively these models handle rare and unusual street scenes.

\subsection{Object Detectors}
In this section we briefly summarize the models that we evaluate.

\begin{table}[tb]
\centering
\scriptsize
  \caption{Information about the models belonging to different categories i.e., open-vocabulary, object detection and OOD object detection. It also specifies the backbone and pre-trained data used for these models in our experiments.}
  \begin{tabular}{@{}llll@{}}
    \toprule
    \textbf{Category} & \textbf{Model} & \textbf{Backbone} & \textbf{Pre-trained Data}\\
    \midrule
    Open-voc & MDETR & EfficientnetB5 & VG,COCO,Flickr30k\\
    & YOLO-World-L & YOLOv8-L & O365,GoldG\\
    & OmDet-Turbo & Swin-T & O365,GoldG\\
    & Grounding-DINO-T & Swin-T & O365,GoldG,Cap4M\\
    \midrule
    OD  & YOLOv8 & YOLOv7 & COCO  \\
    \midrule
    OOD  & VOS & ResNet50 & BDD-100k  \\
    \bottomrule
  \end{tabular}
    \label{tab:method}
\end{table}
\textbf{Grounding DINO} \cite{liu2023grounding} is an innovative open-vocabulary object detector that combines a Transformer-based detector DINO \cite{zhang2022dino} with grounded pre-training. It enables the detection of various objects based on human inputs (so-called prompts) such as category names or referring expressions. The detector operates in three phases: feature enhancement, language-guided query selection, and cross-modality decoding. 
The feature enhancer consists of multiple layers, incorporating deformable self-attention for image features and vanilla self-attention for text features, along with image-to-text and text-to-image cross-attention modules. It extracts multiscale features from the image using Swin Transformer \cite{liu2021swin} and from the text using BERT \cite{devlin2018bert}. These features are then fed into a feature enhancer for cross-modality fusion.
The language-guided query selection module selects features relevant to the input text as decoder queries. The cross-modality decoder merges features from both image and text modalities, employing layers for self-attention, image cross-attention, text cross-attention, and feed-forward processing.

\textbf{YOLO-World} \cite{cheng2024yolo} is an open-set model that enhances the widely recognized YOLO object detection model with open-vocabulary detection capabilities through vision-language modeling. 
It is based on a new re-parameterizable Vision-Language Path Aggregation Network (RepVL-PAN) and region-text contrastive loss to facilitate interaction between visual and linguistic information. The architecture begins with the Text Encoder pre-trained by CLIP \cite{radford2021learning}, which converts the input text into embeddings. The Image Encoder processes the input image, generating multi-scale image features. The RepVL-PAN is then used to fuse image and text features at multiple levels.

\textbf{MDETR} \cite{kamath2021mdetr} is an end-to-end text-modulated detection method derived from DETR \cite{carion2020end}. It utilizes a convolutional backbone for visual feature extraction and a language model, RoBERTa \cite{liu2019roberta}, for text feature extraction. These modalities' features are projected into a shared embedding space, combined and passed into a transformer encoder-decoder.

\textbf{OmDet-Turbo} \cite{zhao2024real} introduces an Efficient Fusion Head (EFH) module, aiming for real-time performance. The overall OmDet-Tubro model features a text backbone, an image backbone and an EFH module. The text backbone, i.e., CLIP, encodes prompts and labels, generating embeddings and token-level outputs for prompts. The image backbone extracts multiscale features from the input image. EFH includes an Efficient Language-Aware Encoder (ELA-Encoder) and Decoder (ELA-Decoder). ELA-Encoder selects top-K initial queries, fused with prompt embedding for language-guided multi-modality queries in ELA-Decoder. 

\textbf{GD Prompt Ensemble} we used an ensemble approach with Grounding DINO by using multiple prompts to individually query the model. Specifically, we generated separate predictions for each prompt and then combined these predictions through a weighted average of the confidence scores. Each prompt contributed differently to the final prediction, and the weights assigned to each prompt's predictions were determined based on their performance. After combining the predictions with weighted averaging, we applied Non-Maximum Suppression (NMS) to the fused bounding boxes to eliminate redundant detections.

We use the above mentioned open-vocabulary models for OOD object by providing them with text prompts and explore to what extent they could be utilized for OOD object detection. We also use a classical closed-vocabulary object detection model, Yolov8 \cite{Jocher_Ultralytics_YOLO_2023} to get an overview of how much does it lag as compared to an open-vocabulary model in terms of OOD object detection.

\subsection{Metrics}
We used  the well-known object detection metrics, Average Precision (AP) and Average Recall (AR), as these metrics are present in the OoDIS benchmark. Additionally, true positives (TP), false positives (FP), and false negatives (FN) are also reported to gain insight into the specific types of errors made by the models experimented.
To assess the performance on the LostAndFound dataset, we reported additional metrics beyond those provided in the OoDIS benchmark. Since the ground truth for LostAndFound is publicly available, we conducted more experiments with lower threshold. Predictions were considered true positives (TP) if the Intersection over Union (IoU) score was above 0.1. We selected this low IoU threshold because, although objects were often detected, the overlap between the ground truth annotation and the predicted bounding box did not consistently exceed 0.5.
To avoid disregarding true positives, we kept the IoU threshold below 0.5. Before counting the TP and FP, we applied Non-Maximum Suppression (NMS) for each prompt separately to eliminate redundant predictions. 
We reported additional AP scores across three custom IoU threshold intervals using COCO evaluation, i.e., AP\textsubscript{10}, AP\textsubscript{10..75}, AP\textsubscript{20..75} with a step size of 0.5.

\section{Experiments}
In this section, we briefly describe the datasets and subsequently, we discuss in detail the experiments performed on those datasets using open-vocabulary models.

\subsection{Dataset Description:}

In our experiments we choose datasets that are located in the long tail of street scenes. Thus they can be viewed as OOD with respect to the datasets that present frequent events in street scenes such as Cityscapes \cite{Cordts2016Cityscapes} and or BDD100K \cite{yu2020bdd100k}. These datasets can help us evaluate the generalization ability of open-vocabulary models in handling OOD situations.

\textbf{RoadObstacle21} \cite{nekrasov2024oodis, chan2021segmentmeifyoucan} consists of 447 images capturing various road scenes with different road surfaces, lighting conditions including daytime and nighttime scenes, and weather conditions such as sunny and snowstorm. 
The OOD objects are characterized by their presence at different distances from the camera and varying sizes relative to the scene. The road surfaces include instances of asphalt, concrete, gravel and bricks, reflecting the diversity of road surfaces encountered in real-world environments. The OOD objects present in this dataset include elements like stuffed toys, bottles, sleighs, tree stumps etc., thus elements that are typically not present in a road scene.

\textbf{RoadAnomaly21} \cite{nekrasov2024oodis, chan2021segmentmeifyoucan} comprises 100 images featuring very large objects in various environments. Unlike typical road scene datasets, this dataset is slightly different and focuses on capturing scenes with exceptionally large objects rather than road-specific scenarios. The dataset is curated for the task of large object anomaly detection. 
Notably, the anomalies in this dataset predominantly consist of animals. This dataset does not specify a particular region of interest, instead, anomalies can occur anywhere in the image.

\textbf{LostAndFound} \cite{pinggeralost} comprises images captured at five distinct locations. surface textures, and pronounced changes in illumination conditions. This dataset is curated for the task of anomaly detection, focusing on identifying smaller-sized hazardous objects placed at different distances. Each location represents a different setting or environment where lost and found items may occur. The test set of LostAndFound dataset is divided into five subsets, with each subset containing a varying number of images captured at the respective locations with: 371, 248, 304, 137 and 143 images for each location respectively. Each particular location encompass unique challenges, including irregular road profiles, objects placed at considerable distances and varying road.

\textbf{Fishyscapes L\&F} \cite{nekrasov2024oodis} The Fishyscapes L\&F dataset is a specialized subset derived from the original LostAndFound dataset. This subset consists of 275 images, selected from both the training and test sets of the LostAndFound dataset. However, the specific images included in the Fishyscapes L\&F subset have not been disclosed publicly.

\subsection{Quantitative Experiments}

To comprehensively assess the performance of open-vocabulary models when confronted with unusual objects in street scenes, we conducted a series of experiments utilizing the OoDIS benchmark and the LostAndFound dataset. These datasets encapsulate diverse characteristics reflecting typical road scenes that may occur as rare scenarios.

\paragraph{\textbf{Evaluation protocol:}} 
Our primary objective remains to ascertain the detection of unusual objects, irrespective of the precision of bounding box localization relative to ground truth annotations. 
Notably, there may be instances where objects are detected but not precisely localized, potentially leading to diminished IoU score. To account for these detections, we used a lower IoU threshold in LostAndFound experiments.
However, this adjustment was not applicable to the other two datasets in the benchmark.

To focus on objects on the road in LostAndFound, we initially filtered out predictions outside the designated region of interest, which excluded edge objects. To address this, we computed the intersection of predicted bounding boxes with the region of interest and included boxes that exceeded a predefined threshold. Although this adjustment included more edge objects, it also increased false positives, as many side objects were not marked as anomalies in the ground truth annotations.
For RoadObstacle21, we used the model's predicted road region by prompting it, to filter predictions. RoadAnomaly21 predictions were not filtered and were evaluated without a specific region of interest.

\paragraph{\textbf{Prompt Engineering:}} In our experimental setup, we provided the previously mentioned models with broad and generic textual inputs to assess their capacity for textual comprehension. We deliberately avoided employing highly specific terminologies corresponding to the objects depicted in the images. In the RoadObstacle21 and the LostAndFound dataset we nowhere explicitly named any of the unusual objects present on the road surface. 
The prompts had to be fairly different for each separate dataset to optimize performance.

Interestingly, we noted a trend where shorter text prompts yielded superior results in some instances. However, for certain subcategories, such as in the LostAndFound dataset, subset 4 in Table \ref{tab:result2}, we observed poorer results with shorter text prompt \textit{'objects'} and superior results with the prompt \textit{'unusual items unexpectedly found along an isolated road'}.

During our experiments we noticed that combining multiple textual prompts often led to a reduction in the number of true positives (TP) and AP. For example in Table \ref{tab:GD1}, Grounding DINO achieves better results when prompted separately using two different prompts. However, this was not the case when they were combined and prompted together as a single prompt, indicating potential challenges in model comprehension when confronted with diverse textual inputs.  
As shown in Table \ref{tab:GD1}, we also tested an ensemble approach with Grounding DINO by prompting it separately with two different prompts and then combining the results using a weighted average of the predictions. 
The ensemble results in Table \ref{tab:GD1} show the impact of two different weight distributions i.e., 0.6--0.4 and 0.8--0.2\, respectively. This method is only slightly sensitive w.r.t.\ the weight distributions, however, this approach shows better performance as compared to prompting the model with several prompts together, which leads to suboptimal results.

Additionally, we explored the impact of adjusting the objectness score thresholds on the interpretation of results. For the sake of brevity, we maintained a consistent threshold across all experiments within the same dataset.
Our experiments yielded the best results on RoadObstacle21 dataset.

\begin{table*}[tb]
  \centering
  \scriptsize
    \caption{Comparison of Grounding DINO's performance on the RoadObstacle21 dataset using separate prompts, combined prompts, and the GD Prompt Ensemble method.}
  \begin{tabular}{c|ccccccccc}
    \toprule
      Prompt & AP\textsubscript{50..95} & AP\textsubscript{50} & AP\textsubscript{75} & AP\textsubscript{S} & AP\textsubscript{M} & AP\textsubscript{L} & TP & FP & FN \\
    \midrule
    Foreground objects & 0.483 & 0.768 & 0.521 & \textbf{0.362} & \textbf{0.713} & 0.830 & 457 & 350 & 100  \\
    
    Objects on road & 0.398 & 0.727 & 0.375 & 0.243 & 0.655 & 0.805 & 432 & 489 & 125  \\
    Combined prompt & 0.219 & 0.525 & 0.141 & 0.083 & 0.444 & 0.644 & 323 & 337 & 234  \\
    \midrule
    GD Prompt Ensemble & \textbf{0.485} & \textbf{0.776} & \textbf{0.522} & 0.359 & 0.712 & \textbf{0.841} & 457 & 350 & 100  \\
    GD Prompt Ensemble & \textbf{0.485} & 0.773 & \textbf{0.522} & 0.360 & \textbf{0.713} & 0.835 & 457 & 350 & 100 \\
    \bottomrule
  \end{tabular}
  \label{tab:GD1}
\end{table*}

\paragraph{\textbf{Prompt Engineering for RoadObstacle21:}}
In Table \ref{tab:result1} we present a selection of three textual prompts for RoadObstacle21 dataset, devised through a systematic process of experimentation and refinement. The prompts include \textit{'foreground objects'}, \textit{'objects on road'} and \textit{'unexpected objects on road'}. These prompts were meticulously crafted to get optimal performance from Grounding DINO.
However, we do not claim that there is no room for further improvement of the prompts. We used an objectness score threshold of 0.1. Our choice of such a low threshold was driven by a desire to maximize the retrieval of true positives from the model's predictions. However, this deliberate selection also led to an increase in the number of false positives. 

We strived to use very simple text prompts that could work well with the whole dataset. The idea behind using a simple but effective text prompt is to test if the model is able to grasp the underlying meaning behind the input text. Giving the name of each object present in the dataset would have oversimplified the task and would have made it extremely trivial for the model to locate objects.

Each prompt had varying results considering the sizes of the objects as well as with different IoU thresholds.
Grounding DINO achieved an AP\textsubscript{50} of 76.8\% using the prompt \textit{'foreground objects'}, while for AP\textsubscript{75}, AP\textsubscript{50..95} it achieved 52.1\% and 48.3 \% respectively. Indicating higher confidence in the predictions as compared to the predictions using other prompts. 
The other models significantly underperformed compared to Grounding DINO.
Figure \ref{fig:roadobstacle} in the Appendix, presents the qualitative results on RoadObstacle21.

\begin{table*}[tb]
  \centering
  \scriptsize
    \caption{Performance of various models on the RoadObstacle21 dataset with different prompts.}
  \begin{tabular}{c|cccccccccc}
    \toprule
    Model & Prompt & AP\textsubscript{50..95} & AP\textsubscript{50} & AP\textsubscript{75} & AP\textsubscript{S} & AP\textsubscript{M} & AP\textsubscript{L} & TP & FP & FN \\
    \midrule
    Grounding & Foreground objects & \textbf{0.483} & \textbf{0.768} & \textbf{0.521} & \textbf{0.362} & \textbf{0.713} & \textbf{0.830} & 457 & 350 & 100  \\
    
    DINO & Objects on road & 0.398 & 0.727 & 0.375 & 0.243 & 0.655 & 0.805 & 432 & 489 & 125  \\
    & Unexpected objects on road & 0.152 & 0.405 & 0.077 & 0.031 & 0.335 & 0.684 & 256 & 610 & 301  \\
    \midrule 
    YOLO-World & Foreground objects & 0.072 & 0.214 & 0.038 & 0.006 & 0.152 & 0.472 & 164 & 359 & 393  \\
    
    & Objects on road & 0.054 & 0.178 & 0.025 & 0.004 & 0.119 & 0.352 & 163 & 434 & 394  \\
    & Unexpected objects on road & 0.064 & 0.202 & 0.032 & 0.005 & 0.138 & 0.429 & 162 & 682 & 395  \\
    \midrule 
    MDetr & Foreground objects & 0.067 & 0.218 & 0.032 & 0.009 & 0.145 & 0.442 & 161 & 683 & 396  \\
    
    & Objects on road & 0.046 & 0.157 & 0.020 & 0.004 & 0.088 & 0.362 & 130 & 337 & 427  \\
    & Unexpected objects on road & 0.064 & 0.204 & 0.030 & 0.006 & 0.127 & 0.443 & 147 & 409 & 410  \\
     \midrule 
    OmDet & Foreground objects & 0.195 & 0.332 & 0.211 & 0.079 & 0.404 & \textbf{0.573} & 195 & 29 & 362  \\
    
    & Objects on road & \textbf{0.284} & \textbf{0.473} & \textbf{0.300} & \textbf{0.199} & \textbf{0.458} & 0.493 & 278 & 39 & 279  \\
    
    & Unexpected objects on road & 0.195 & 0.362 & 0.192 & 0.119 & 0.353 & 0.428 & 221 & 30 & 336  \\    
    \bottomrule
  \end{tabular}
  \label{tab:result1}
\end{table*}

\begin{table}[tb]
  \centering
    \scriptsize
  \caption{Performance on LostAndFound using Grounding DINO. In this table the complete Prompt 1 is: \emph{Unusual items unexpectedly found along an isolated road},
Prompt 2: \emph{Unexpected items abandoned scattered along a road}}
\begin{tabular}{@{}lp{3cm}llllllllll@{}}

    \toprule
    Subset & Prompt & AP\textsubscript{10} & AP\textsubscript{10..75} & AP\textsubscript{20..75} & AP\textsubscript{50..95} & AP\textsubscript{S} & AP\textsubscript{M} & AP\textsubscript{L} & TP & FP & FN \\
    \midrule
    1 & 1. Unusual items.. & 0.323 & 0.159 & 0.134 & 0.032 & \textbf{0.228} & 0.620 & 0.752 & 208 & 111 & 339  \\
    & 2. Items abandoned.. & 0.158 & 0.094 & 0.084 & 0.024 & 0.119 & 0.689 & 0.353 & 201 & 576 & 346  \\
    & 3. Unexpected items & 0.286 & 0.283 & 0.283 & 0.233 & 0.063 & 0.817 & 0.986 & 192 & 232 & 355  \\
    & 4. Objects & \textbf{0.330} & \textbf{0.322} & \textbf{0.322} & \textbf{0.258} & 0.100 & \textbf{0.839} & \textbf{0.987} & 219 & 218 & 328  \\
    \midrule
    2 & 1. Unusual items.. & 0.396 & 0.202 & 0.172 & 0.045 & \textbf{0.267} & 0.786 & 0.498 & 232 & 185 & 236  \\
     & 2. Items abandoned.. & 0.282 & 0.161 & 0.145 & 0.041 & 0.119 & 0.687 & 0.573 & 175 & 128 & 293 \\
     & 3. Unexpected items & 0.339 & 0.316 & 0.314 & 0.229 & 0.168 & 0.746 & 0.925 & 201 & 120 & 267 \\
     & 4. Objects & \textbf{0.401} & \textbf{0.379} & \textbf{0.376} & \textbf{0.273} & 0.257 & \textbf{0.826} & \textbf{0.921} & 238 & 122 & 230 
     \\ 
    \midrule
    3 & 1. Unusual items.. & 0.321 & 0.183 & 0.164 & 0.043 & \textbf{0.208} & 0.663 & 0.631 & 178 & 246 & 211  \\
    & 2. Items abandoned.. & 0.273 & 0.184 & 0.172 & 0.058 & 0.059 & 0.655 & 0.752 & 132 & 159 & 257  \\
    & 3. Unexpected items & 0.339 & 0.330 & 0.329 & 0.255 & 0.112 & 0.671 & \textbf{0.974} & 147 & 79 & 242  \\
    & 4. Objects & \textbf{0.379} & \textbf{0.364} & \textbf{0.363} & \textbf{0.277} & 0.162 & \textbf{0.720} & 0.958 & 168 & 79 & 221  \\
    \midrule
    4 & 1. Unusual items.. & \textbf{0.464} & 0.222 & 0.184 & 0.048 & \textbf{0.307} & \textbf{0.873} & 0.842 & 168 & 86 & 153  \\
    & 2. Items abandoned.. & 0.307 & 0.170 & 0.152 & 0.044 & 0.099 & 0.789 & 0.475 & 106 & 48 & 215  \\
    & 3. Unexpected items & 0.228 & 0.226 & 0.226 & 0.183 & 0.040 & 0.545 & 0.683 & 73 & 1 & 248  \\
    & 4. Objects & 0.340 & \textbf{0.312} & \textbf{0.308} & \textbf{0.239} & 0.129 & 0.861 & \textbf{0.994} & 127 & 54 & 194  \\
    \midrule
    5 & 1. Living beings & \textbf{0.527} & \textbf{0.362} & \textbf{0.335} & \textbf{0.225} & \textbf{0.584} & \textbf{0.696} & \textbf{0.636} & 177 & 66 & 26 \\
    \bottomrule
      \end{tabular}
\label{tab:result2}
\end{table}

\begin{table}[tb]
  \centering
  \setlength{\tabcolsep}{2pt}
  \scriptsize
    \caption{Performance on RoadAnomaly21. In this table the results are based on the aggregated prompt and that refers to 'foreground big objects . sheep . animals'}
  \begin{tabular}{c|cccccccccc}
    \toprule
    Model & AP\textsubscript{50..95} & AP\textsubscript{50} & AP\textsubscript{75} & AP\textsubscript{S} & AP\textsubscript{M} & AP\textsubscript{L} & AR\textsubscript{10}  & TP & FP & FN \\
    \midrule
    Grounding DINO & 0.195 & 0.270 & 0.203 & \textbf{0.069} & 0.156 & 0.408 & 0.270 & 328 & 309 & 411 \\
    YOLO-World & \textbf{0.212} & \textbf{0.334} & \textbf{0.211} & 0.038 & \textbf{0.158} & \textbf{0.466} & \textbf{0.334} & 368 & 621 & 371 \\ 
    MDetr & 0.149 & 0.261 & 0.153 & 0.002 & 0.039 & 0.407 & 0.261 & 285 & 647 & 454  \\
    OmDet & 0.113 & 0.160 & 0.116 & 0.014 & 0.066 & 0.277 & 0.160 & 336 & 312 & 403 \\
    \midrule
    GD Prompt Ensemble & \textbf{0.227} & \textbf{0.333} & \textbf{0.232} & \textbf{0.142} & \textbf{0.172} & \textbf{0.409} & \textbf{0.333} & 322 & 184 & 417 \\
    \bottomrule
  \end{tabular}
  \label{tab:anomaly}
\end{table}

\paragraph{\textbf{Prompt Engineering for LostAndFound:}}
In Table \ref{tab:result2} we present the experimental results on LostAndFound using Grounding DINO. We treated each location within this dataset individually. Due to the notable uneven distribution of images across those locations, we decided to evaluate them separately into distinct subsets. We represent four prompts for each subsection, however, for the subset 5 we show only one prompt which alone performed quite well. This subset was also fairly different than the other subsets, as it contained humans as anomalies while the rest contained random objects.
The predominant reasons for the increased number of false positives were the multiple detections for the same object. Additionally, the presence of road features such as gutters and roadside drainage frequently resulted in misclassifications as objects during the detection process. 

We observe an interesting pattern in Table \ref{tab:result2}, where prompt \textit{'unusual items unexpectedly found along an isolated road'} consistently obtained better AP for small objects as well as for medium sized objects in subset 4. While in subset 3, the prompt \textit{'unexpected items'} achieved an AP of 97.4\% for large objects. In rest of the cases the most simple and short prompt \textit{'objects'} resulted in superior results. For subset 5 we used a different prompt \textit{'living beings'}, it resulted in a 
significantly low FN count of 26 with an optimal score of 52.7\% for AP\textsubscript{10}.
In Appendix Table \ref{tab:result2.2}, we present results using other open-vocabulary models. Whereas the results on Fishyscapes L\&F can be found in Appendix Table \ref{tab:fs}. Additionally, we illustrate the qualitative results in Appendix Figure \ref{fig:lostandfound}.

\paragraph{\textbf{Prompt Engineering for RoadAnomaly21:}}
Table \ref{tab:anomaly} presents the results for RoadAnomaly21 dataset. This dataset is fairly different from typical road scene datasets. Here, we used a combination of three different prompts together such as 
 \textit{'foreground big objects', 'sheep', 'animals'}. Interestingly this combination worked better rather than a single text. Changing the prompt from \textit{'foreground objects'} to \textit{'foreground big objects'} resulted in improved results. As this dataset contained larger objects as compared to RoadObstacle21, where the prompt \textit{'foreground objects'} performed quite well, adding the keyword \textit{'big'} helped the model locate objects better.

The main challenges with this dataset are the large size of the objects, that often occupy nearly half of the image, leading to predictions that cover the entire image.
Additionally, in many cases the objects are either very close to each other or are overlapping, leading to a single prediction. The dataset predominantly contained animals specifically a flock of sheep. Thus, we also considered more specific prompt such as \textit{'sheep'} and \textit{'animals'}.
With GDino Prompt Ensemble we used only two prompts i.e., \textit{'animals'} and \textit{'foreground big objects'} with a uniform weight distribution of 0.5--0.5.

Recognizing the presence of animals in a road scene, despite their status as out-of-distribution objects, should be possible for any open-vocabulary object detector. However, the challenge arises when attempting to localize each individual animal within a flock. While the model may successfully identify the presence of animals, discerning each individual is challenging. In order to see whether Grounding DINO is able to find most of them, we provided the models with more help by using a combination of specific and precise prompts. However, it is worth noting that we did not use the specific prompt \textit{'sheep'} in the GD Prompt Ensemble method, yet it outperforms the other methods.
Figure \ref{fig:roadanomaly} in Appendix presents the qualitative results on RoadAnomaly21.

\subsection{Discussion}

\paragraph{\textbf{Models Performance Across Datasets:}}
Grounding DINO consistently outperformed other models on RoadObstacle21 dataset with a notable margin (see Table \ref{tab:result1}) achieving a high score of 48\% for AP\textsubscript{50..95}, suggesting accurate localization of the predicted objects as well as higher confidence levels. In contrast, the other models struggled on this specific metric.
OmDet reached the second rank with 28.4\% for AP\textsubscript{50..95} metric. This model mostly struggled with smaller objects thus resulting in a lower count of TPs but at the same time performing well on medium and larger objects. In general this model achieved fairly low FP counts.
A different pattern was noted with YOLO-World and MDETR, both resulted in lower count of TP with a higher number of FP and considerably lagged behind Grounding DINO and OmDet. 

MDETR and Grounding DINO achieved similar results for AP\textsubscript{10} on the LostAndFound dataset. However, Grounding DINO exhibited superior performance on stricter metric AP\textsubscript{50..95} YOLO-World showed good performance on detecting medium and large objects but was significantly inferior in detecting small objects. Grounding DINO consistently performed well with shorter prompts. In Table \ref{tab:result2} subset 1 of the LostAndFound dataset, Grounding DINO obtained a significantly lower score of 15.8\% for AP\textsubscript{10} with the prompt 'unexpected items abandoned scattered along a road', whereas 
MDETR achieved better performance using the same prompt resulting in a comparatively higher AP\textsubscript{10} score of 38.9\%. See Table \ref{tab:result2.2} in Appendix for the results on MDETR, YOLO-World and OmDet.

In Table \ref{tab:GD1}, our experiments revealed that combining multiple prompts yielded suboptimal results for Grounding DINO, compared to when individual prompts were applied separately.
However, on RoadAnomaly21 dataset shown in table \ref{tab:anomaly}, the combined prompt resulted in fairly good performance, with YOLO-World outperforming all other models. Since this dataset consists of large and closely placed objects. This suggests that YOLO-World is able to distinguish better between overlapping objects and not considering them as a single entity. It is also worth noting that MDETR performs as good as Grounding DINO with respect to the AP for large objects on RoadAnomaly21, achieving a score of 40\%, shown in Table \ref{tab:anomaly}.

\begin{figure}[tb]
\centering
  \begin{subfigure}{0.38\textwidth}
    \includegraphics[width=\linewidth, height=2.4cm]{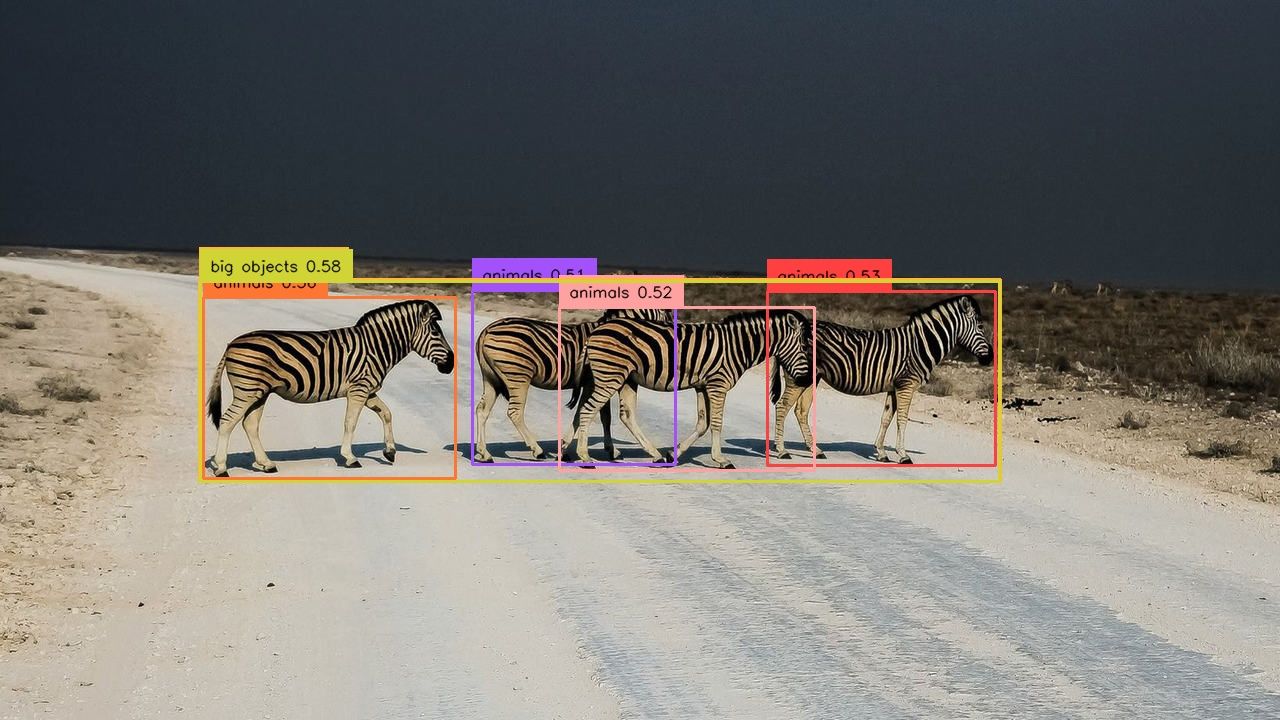}
  \end{subfigure}
  \begin{subfigure}{0.38\textwidth}
    \includegraphics[width=\linewidth, height=2.4cm]{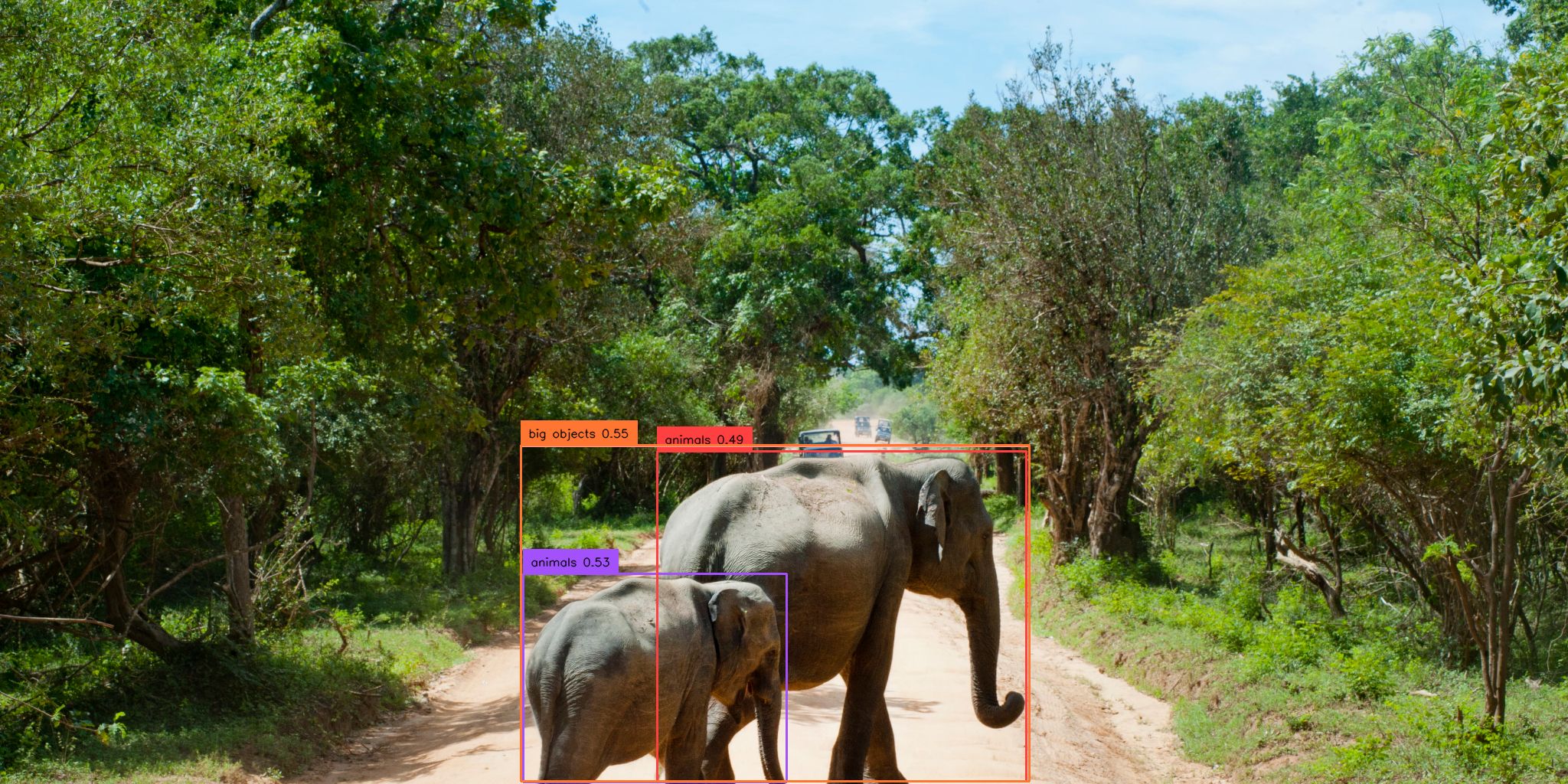}
  \end{subfigure}
\caption{An illustration of multiple predictions with different labels, where the labels assigned the highest confidence are the least precise in the context of the scene.}
\label{fig:fig5}
\end{figure}

\paragraph{\textbf{Limitations of Current Object Detection Models:}}
In Table \ref{tab:discussion} we present results on all object detection models experimented. Where it can be clearly seen that open-vocabulary models, as expected Grounding DINO and OmDet, outperform YOLOv8 object detection model and the VOS OOD object detection method. 
Since, open-vocabulary models are trained on extensive amount of data, it is rational to expect them to perform better than a basic object detection model. This broader knowledge base equips them with the ability to outperform more basic object detection models. On the other hand, VOS focuses on correctly classifying ID and OOD objects for given predictions, rather than optimizing the localization aspect by maximizing the detection of OOD objects.
We further observed that open-vocabulary models do not consistently interpret the semantic meaning of text prompts accurately. Specifically, these models may struggle with negations, which is in accordance with \cite{jang2023can}, resulting in erroneous predictions for objects that are explicitly negated. Additionally, we frequently encountered multiple predictions for the same object, each with different labels and varying confidence scores, shown in Figure \ref{fig:fig5}. We also identified discrepancies in the precision of bounding boxes associated with different confidence scores, where predictions with higher confidence sometimes exhibited lower precision compared to their lower confidence counterparts.

\begin{table}[tb]
\centering
\scriptsize
  \caption{Comparative results of open-vocabulary and object detection models on RoadObstacle21.}
  \setlength{\tabcolsep}{2pt} 
  \begin{tabular}{c|ccccccccc}
    \toprule
    \textbf{Method} & \textbf{AP\textsubscript{50..95}} & \textbf{AP\textsubscript{50}} & \textbf{AP\textsubscript{75}} & \textbf{AP\textsubscript{S}} & \textbf{AP\textsubscript{M}} & \textbf{AP\textsubscript{L}} & \textbf{TP} & \textbf{FP} & \textbf{FN} \\
    \midrule
    MDETR & 0.067 & 0.218 & 0.032 & 0.009 & 0.145 & 0.442 & 161 & 683 & 396 \\
    YOLO-World & 0.072 & 0.214 & 0.038 & 0.006 & 0.152 & 0.472 & 164 & 359 & 393  \\
    OmDet-Turbo & 0.284 & 0.473 & 0.300 & 0.199 & 0.458 & 0.493 & 278 & 39 & 279 \\
    Grounding DINO & \textbf{0.483} & \textbf{0.768} & \textbf{0.521} & \textbf{0.362} & \textbf{0.713} & \textbf{0.830} & 457 & 350 & 100 \\ 

    \midrule
    YOLOv8-L & 0.215 & 0.343 & 0.225 & 0.125 & 0.440 & 0.551 & 412 & 3994 & 145 \\
    \midrule
    VOS & 0.022 & 0.058 & 0.011 & 0.016 & 0.050 & 0.064 & 56 & 91 & 501 \\
    \bottomrule
  \end{tabular}
  \label{tab:discussion}
\end{table}

\section{Conclusion}
In this work we present a study on three challenging datasets for OOD object detection using open-vocabulary models. Our experiments reveal the potential of open-vocabulary models for addressing the OOD object detection problem, a domain where their application has not been explored previously.
Our findings show that open-vocabulary models have a lot of potential for OOD object detection tasks. 
Despite those promising capabilities, there are clear limitations observed in certain scenarios. The presented results, although encouraging, are far from perfect. We anticipate that with further advancements and refinement, these limitations can be effectively addressed over time. Furthermore, we provide the community with a simple prompt-based baseline for OOD object detection.

\clearpage  

%
%
\bibliographystyle{splncs04}
\bibliography{main}

\begin{thebibliography}{10}
\providecommand{\url}[1]{\texttt{#1}}
\providecommand{\urlprefix}{URL }
\providecommand{\doi}[1]{https://doi.org/#1}

\bibitem{blum2019fishyscapes}
Blum, H., Sarlin, P.E., Nieto, J., Siegwart, R., Cadena, C.: Fishyscapes: A benchmark for safe semantic segmentation in autonomous driving. In: proceedings of the IEEE/CVF international conference on computer vision workshops. pp.~0--0 (2019)

\bibitem{bommasani2021opportunities}
Bommasani, R., Hudson, D.A., Adeli, E., Altman, R., Arora, S., von Arx, S., Bernstein, M.S., Bohg, J., Bosselut, A., Brunskill, E., et~al.: On the opportunities and risks of foundation models. arXiv preprint arXiv:2108.07258  (2021)

\bibitem{carion2020end}
Carion, N., Massa, F., Synnaeve, G., Usunier, N., Kirillov, A., Zagoruyko, S.: End-to-end object detection with transformers. In: European conference on computer vision. pp. 213--229. Springer (2020)

\bibitem{chan2021segmentmeifyoucan}
Chan, R., Lis, K., Uhlemeyer, S., Blum, H., Honari, S., Siegwart, R., Fua, P., Salzmann, M., Rottmann, M.: Segmentmeifyoucan: A benchmark for anomaly segmentation. arXiv preprint arXiv:2104.14812  (2021)

\bibitem{chan2021entropy}
Chan, R., Rottmann, M., Gottschalk, H.: Entropy maximization and meta classification for out-of-distribution detection in semantic segmentation. In: Proceedings of the ieee/cvf international conference on computer vision. pp. 5128--5137 (2021)

\bibitem{cheng2022masked}
Cheng, B., Misra, I., Schwing, A.G., Kirillov, A., Girdhar, R.: Masked-attention mask transformer for universal image segmentation. In: Proceedings of the IEEE/CVF conference on computer vision and pattern recognition. pp. 1290--1299 (2022)

\bibitem{cheng2024yolo}
Cheng, T., Song, L., Ge, Y., Liu, W., Wang, X., Shan, Y.: Yolo-world: Real-time open-vocabulary object detection. arXiv preprint arXiv:2401.17270  (2024)

\bibitem{Cordts2016Cityscapes}
Cordts, M., Omran, M., Ramos, S., Rehfeld, T., Enzweiler, M., Benenson, R., Franke, U., Roth, S., Schiele, B.: The cityscapes dataset for semantic urban scene understanding. In: Proc. of the IEEE Conference on Computer Vision and Pattern Recognition (CVPR) (2016)

\bibitem{devlin2018bert}
Devlin, J., Chang, M.W., Lee, K., Toutanova, K.: Bert: Pre-training of deep bidirectional transformers for language understanding. arXiv preprint arXiv:1810.04805  (2018)

\bibitem{di2021pixel}
Di~Biase, G., Blum, H., Siegwart, R., Cadena, C.: Pixel-wise anomaly detection in complex driving scenes. In: Proceedings of the IEEE/CVF conference on computer vision and pattern recognition. pp. 16918--16927 (2021)

\bibitem{du2024uncovering}
Du, H., Zhang, S., Xie, B., Nan, G., Zhang, J., Xu, J., Liu, H., Leng, S., Liu, J., Fan, H., et~al.: Uncovering what why and how: A comprehensive benchmark for causation understanding of video anomaly. In: Proceedings of the IEEE/CVF Conference on Computer Vision and Pattern Recognition. pp. 18793--18803 (2024)

\bibitem{du2022vos}
Du, X., Wang, Z., Cai, M., Li, Y.: Vos: Learning what you don't know by virtual outlier synthesis. arXiv preprint arXiv:2202.01197  (2022)

\bibitem{elhafsi2023semantic}
Elhafsi, A., Sinha, R., Agia, C., Schmerling, E., Nesnas, I.A., Pavone, M.: Semantic anomaly detection with large language models. Autonomous Robots  \textbf{47}(8),  1035--1055 (2023)

\bibitem{esmaeilpour2022zero}
Esmaeilpour, S., Liu, B., Robertson, E., Shu, L.: Zero-shot out-of-distribution detection based on the pre-trained model clip. In: Proceedings of the AAAI conference on artificial intelligence. vol.~36, pp. 6568--6576 (2022)

\bibitem{gasperini2023segmenting}
Gasperini, S., Marcos-Ramiro, A., Schmidt, M., Navab, N., Busam, B., Tombari, F.: Segmenting known objects and unseen unknowns without prior knowledge. In: Proceedings of the IEEE/CVF international conference on computer vision. pp. 19321--19332 (2023)

\bibitem{gu2024anomalygpt}
Gu, Z., Zhu, B., Zhu, G., Chen, Y., Tang, M., Wang, J.: Anomalygpt: Detecting industrial anomalies using large vision-language models. In: Proceedings of the AAAI Conference on Artificial Intelligence. vol.~38, pp. 1932--1940 (2024)

\bibitem{hendrycks2016baseline}
Hendrycks, D., Gimpel, K.: A baseline for detecting misclassified and out-of-distribution examples in neural networks. arXiv preprint arXiv:1610.02136  (2016)

\bibitem{hendrycks2019oe}
Hendrycks, D., Mazeika, M., Dietterich, T.: Deep anomaly detection with outlier exposure. Proceedings of the International Conference on Learning Representations  (2019)

\bibitem{huang2021importance}
Huang, R., Geng, A., Li, Y.: On the importance of gradients for detecting distributional shifts in the wild. Advances in Neural Information Processing Systems  \textbf{34},  677--689 (2021)

\bibitem{jang2023can}
Jang, J., Ye, S., Seo, M.: Can large language models truly understand prompts? a case study with negated prompts. In: Transfer learning for natural language processing workshop. pp. 52--62. PMLR (2023)

\bibitem{jiang2024dog}
Jiang, H., Fang, Z., Jiang, X., Zhong, Z., Liu, T., Han, B.: {DOG}: Diffusion-based outlier generation for out-of-distribution detection (2024), \url{https://openreview.net/forum?id=Go8hf9wKJx}

\bibitem{Jocher_Ultralytics_YOLO_2023}
Jocher, G., Chaurasia, A., Qiu, J.: {Ultralytics YOLO} (Jan 2023), \url{https://github.com/ultralytics/ultralytics}

\bibitem{kamath2021mdetr}
Kamath, A., Singh, M., LeCun, Y., Synnaeve, G., Misra, I., Carion, N.: Mdetr-modulated detection for end-to-end multi-modal understanding. In: Proceedings of the IEEE/CVF International Conference on Computer Vision. pp. 1780--1790 (2021)

\bibitem{lakshminarayanan2017simple}
Lakshminarayanan, B., Pritzel, A., Blundell, C.: Simple and scalable predictive uncertainty estimation using deep ensembles. Advances in neural information processing systems  \textbf{30} (2017)

\bibitem{lee2018simple}
Lee, K., Lee, K., Lee, H., Shin, J.: A simple unified framework for detecting out-of-distribution samples and adversarial attacks. Advances in neural information processing systems  \textbf{31} (2018)

\bibitem{li2024anomaly}
Li, A., Zhao, Y., Qiu, C., Kloft, M., Smyth, P., Rudolph, M., Mandt, S.: Anomaly detection of tabular data using llms. arXiv preprint arXiv:2406.16308  (2024)

\bibitem{li2023trustworthy}
Li, B., Qi, P., Liu, B., Di, S., Liu, J., Pei, J., Yi, J., Zhou, B.: Trustworthy ai: From principles to practices. ACM Computing Surveys  \textbf{55}(9),  1--46 (2023)

\bibitem{liang2017enhancing}
Liang, S., Li, Y., Srikant, R.: Enhancing the reliability of out-of-distribution image detection in neural networks. arXiv preprint arXiv:1706.02690  (2017)

\bibitem{lis2019detecting}
Lis, K., Nakka, K., Fua, P., Salzmann, M.: Detecting the unexpected via image resynthesis. In: Proceedings of the IEEE/CVF International Conference on Computer Vision. pp. 2152--2161 (2019)

\bibitem{liu2023grounding}
Liu, S., Zeng, Z., Ren, T., Li, F., Zhang, H., Yang, J., Li, C., Yang, J., Su, H., Zhu, J., et~al.: Grounding dino: Marrying dino with grounded pre-training for open-set object detection. arXiv preprint arXiv:2303.05499  (2023)

\bibitem{liu2020energy}
Liu, W., Wang, X., Owens, J., Li, Y.: Energy-based out-of-distribution detection. Advances in neural information processing systems  \textbf{33},  21464--21475 (2020)

\bibitem{liu2019roberta}
Liu, Y., Ott, M., Goyal, N., Du, J., Joshi, M., Chen, D., Levy, O., Lewis, M., Zettlemoyer, L., Stoyanov, V.: Roberta: A robustly optimized bert pretraining approach. arXiv preprint arXiv:1907.11692  (2019)

\bibitem{liu2021swin}
Liu, Z., Lin, Y., Cao, Y., Hu, H., Wei, Y., Zhang, Z., Lin, S., Guo, B.: Swin transformer: Hierarchical vision transformer using shifted windows. In: Proceedings of the IEEE/CVF international conference on computer vision. pp. 10012--10022 (2021)

\bibitem{mukhoti2018evaluating}
Mukhoti, J., Gal, Y.: Evaluating bayesian deep learning methods for semantic segmentation. arXiv preprint arXiv:1811.12709  (2018)

\bibitem{nekrasov2023ugains}
Nekrasov, A., Hermans, A., Kuhnert, L., Leibe, B.: Ugains: Uncertainty guided anomaly instance segmentation. In: DAGM German Conference on Pattern Recognition. pp. 50--66. Springer (2023)

\bibitem{nekrasov2024oodis}
Nekrasov, A., Zhou, R., Ackermann, M., Hermans, A., Leibe, B., Rottmann, M.: Oodis: Anomaly instance segmentation benchmark. arXiv preprint arXiv:2406.11835  (2024)

\bibitem{oberdiek2022uqgan}
Oberdiek, P., Fink, G., Rottmann, M.: Uqgan: A unified model for uncertainty quantification of deep classifiers trained via conditional gans. Advances in Neural Information Processing Systems  \textbf{35},  21371--21385 (2022)

\bibitem{pinggeralost}
Pinggera, P., Ramos, S., Gehrig, S., Franke, U., Rother, C., Mester, R.: Lost and found: detecting small road hazards for self-driving vehicles. in 2016 ieee. In: RSJ International Conference on Intelligent Robots and Systems (IROS). pp. 1099--1106

\bibitem{radford2021learning}
Radford, A., Kim, J.W., Hallacy, C., Ramesh, A., Goh, G., Agarwal, S., Sastry, G., Askell, A., Mishkin, P., Clark, J., et~al.: Learning transferable visual models from natural language supervision. In: International conference on machine learning. pp. 8748--8763. PMLR (2021)

\bibitem{rai2023unmasking}
Rai, S.N., Cermelli, F., Fontanel, D., Masone, C., Caputo, B.: Unmasking anomalies in road-scene segmentation. In: Proceedings of the IEEE/CVF International Conference on Computer Vision. pp. 4037--4046 (2023)

\bibitem{sun2024clip}
Sun, H., He, R., Han, Z., Lin, Z., Gong, Y., Yin, Y.: Clip-driven outliers synthesis for few-shot ood detection. arXiv preprint arXiv:2404.00323  (2024)

\bibitem{wang2023clipn}
Wang, H., Li, Y., Yao, H., Li, X.: Clipn for zero-shot ood detection: Teaching clip to say no. In: Proceedings of the IEEE/CVF International Conference on Computer Vision. pp. 1802--1812 (2023)

\bibitem{wilson2023safe}
Wilson, S., Fischer, T., Dayoub, F., Miller, D., S{\"u}nderhauf, N.: Safe: Sensitivity-aware features for out-of-distribution object detection. In: Proceedings of the IEEE/CVF International Conference on Computer Vision. pp. 23565--23576 (2023)

\bibitem{xu2024customizing}
Xu, X., Cao, Y., Chen, Y., Shen, W., Huang, X.: Customizing visual-language foundation models for multi-modal anomaly detection and reasoning. arXiv preprint arXiv:2403.11083  (2024)

\bibitem{yu2020bdd100k}
Yu, F., Chen, H., Wang, X., Xian, W., Chen, Y., Liu, F., Madhavan, V., Darrell, T.: Bdd100k: A diverse driving dataset for heterogeneous multitask learning. In: Proceedings of the IEEE/CVF conference on computer vision and pattern recognition. pp. 2636--2645 (2020)

\bibitem{zhang2022dino}
Zhang, H., Li, F., Liu, S., Zhang, L., Su, H., Zhu, J., Ni, L.M., Shum, H.Y.: Dino: Detr with improved denoising anchor boxes for end-to-end object detection. arXiv preprint arXiv:2203.03605  (2022)

\bibitem{zhao2024real}
Zhao, T., Liu, P., He, X., Zhang, L., Lee, K.: Real-time transformer-based open-vocabulary detection with efficient fusion head. arXiv preprint arXiv:2403.06892  (2024)

\end{thebibliography}

\appendix
\section{Supplementary}

\subsection{Results on LostAndFound}
Experimental results on LostAndFound can be found in the Table \ref{tab:result2.2} below. The LostAndFound test set is divided into five different locations, thus we treat those locations separate as subsets and then average the results to get combined results on the complete dataset. The results based on AP\textsubscript{50..95} are similar for all three models. However, MDETR outperforms the other models on the metric AP\textsubscript{10}, AP\textsubscript{10..75} and AP\textsubscript{20..75}.
\begin{table}[!h]
  \centering
  \scriptsize
    \caption{Text 1: \emph{unusual items unexpectedly found along an isolated road},
  Text 2: \emph{unexpected items abandoned scattered along a road}, Text 3: \emph{unexpected items}, Text 4: \emph{objects}, Text 5: \emph{living beings}}
  \begin{tabular}{@{}ccccccccccccc@{}} 
    \toprule
    & & \multicolumn{10}{c}{Performance on LostAndFound using other models} \\
    \cmidrule(lr){4-13}
    Model & Subset & Prompt & AP\textsubscript{10} & AP\textsubscript{10..75} & AP\textsubscript{20..75} & AP\textsubscript{50..95} & AP\textsubscript{S} & AP\textsubscript{M} & AP\textsubscript{L} & TP & FP & FN \\
    \toprule
    MDETR & 1 & Text 2 & 0.389 & 0.193 & 0.166 & 0.049 & 0.307 & 0.785 & 0.822 & 264 & 647 & 283  \\
    & 2 & Text 1 & 0.342 & 0.188 & 0.166 & 0.053 & 0.248 & 0.691 & 0.791 & 215 & 225 & 253  \\
    & 3 & Text 3 & 0.382 & 0.245 & 0.225 & 0.082 & 0.188 & 0.723 & 0.986 & 174 & 354 & 215  \\ 
    & 4 & Text 4 & 0.407 & 0.222 & 0.184 & 0.048 & 0.287 & 0.780 & 0.555 & 149 & 98 & 172  \\
    & 5 & Text 5 & 0.498 & 0.208 & 0.177 & 0.054 & 0.485 & 0.610 & 0.879 & 153 & 32 & 50 \\
    \midrule
    \textbf{Average} & & & \textbf{0.404} & \textbf{0.211} & \textbf{0.184} & \textbf{0.057} & \textbf{0.303} & \textbf{0.718} & \textbf{0.807} & \textbf{191} & \textbf{271} & \textbf{195}
 \\

    \midrule

    YOLO-World & 1 & Text 2 & 0.235 & 0.148 & 0.135 & 0.042 & 0.129 & 0.859 & 0.759 & 204 & 487 & 343  \\
    & 2 & Text 1 & 0.246 & 0.155 & 0.145 & 0.053 & 0.045 & 0.716 & 0.751 & 173 & 259 & 295  \\
    & 3 & Text 3 & 0.203 & 0.148 & 0.143 & 0.061 & 0.040 & 0.453 & 0.774 & 99 & 117 & 290  \\
    & 4 & Text 4 & 0.269 & 0.162 & 0.144 & 0.050 & 0.119 & 0.712 & 0.903 & 112 & 154 & 209  \\
    & 5 & Text 5 & 0.491 & 0.276 & 0.242 & 0.084 & 0.277 & 0.739 & 0.895 & 129 & 24 & 74 \\
    \midrule
    \textbf{Average} & & & \textbf{0.289} & \textbf{0.178} & \textbf{0.162} & \textbf{0.058} & \textbf{0.122} & \textbf{0.696} & \textbf{0.816} & \textbf{143} & \textbf{208} & \textbf{242} \\
    \midrule

    OmDet & 1 & Text 2 & 0.248 & 0.131 & 0.114 & 0.033 & 0.490 & 0.742 & 0.558 & 353 & 867 & 194  \\ 
    & 2 & Text 1 & 0.113 & 0.080 & 0.075 & 0.035 & 0.224 & 0.613 & 0.315 & 242 & 950 & 226  \\ 
    & 3 & Text 3 & 0.029 & 0.029 & 0.029 & 0.024 & 0.0 & 0.030 & 0.264 & 10 & 2 & 379 \\ 
    & 4 & Text 4 & 0.182 & 0.142 & 0.134 & 0.100 & 0.030 & 0.406 & 0.704 & 58 & 10 & 263  \\
    & 5 & Text 5 & 0.132 & 0.121 & 0.120 & 0.080 & 0.040 & 0.329 & 0.047 & 50 & 59 & 153 \\
    \midrule
    \textbf{Average} & & & \textbf{0.141} & \textbf{0.101} & \textbf{0.094} & \textbf{0.054} & \textbf{0.157} & \textbf{0.424} & \textbf{0.378} & \textbf{143} & \textbf{378} & \textbf{243} \\

    \bottomrule
  \end{tabular}
  \label{tab:result2.2}
\end{table}
\subsection{Results on Fishyscapes L\&F}
Since Fishyscapes L\&F is the filtered subset of LostAndFound, we submitted the results to the OoDIS benchmark only with the prompt '\textit{objects}'.
As shown in Table \ref{tab:fs}, Grounding DINO out performs all the other models with a significant margin on all metrics. MDETR ranks second as in the experiments on LostAndFound.

\begin{table*}[!h]
  \centering
  \setlength{\tabcolsep}{2pt}
  \scriptsize
    \caption{In this table the results are based on the prompt \textit{'objects'}}
  \begin{tabular}{c|cccccccccc}
    \toprule
    & \multicolumn{10}{c}{Performance on Fishyscapes L\&F} \\
    \cmidrule(lr){2-11}
    Model & AP\textsubscript{50..95} & AP\textsubscript{50} & AP\textsubscript{75} & AP\textsubscript{S} & AP\textsubscript{M} & AP\textsubscript{L} & AR\textsubscript{10}  & TP & FP & FN \\
    \midrule
    Grounding DINO & 0.227 & 0.334 & 0.245 & 0.098 & 0.481 & 0.362 & 0.334 & 173 & 170 & 266 \\
    YOLO-World & 0.050 & 0.154 & 0.010 & 0.002 & 0.120 & 0.163 & 0.074 & 99 & 340 & 340 \\ 
    MDetr & 0.059 & 0.184 & 0.026 & 0.007 & 0.130 & 0.174 & 0.184 & 114 & 547 & 325  \\
    OmDet & 0.024 & 0.055 & 0.013 & 0.002 & 0.075 & 0.041 & 0.055 & 68 & 328 & 371 \\
    \bottomrule
  \end{tabular}
  \label{tab:fs}
\end{table*} 
\newpage
\subsection{Qualitative results}
\begin{figure}[!h]
\centering
  \captionsetup{justification=centering,margin=0.5cm}
  \begin{subfigure}{0.25\textwidth}
    \includegraphics[width=\linewidth]{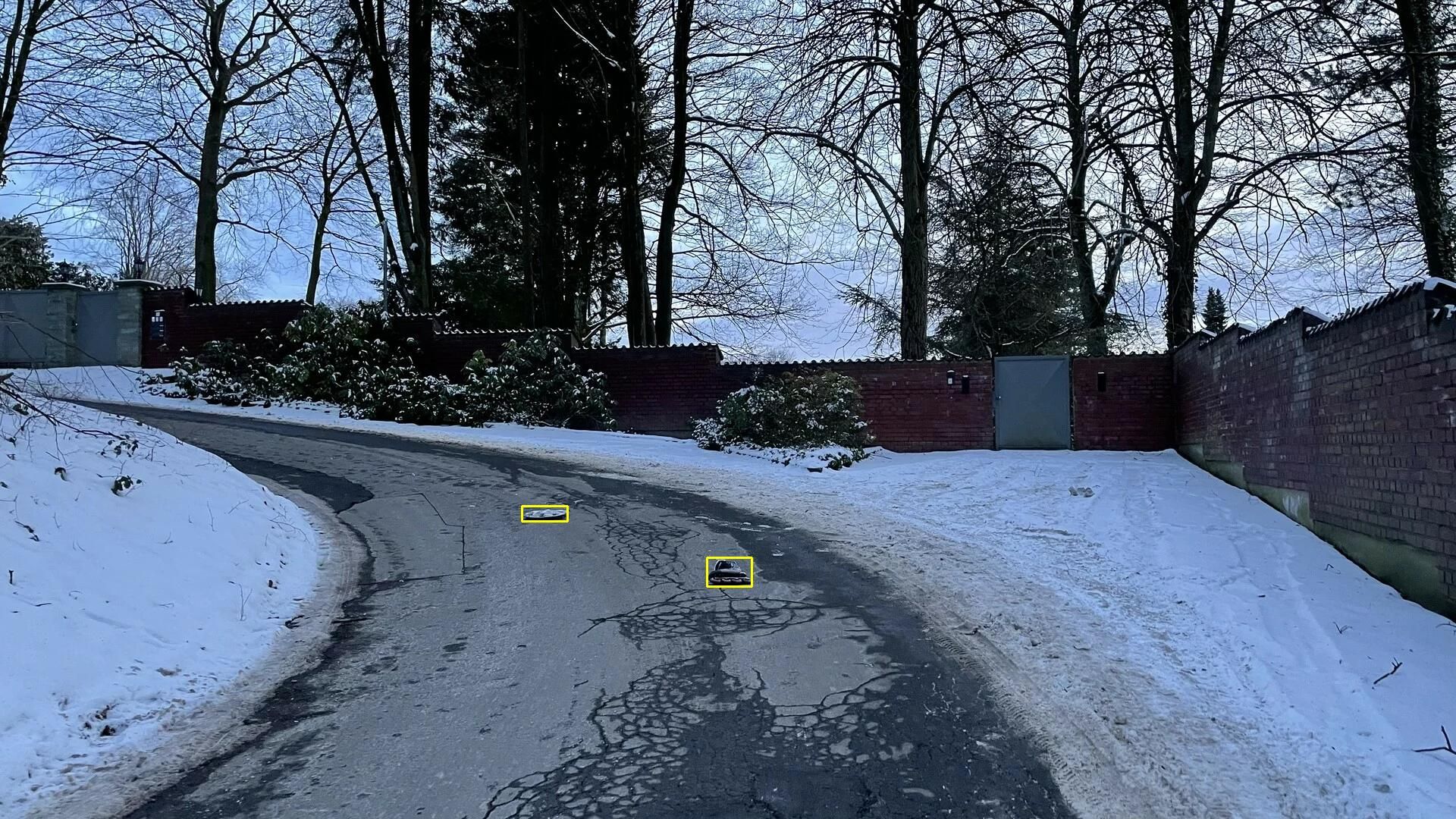}
  \end{subfigure}%
  \begin{subfigure}{0.25\textwidth}
    \includegraphics[width=\linewidth]{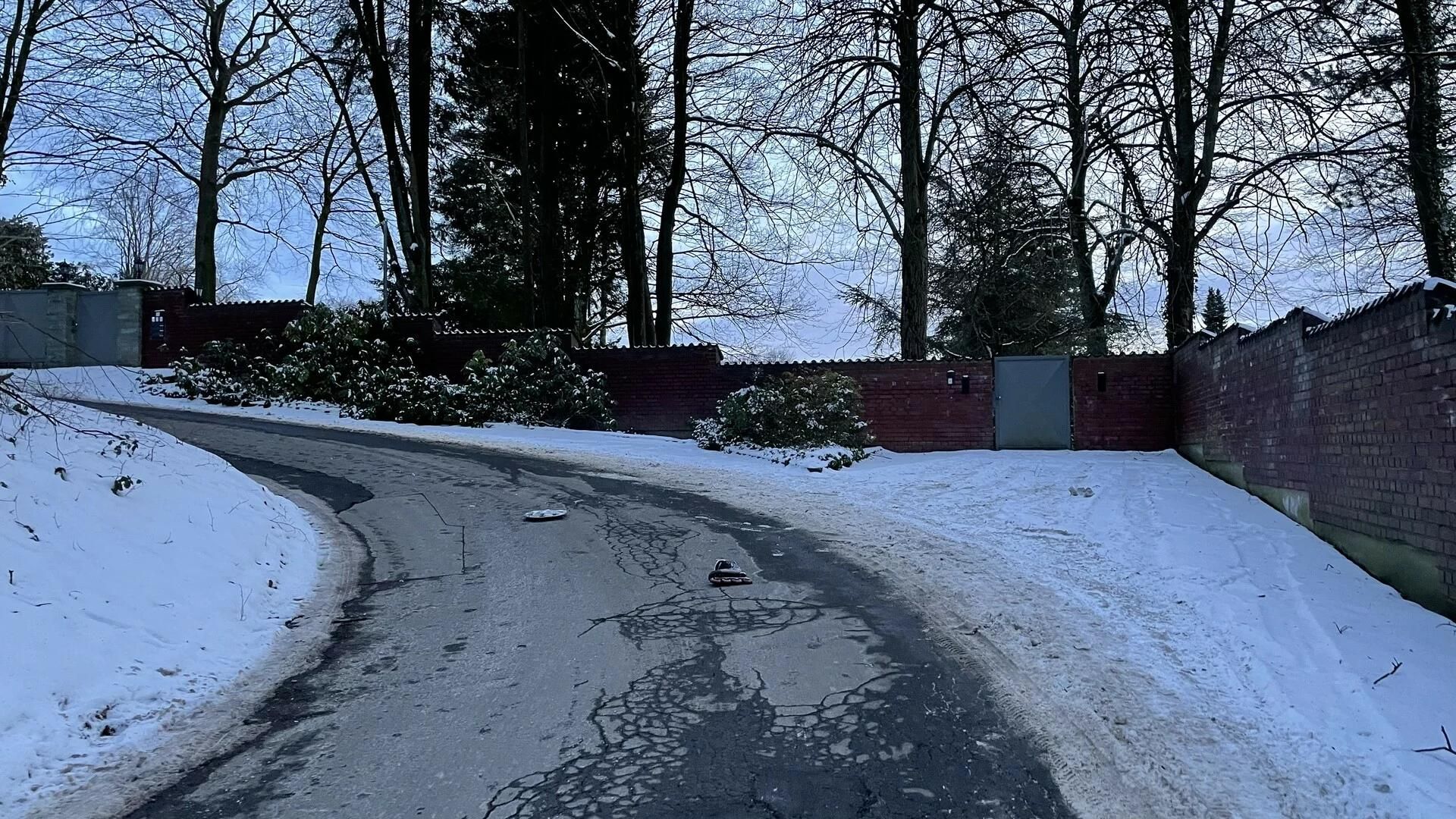}
  \end{subfigure}%
  \begin{subfigure}{0.25\textwidth}
    \includegraphics[width=\linewidth]{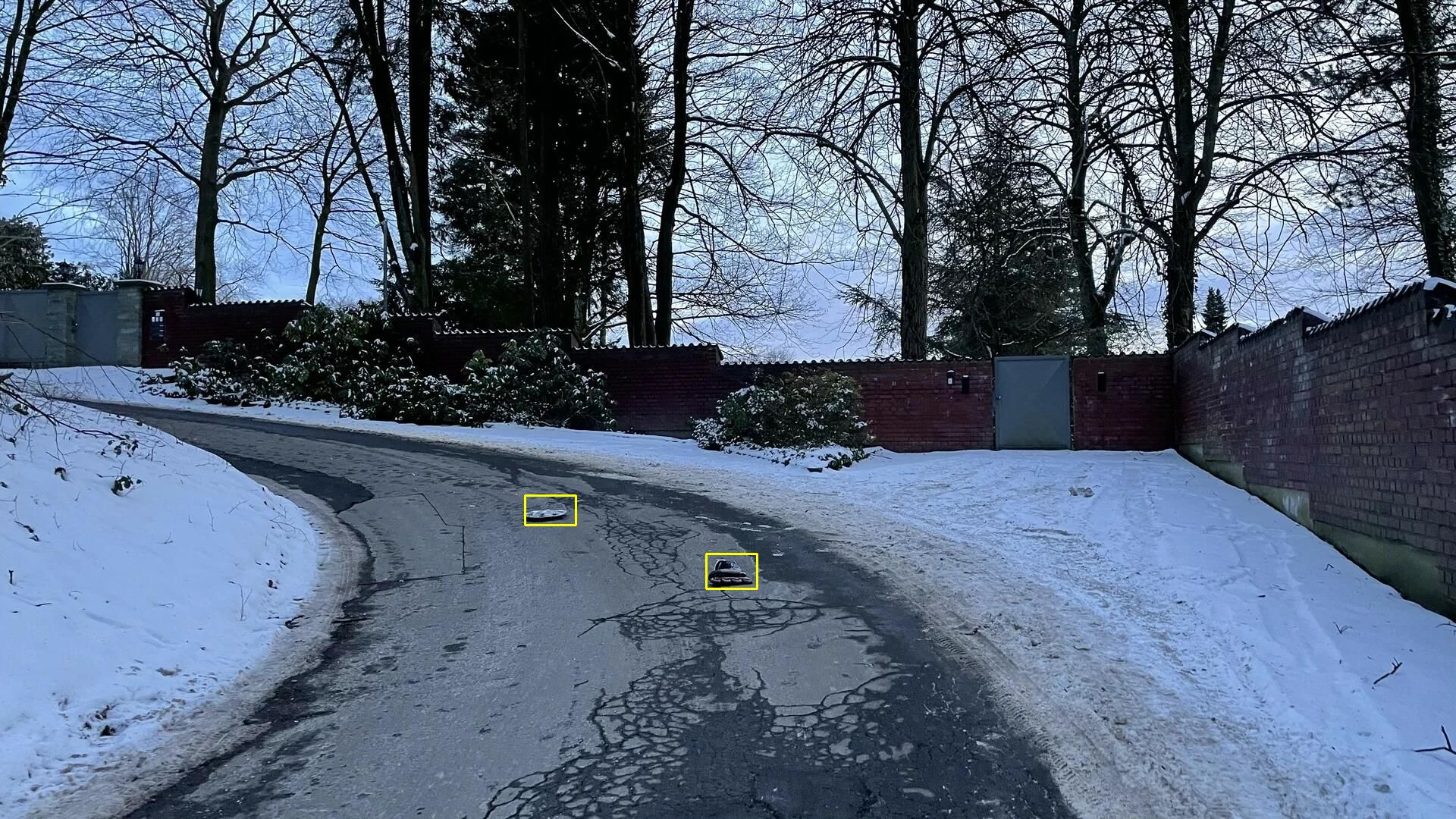}
  \end{subfigure}%
  \begin{subfigure}{0.25\textwidth}
    \includegraphics[width=\linewidth]{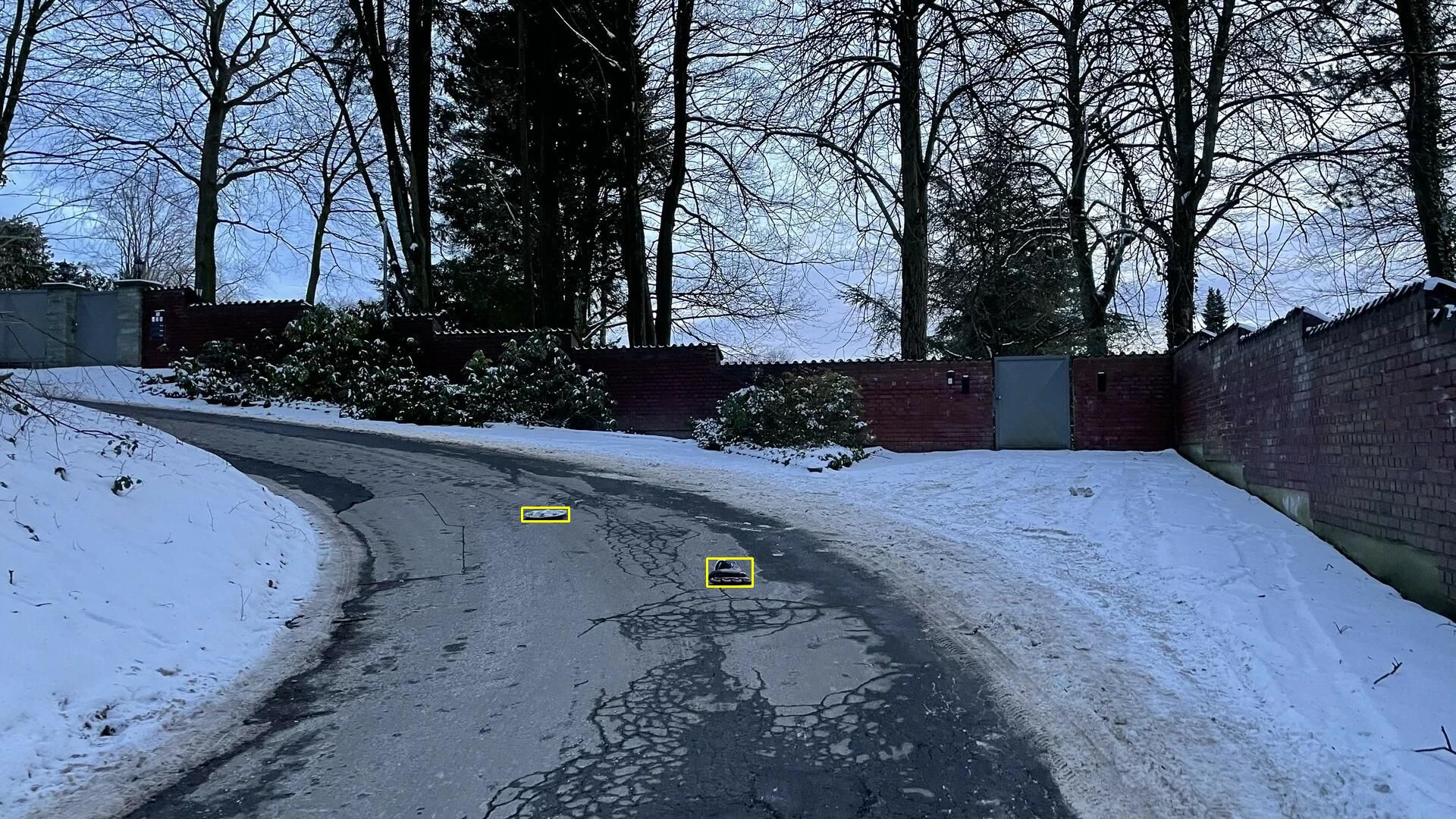}
    
  \end{subfigure}
  \begin{subfigure}{0.25\textwidth}
    \includegraphics[width=\linewidth]{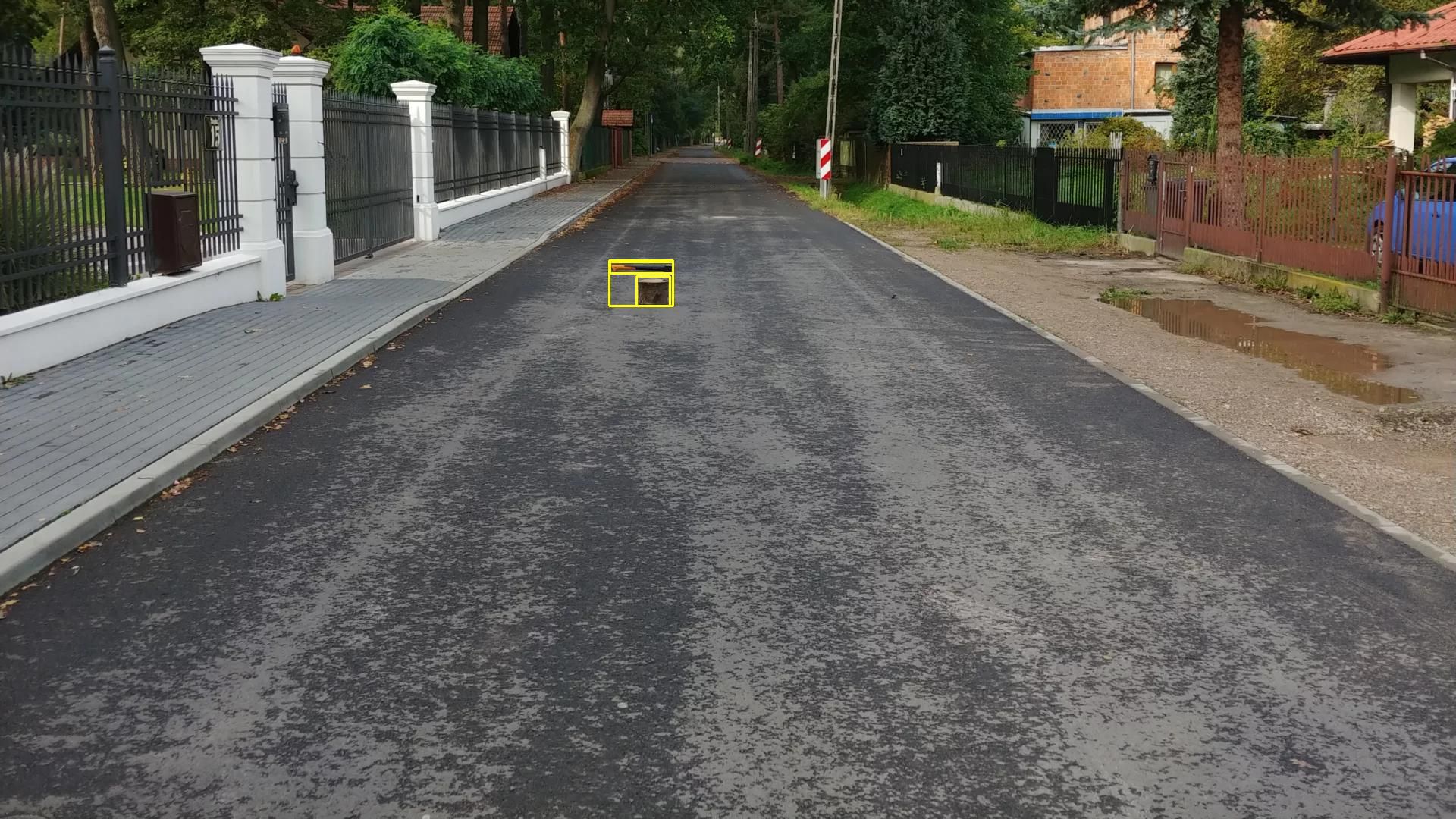}
  \end{subfigure}%
  \begin{subfigure}{0.25\textwidth}
    \includegraphics[width=\linewidth]{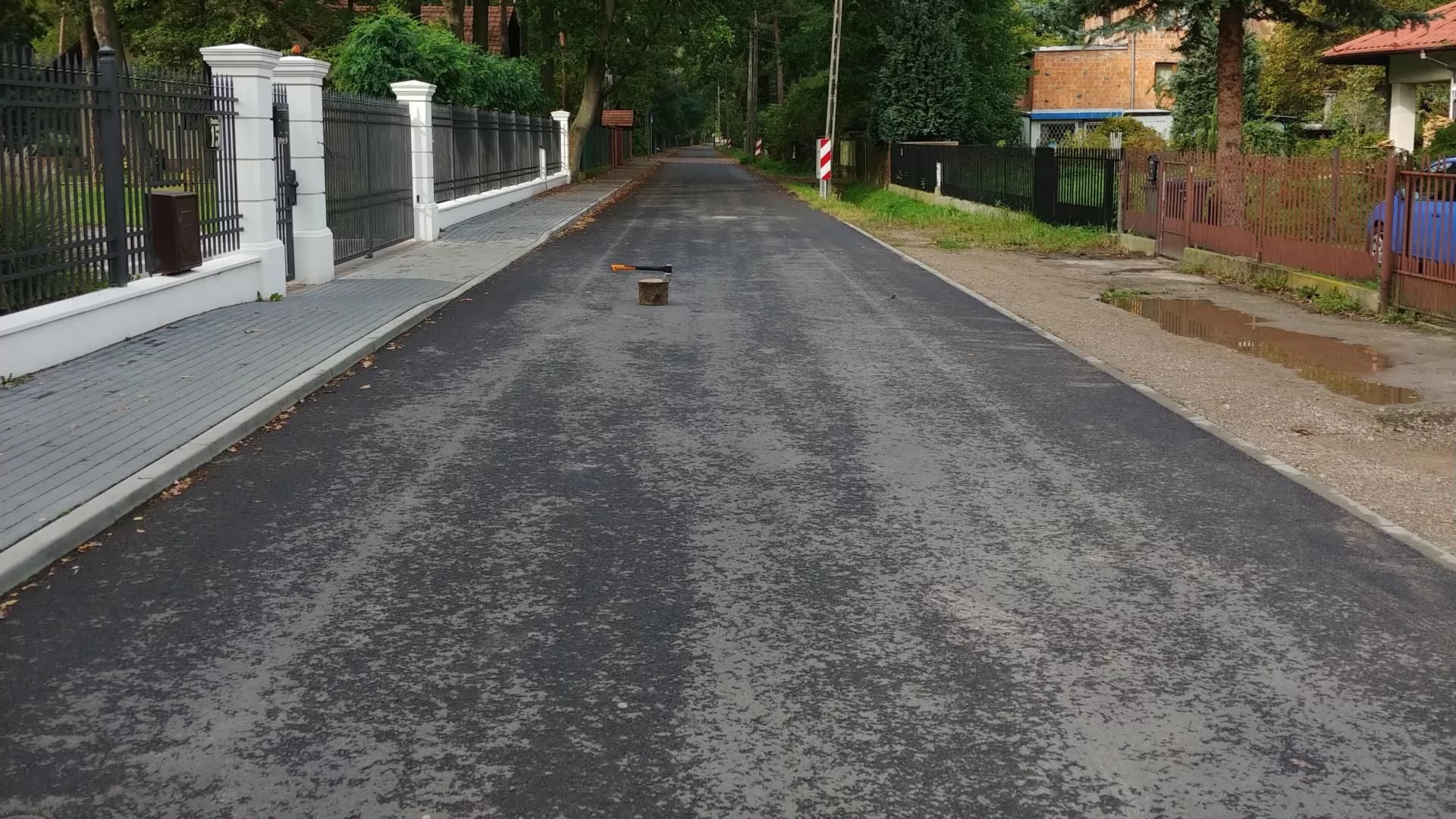}
  \end{subfigure}%
  \begin{subfigure}{0.25\textwidth}
    \includegraphics[width=\linewidth]{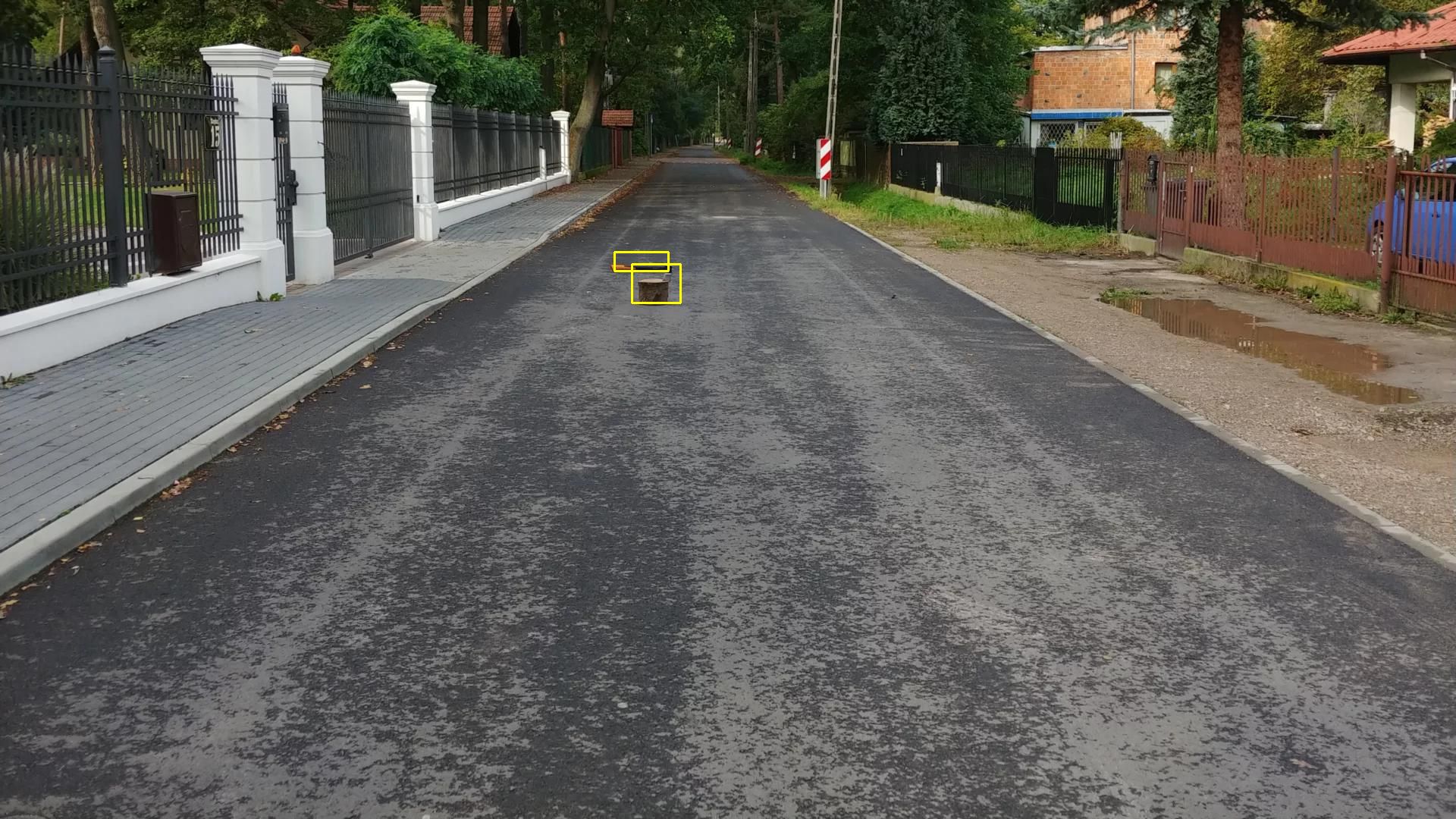}
  \end{subfigure}%
  \begin{subfigure}{0.25\textwidth}
    \includegraphics[width=\linewidth]{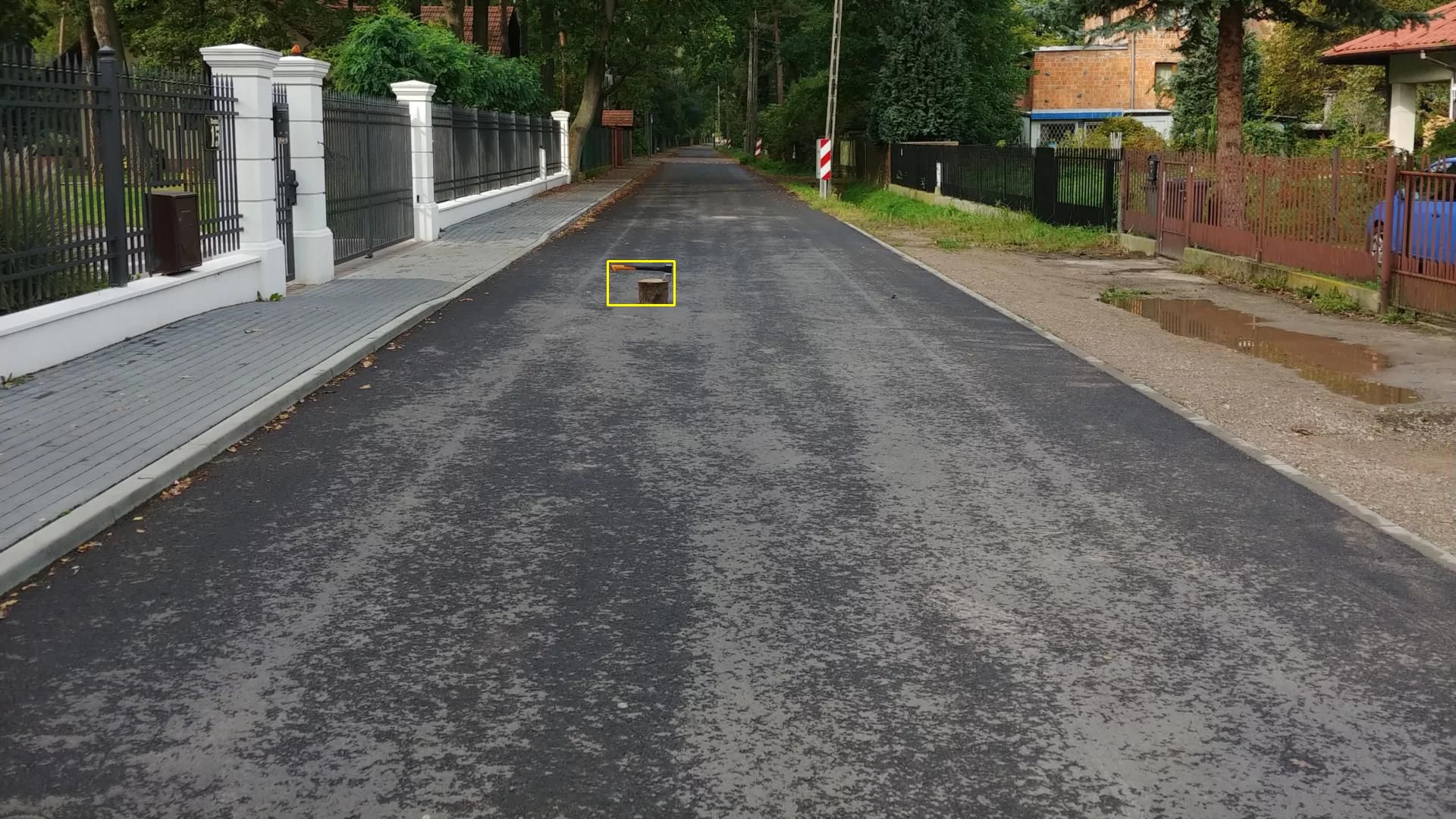}
  \end{subfigure}
  
  \begin{subfigure}{0.25\textwidth}
    \includegraphics[width=\linewidth]{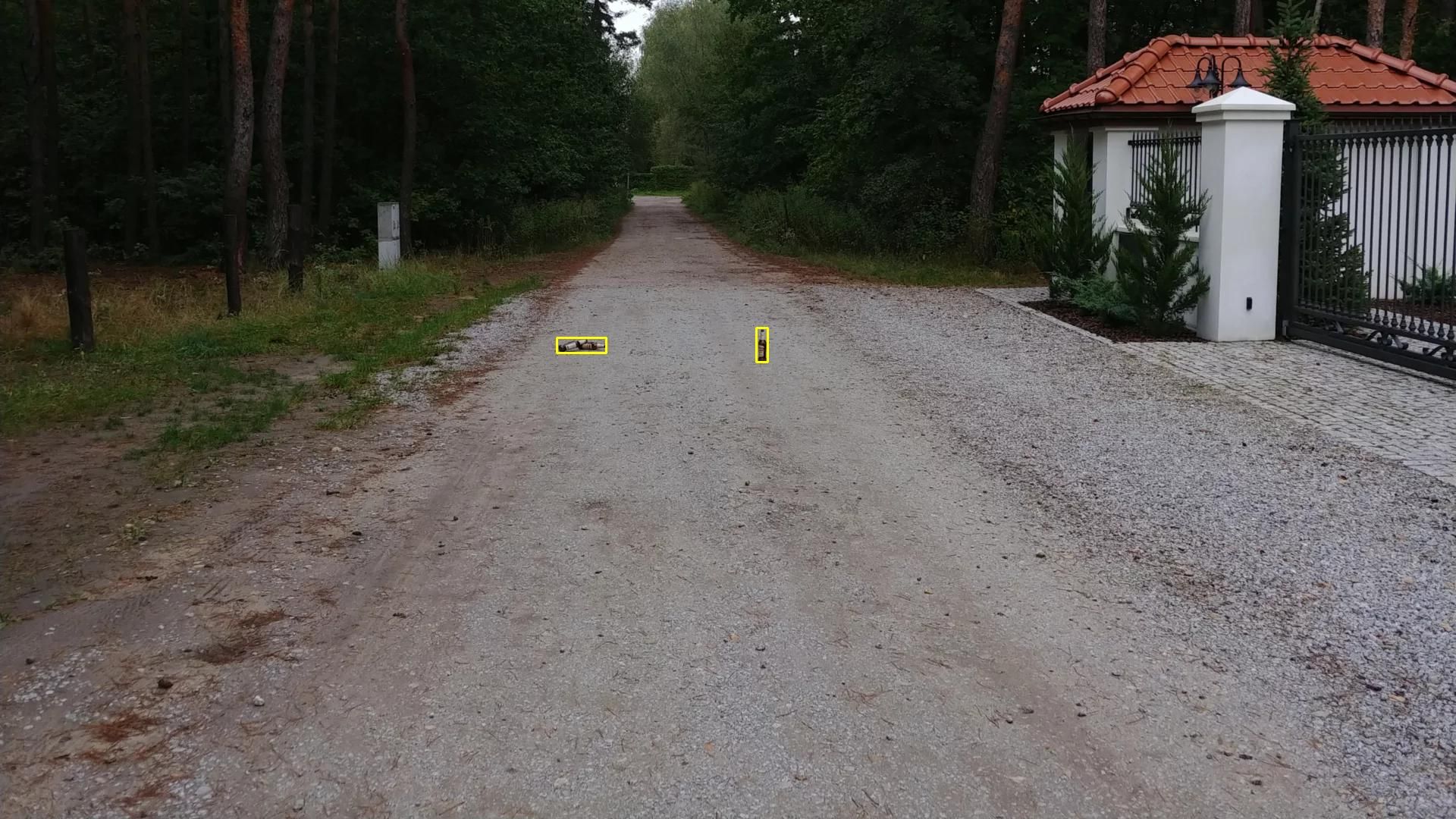}
  \end{subfigure}%
  \begin{subfigure}{0.25\textwidth}
    \includegraphics[width=\linewidth]{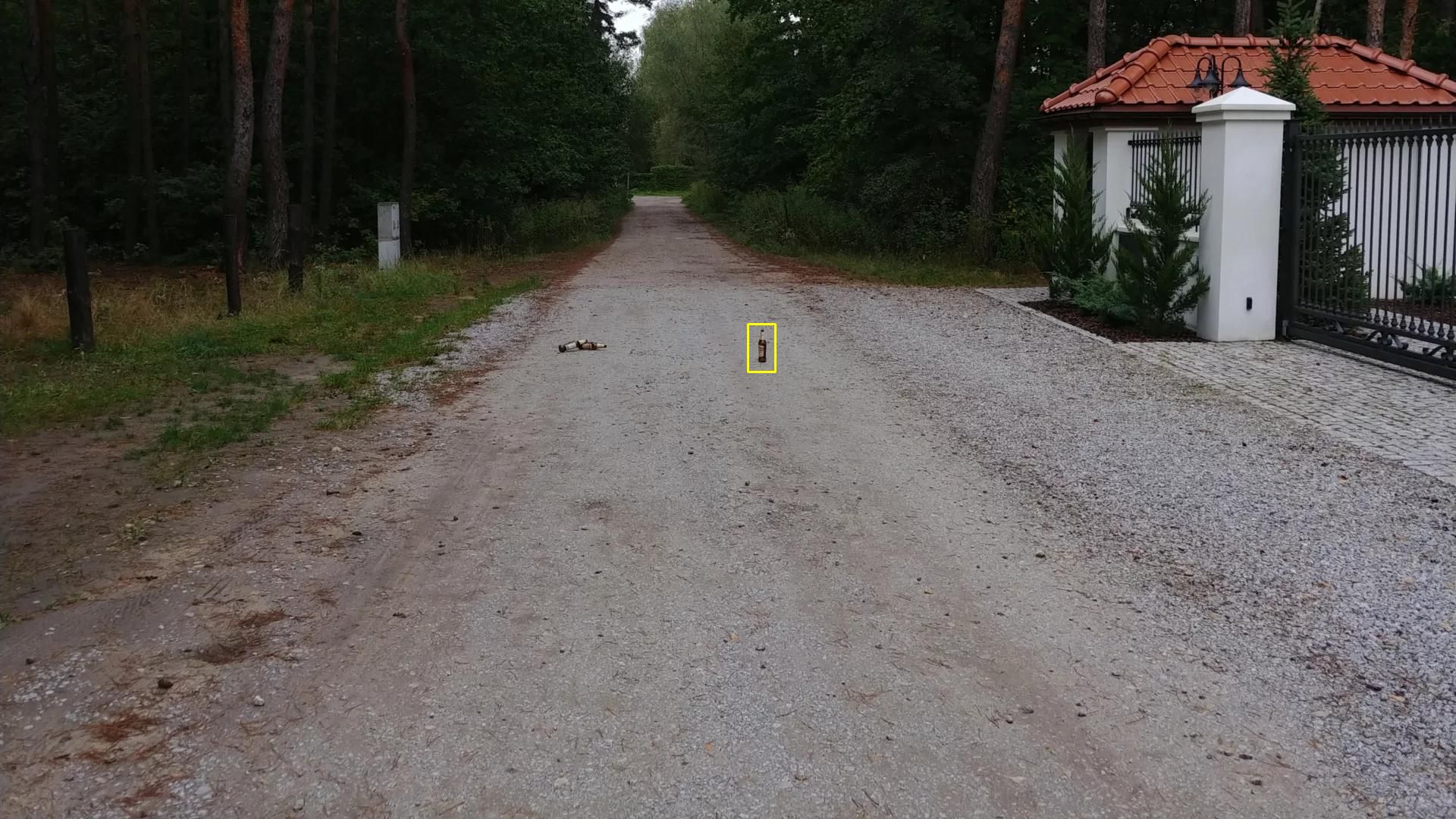}
  \end{subfigure}%
  \begin{subfigure}{0.25\textwidth}
    \includegraphics[width=\linewidth]{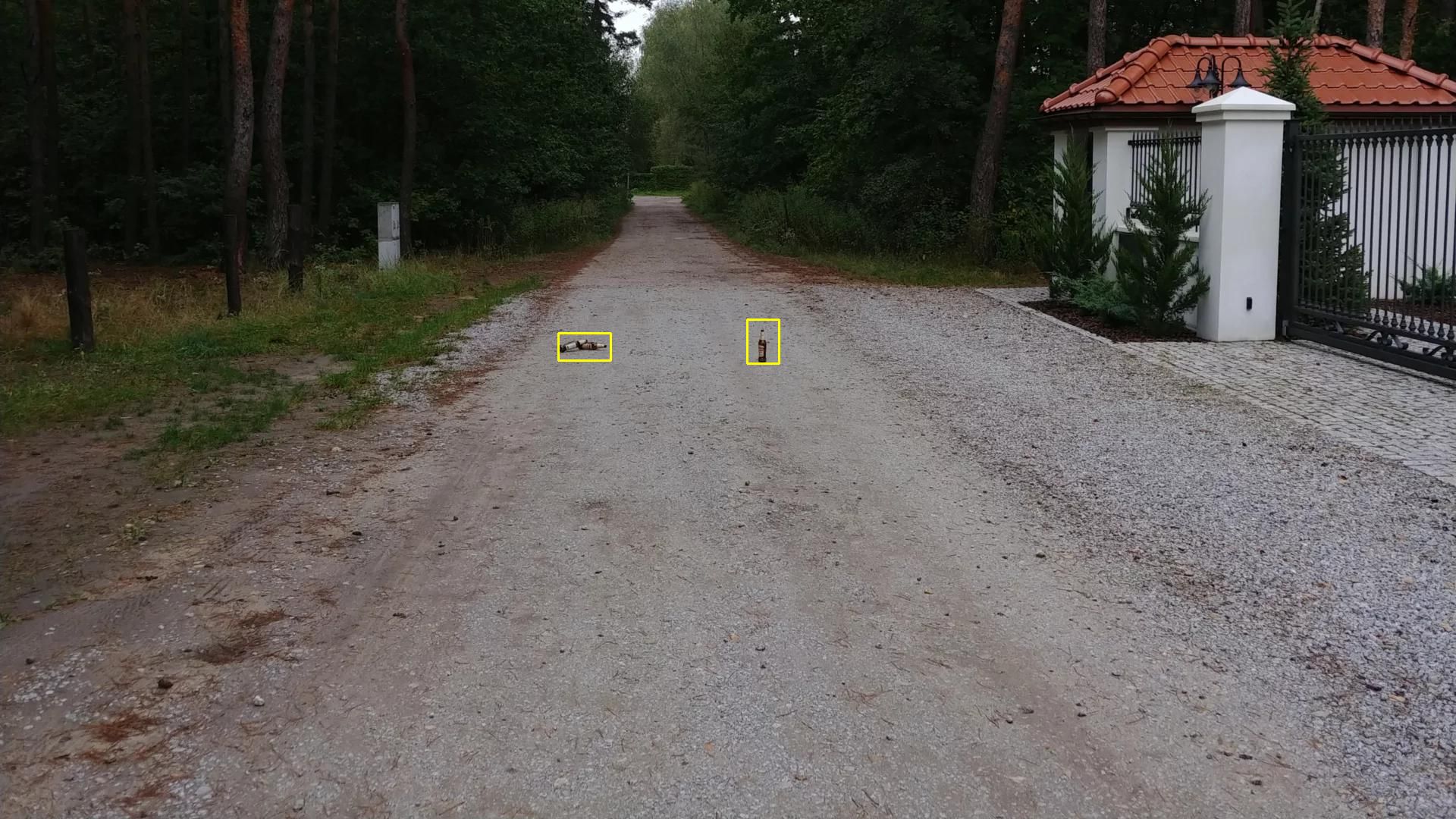}
  \end{subfigure}%
  \begin{subfigure}{0.25\textwidth}
    \includegraphics[width=\linewidth]{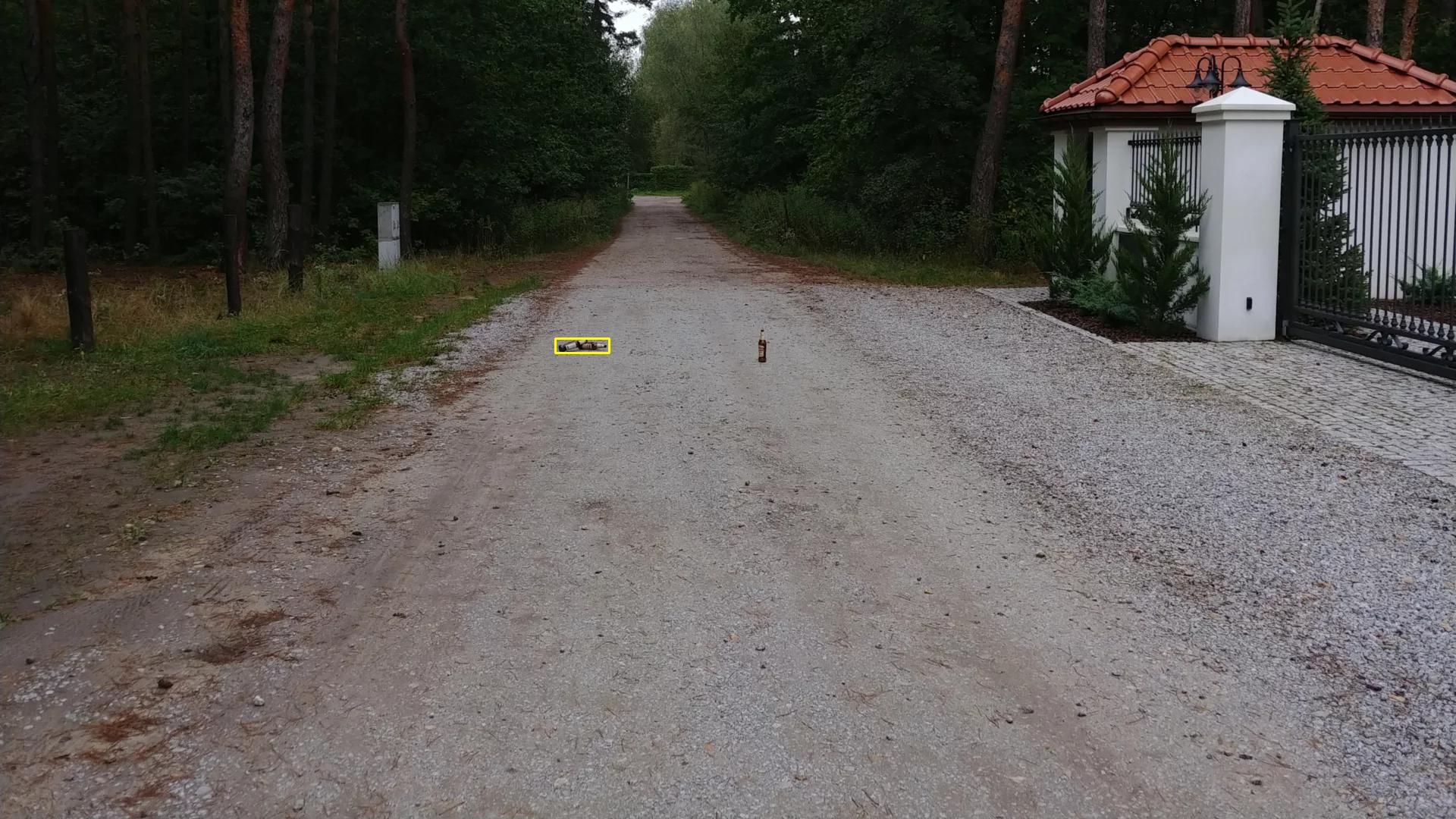}
  \end{subfigure}

  \begin{subfigure}{0.25\textwidth}
    \includegraphics[width=\linewidth]{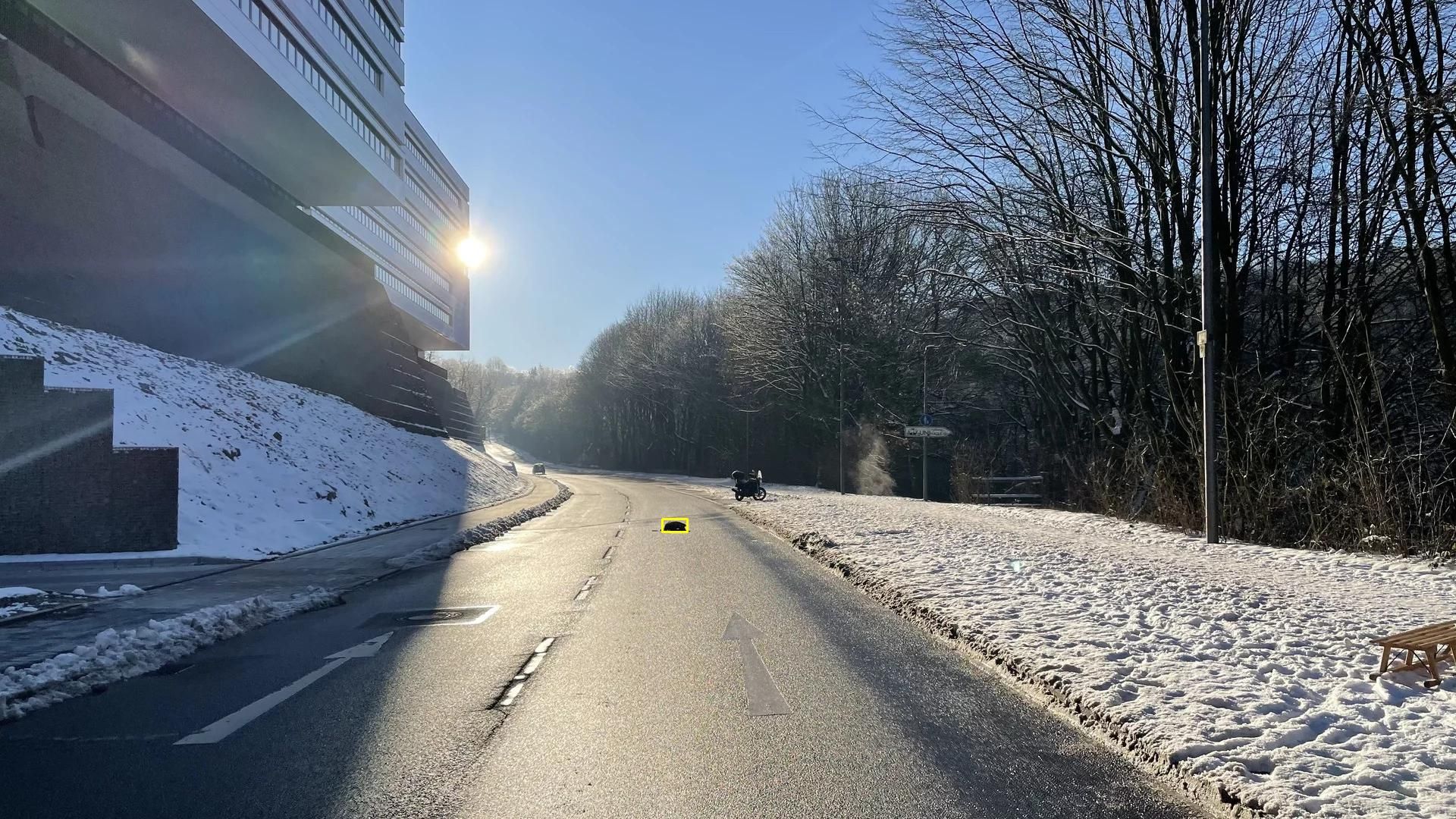}
  \end{subfigure}%
  \begin{subfigure}{0.25\textwidth}
    \includegraphics[width=\linewidth]{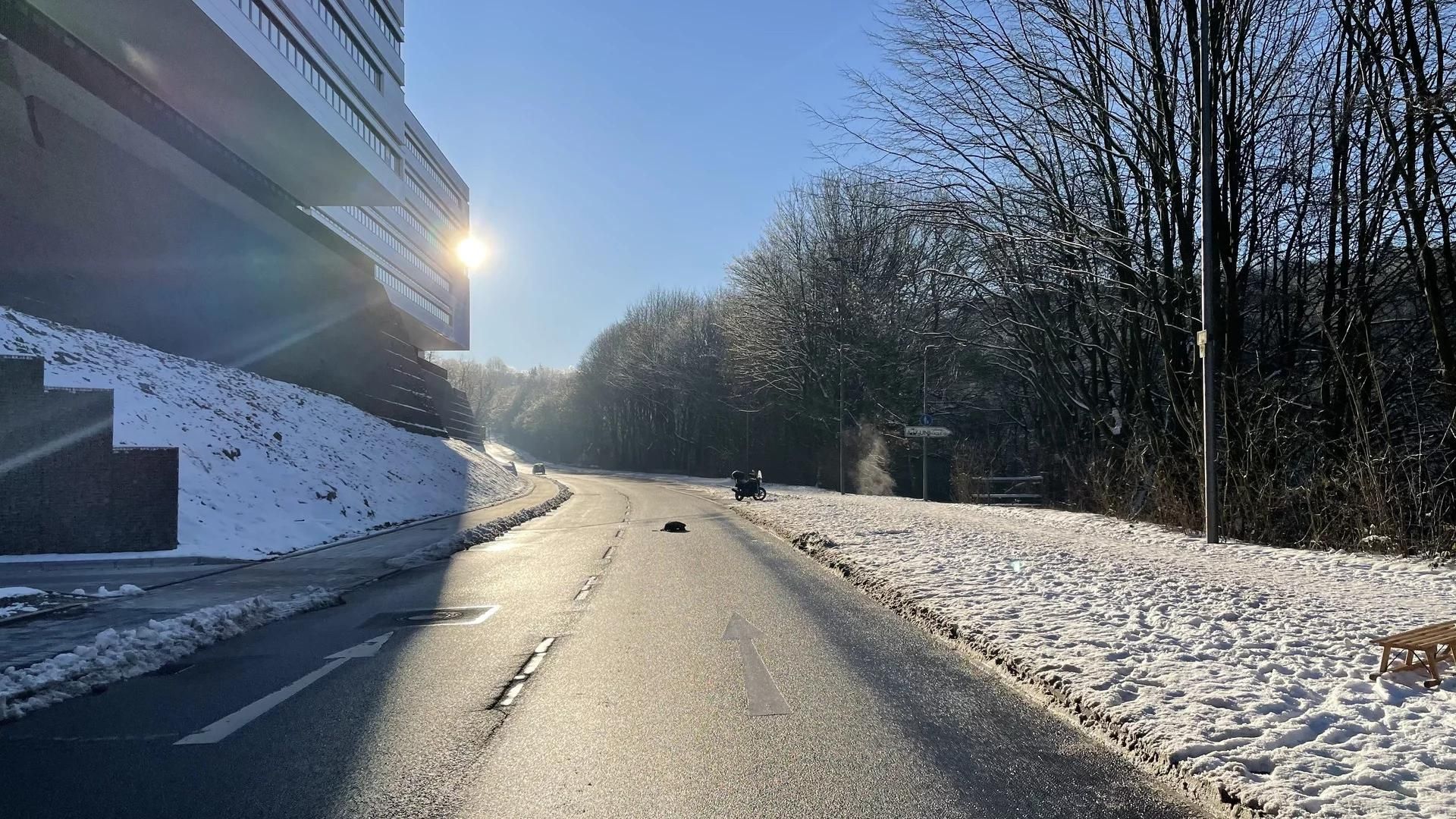}
  \end{subfigure}%
  \begin{subfigure}{0.25\textwidth}
    \includegraphics[width=\linewidth]{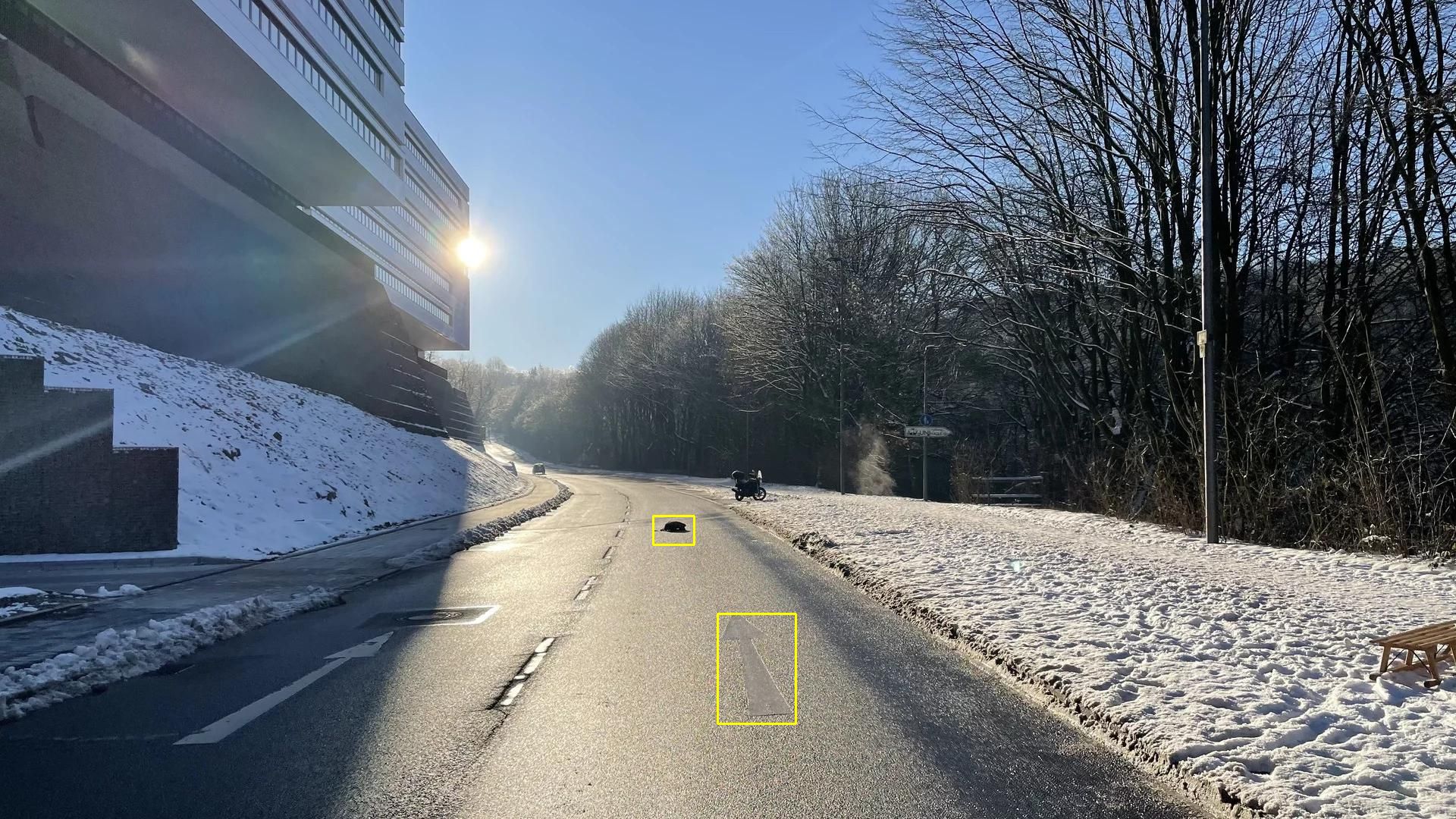}
  \end{subfigure}%
  \begin{subfigure}{0.25\textwidth}
    \includegraphics[width=\linewidth]{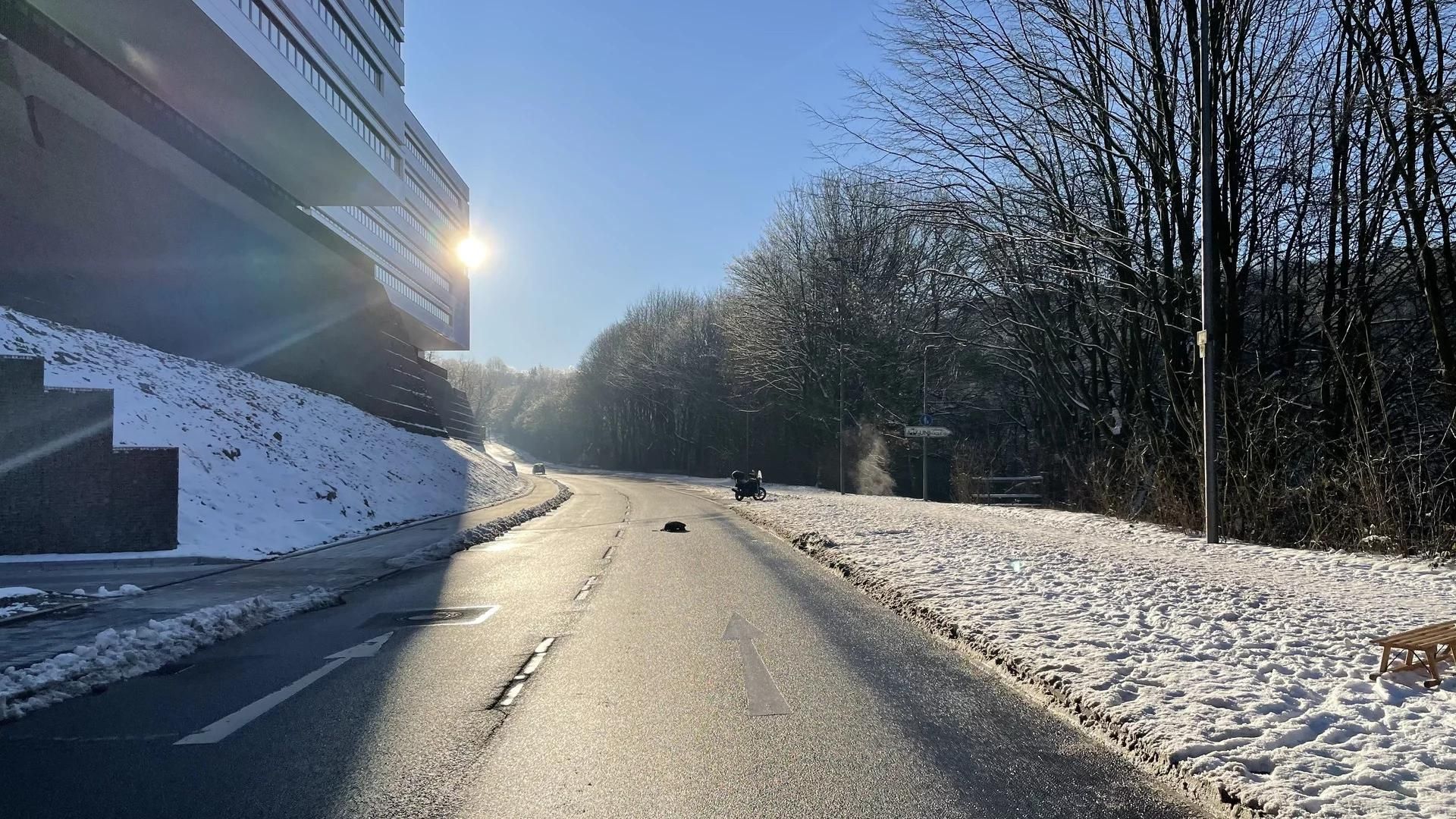}
  \end{subfigure}

    \begin{subfigure}{0.25\textwidth}
    \includegraphics[width=\linewidth]{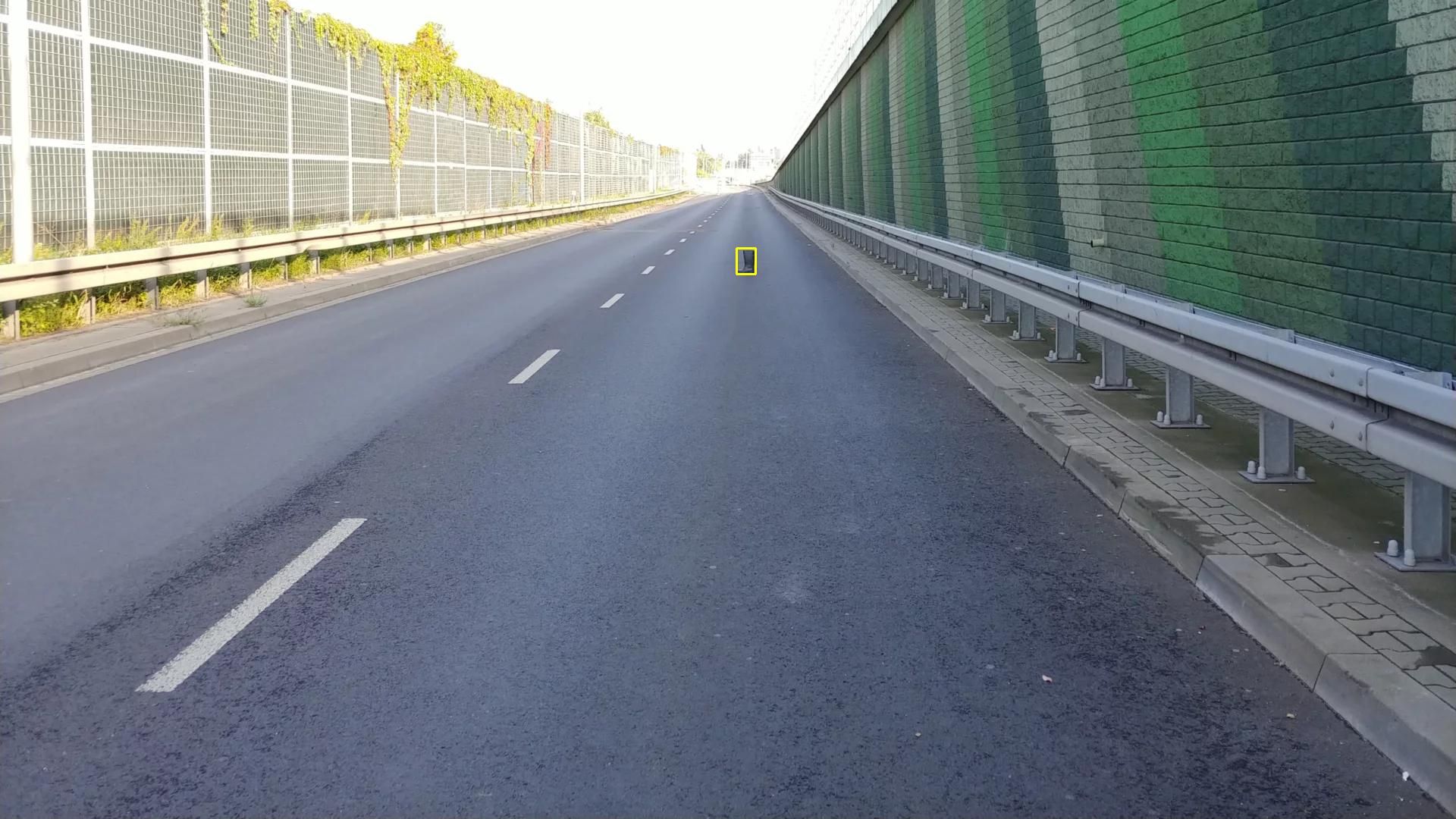}
    \caption{Grounding DINO}
  \end{subfigure}%
  \begin{subfigure}{0.25\textwidth}
    \includegraphics[width=\linewidth]{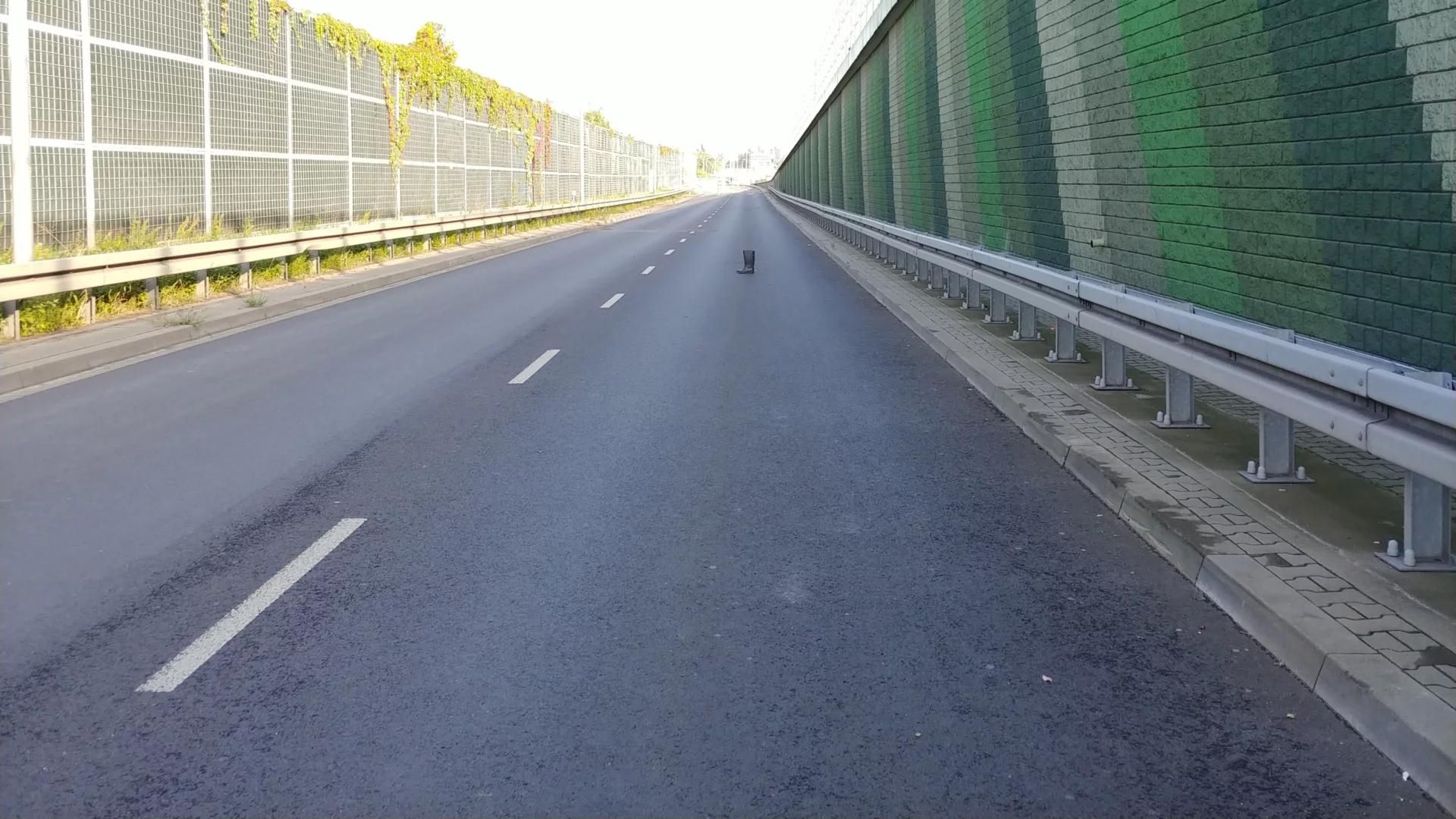}
    \caption{YOLO-World}
  \end{subfigure}%
  \begin{subfigure}{0.25\textwidth}
    \includegraphics[width=\linewidth]{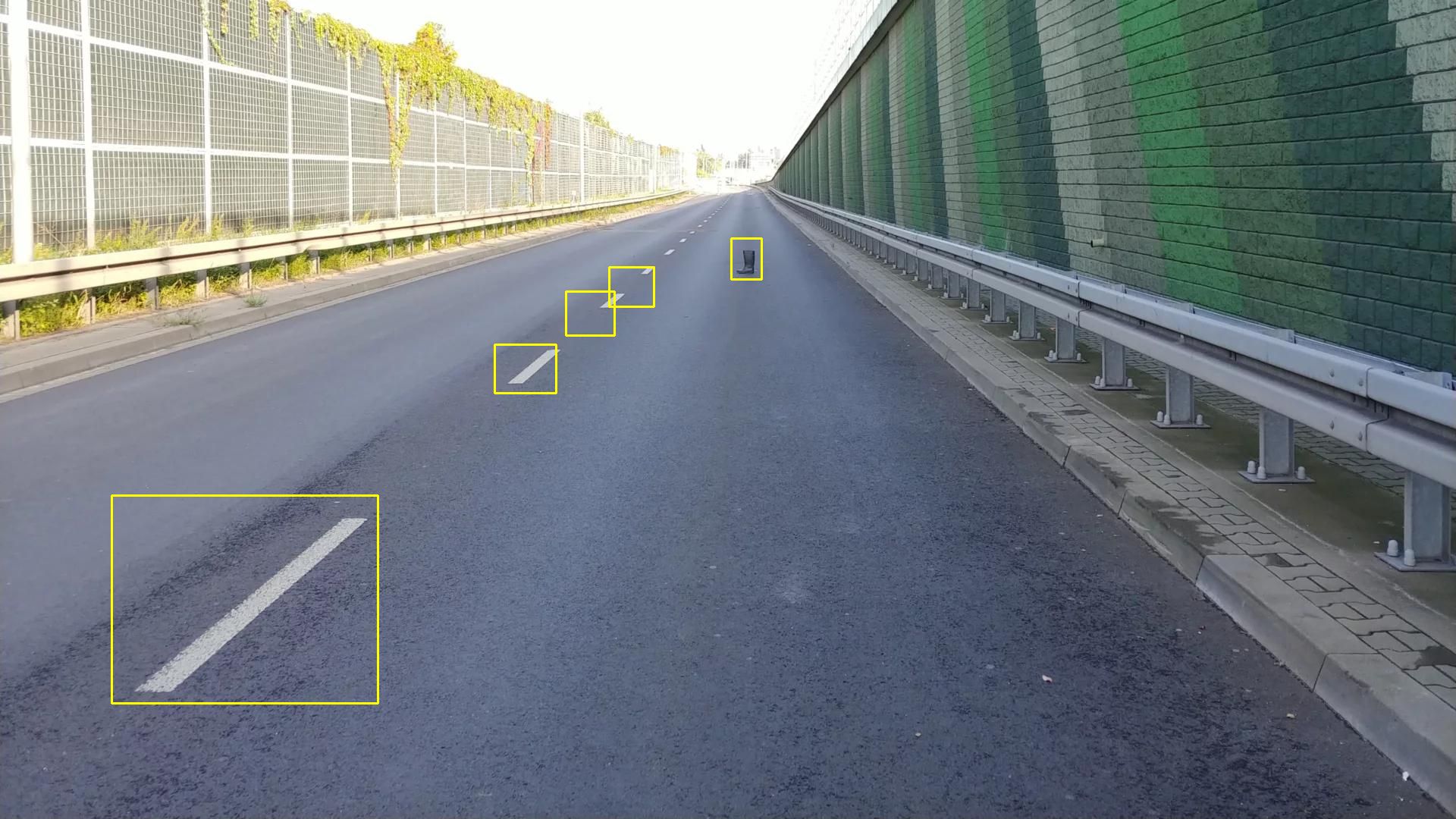}
    \caption{MDETR}
  \end{subfigure}%
  \begin{subfigure}{0.25\textwidth}
    \includegraphics[width=\linewidth]{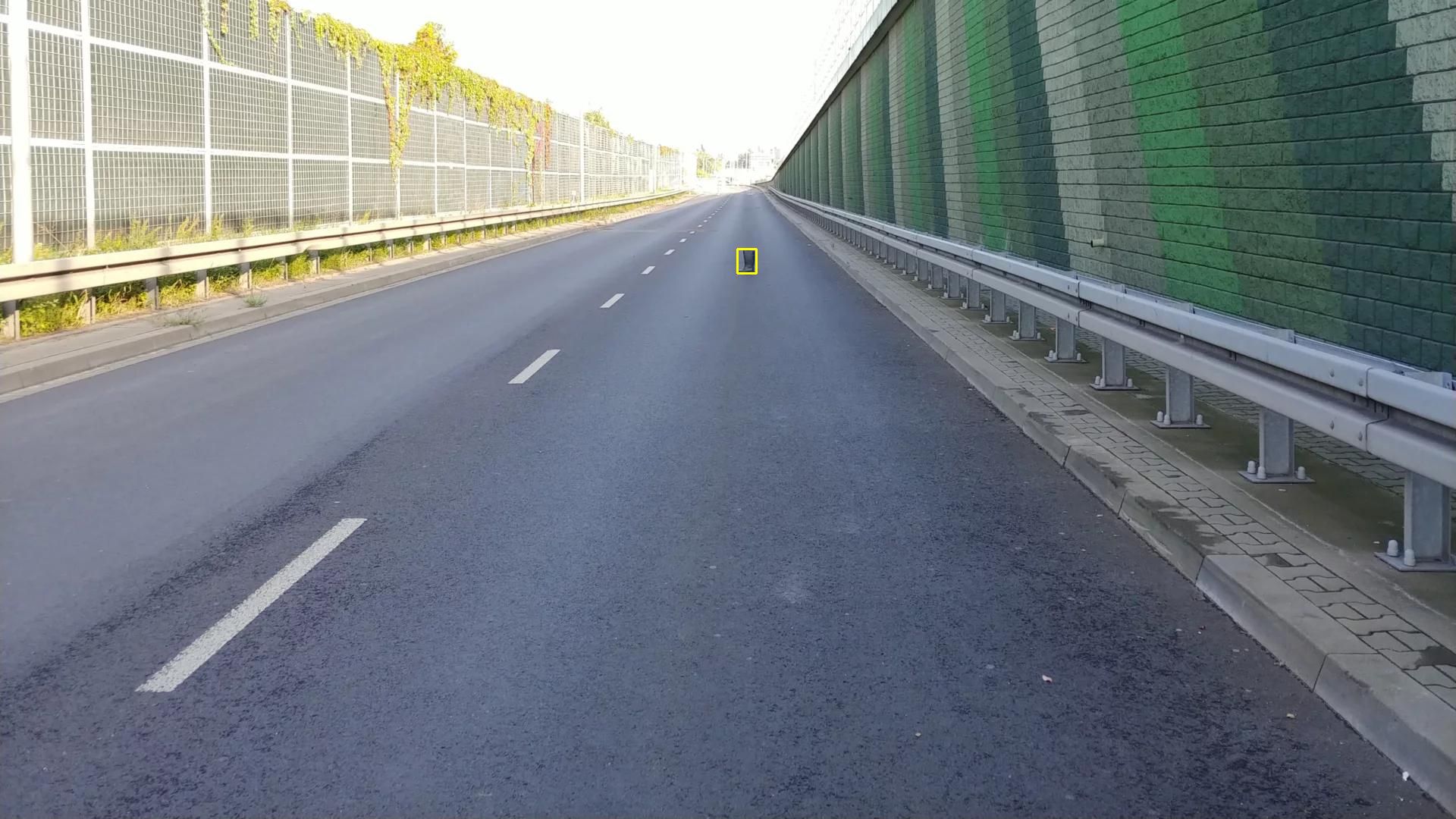}
    \caption{OmDet}
  \end{subfigure}

    \caption{Qualitative results on the RoadObstacle21 dataset. Each image featuring different weather and road condition with OOD objects of differing sizes and distances.}
    \label{fig:roadobstacle}
\end{figure}

In this section we provide qualitative results on RoadObstacle21, RoadAnomaly21 and LostAndFound using open-vocabulary models. It can be observed in Figure \ref{fig:roadobstacle} that Grounding DINO accurately detects the OOD objects, whereas YOLO-World mostly misses them. Conversely, OmDet performs best in detecting the OOD object in the second row's image, where both Grounding DINO and MDETR fails to detect the object as a single entity. Additionally, MDETR produces several false positives.

\begin{figure}[!h]
\centering
  \captionsetup{justification=centering,margin=0.5cm}
  \begin{subfigure}{0.25\textwidth}
    \includegraphics[width=\linewidth]{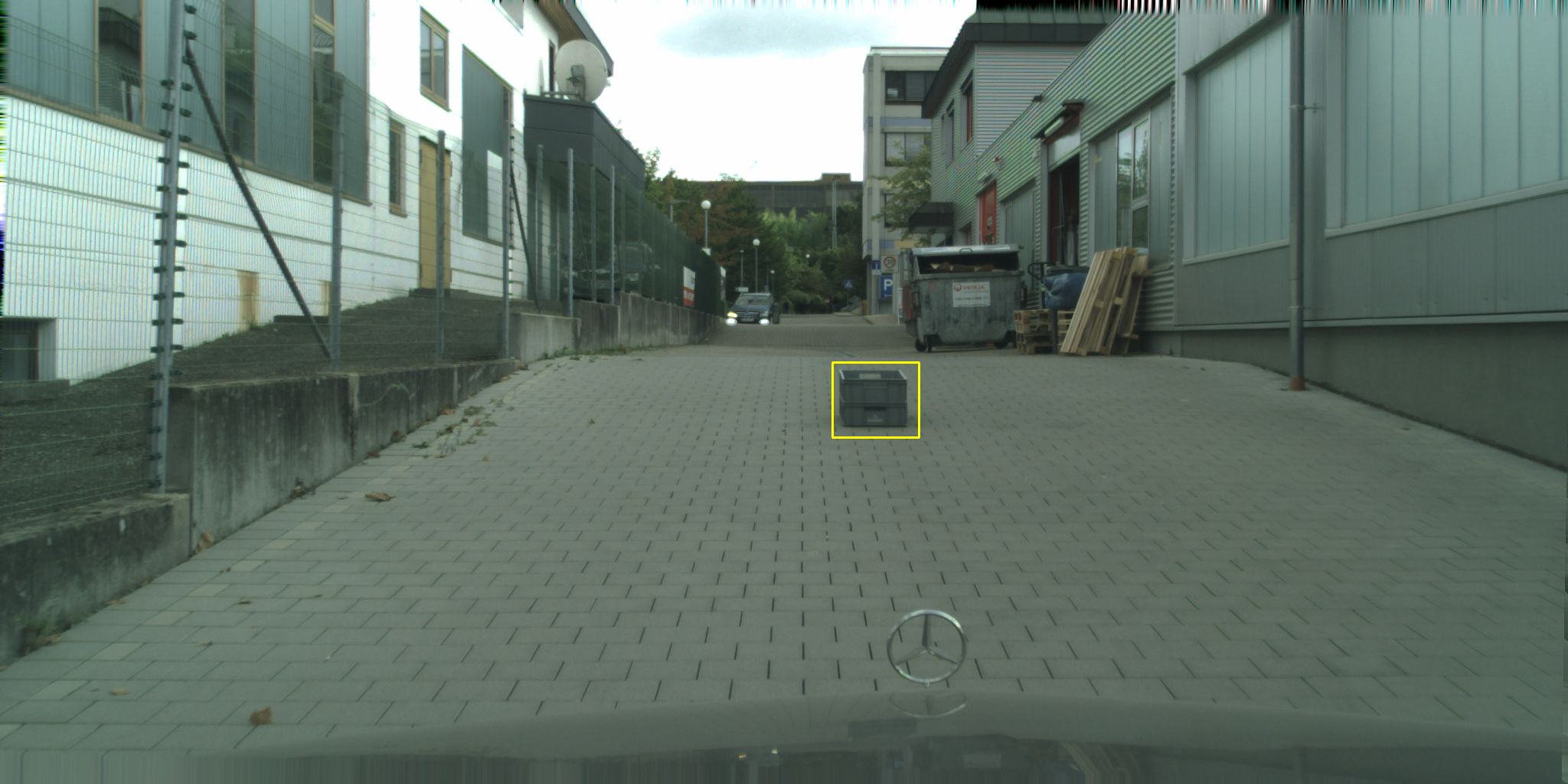}
  \end{subfigure}%
  \begin{subfigure}{0.25\textwidth}
    \includegraphics[width=\linewidth]{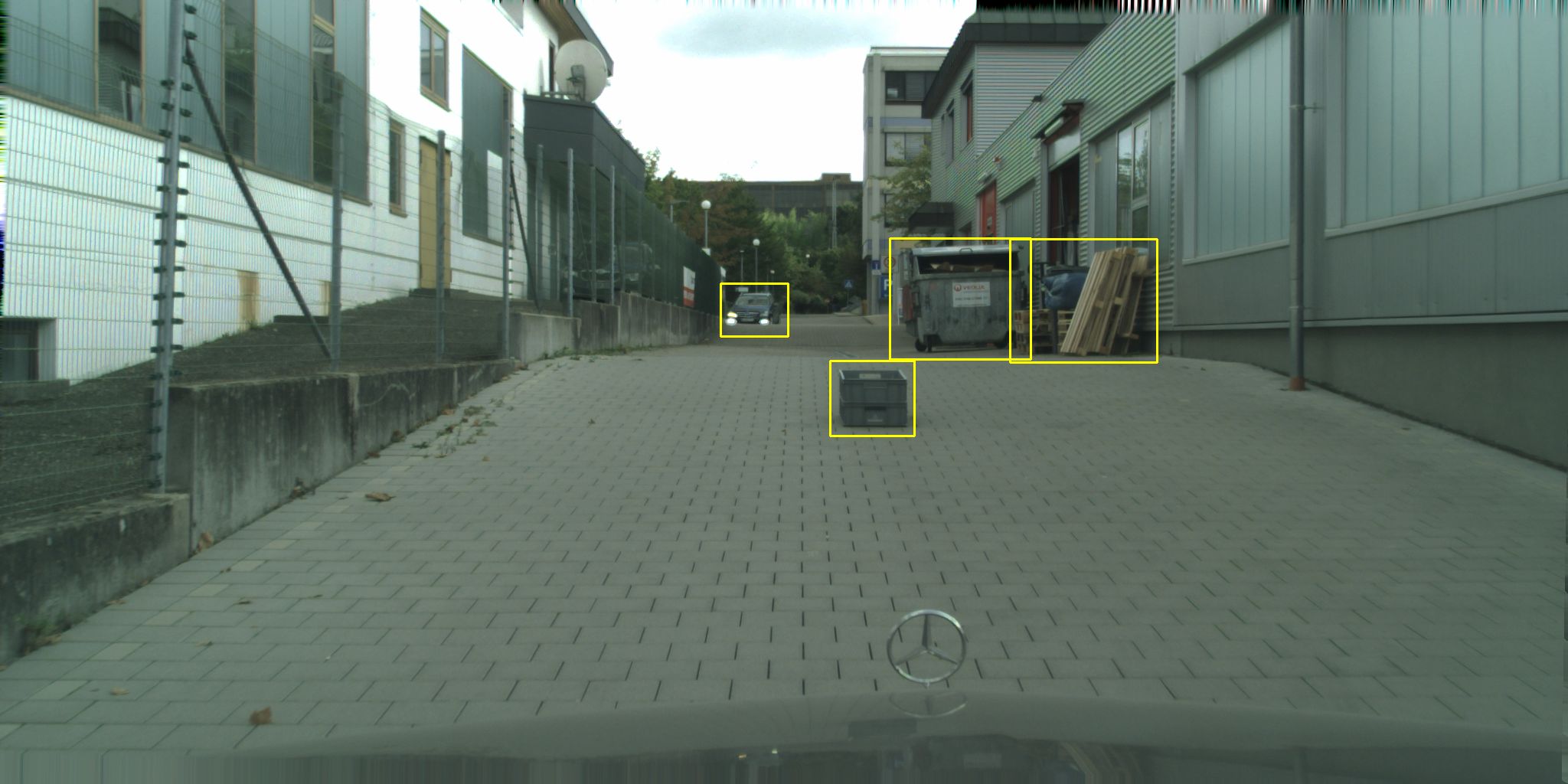}
  \end{subfigure}%
  \begin{subfigure}{0.25\textwidth}
    \includegraphics[width=\linewidth]{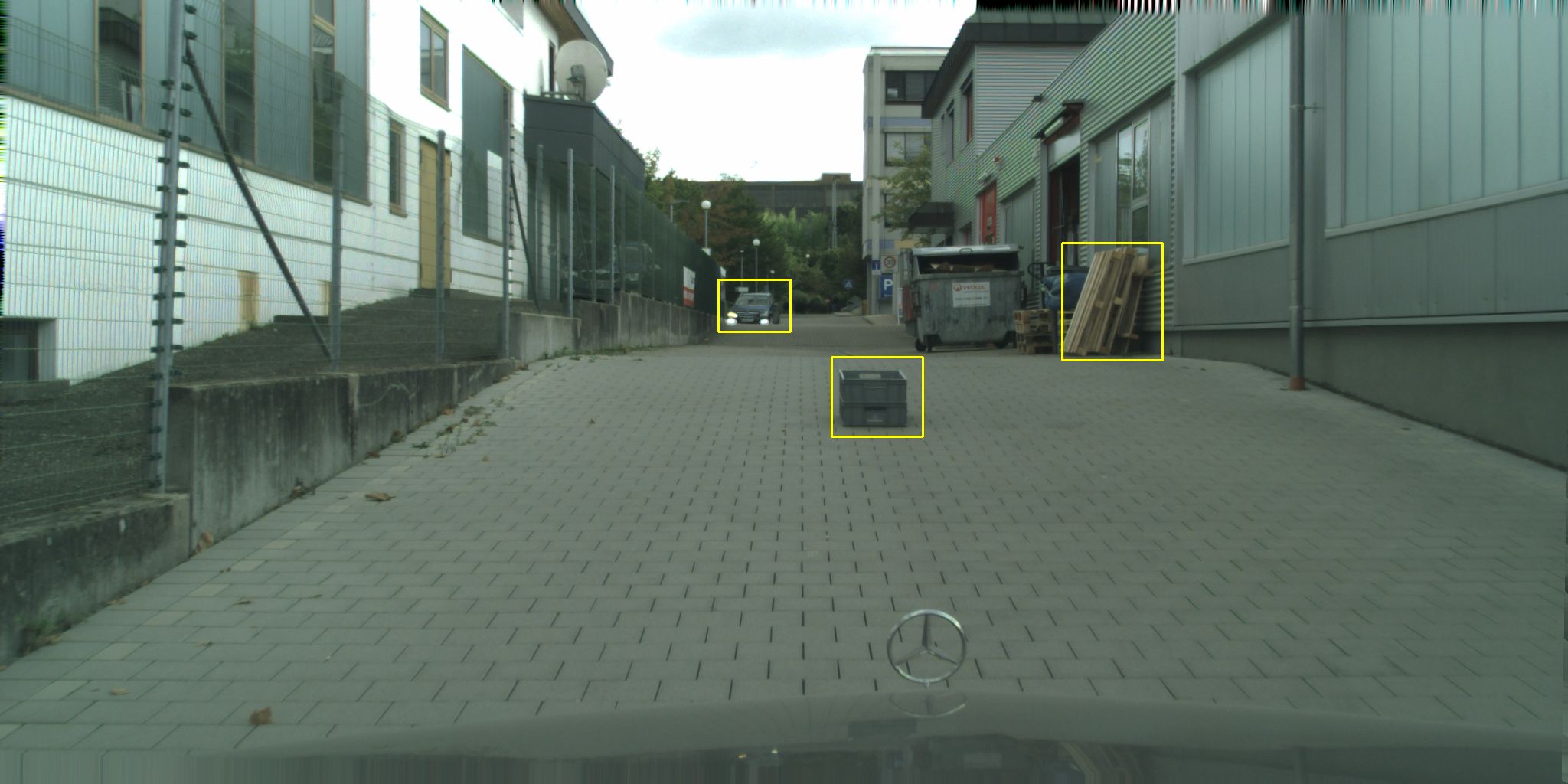}
  \end{subfigure}%
  \begin{subfigure}{0.25\textwidth}
    \includegraphics[width=\linewidth]{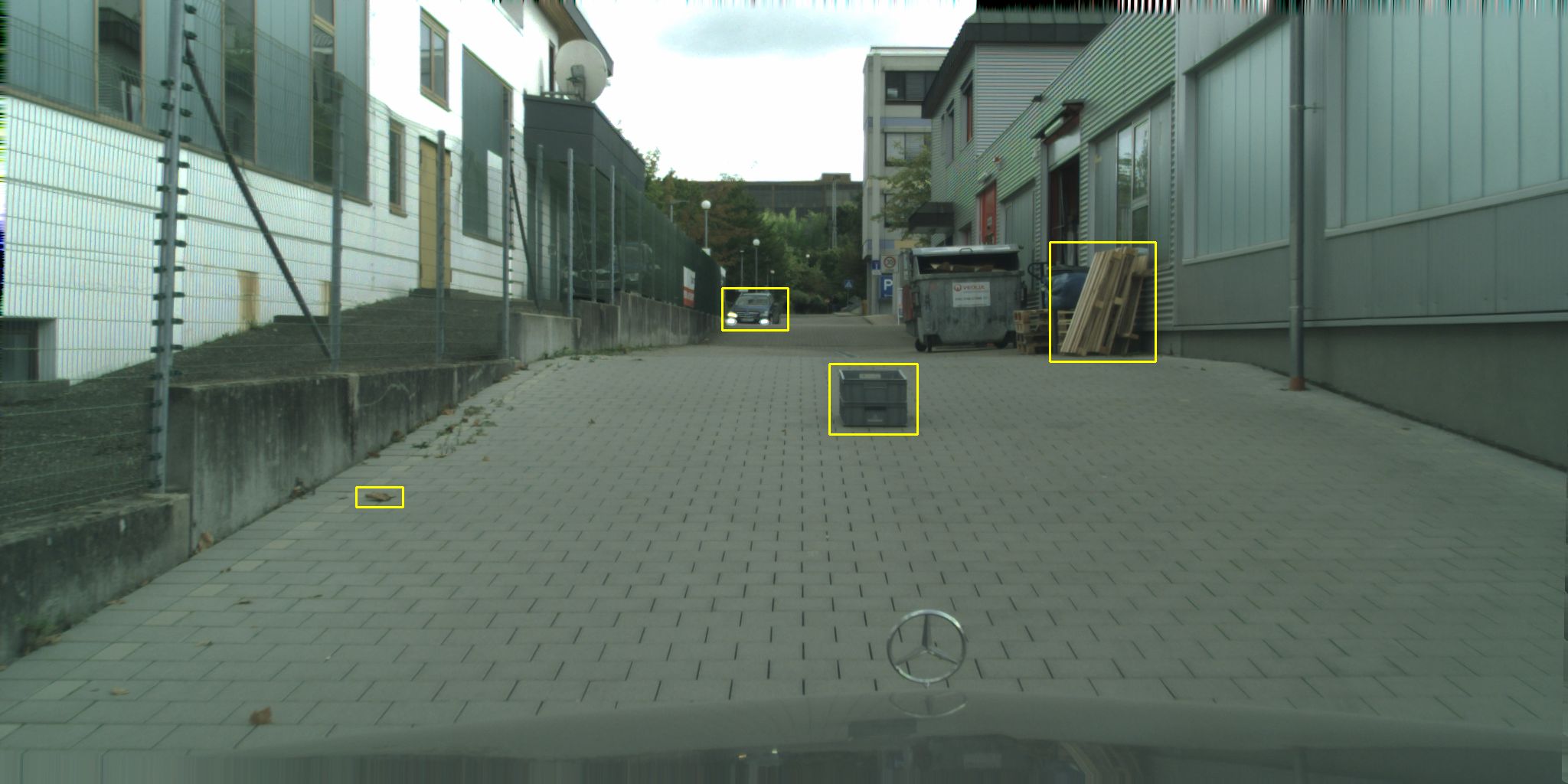}
  \end{subfigure}
  
\begin{subfigure}{0.25\textwidth} 
    \includegraphics[width=\linewidth]{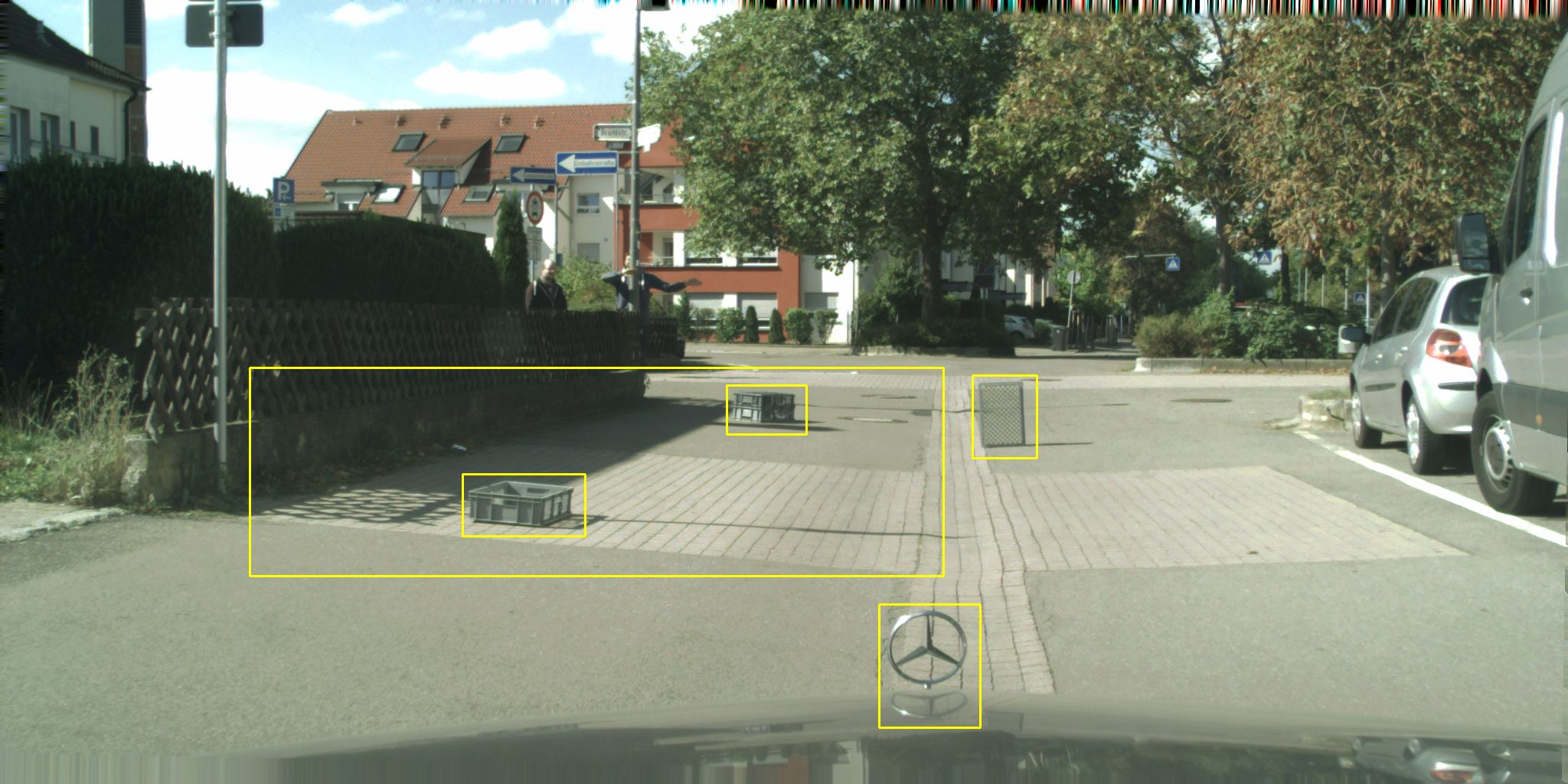}
  \end{subfigure}%
  \begin{subfigure}{0.25\textwidth}
    \includegraphics[width=\linewidth]{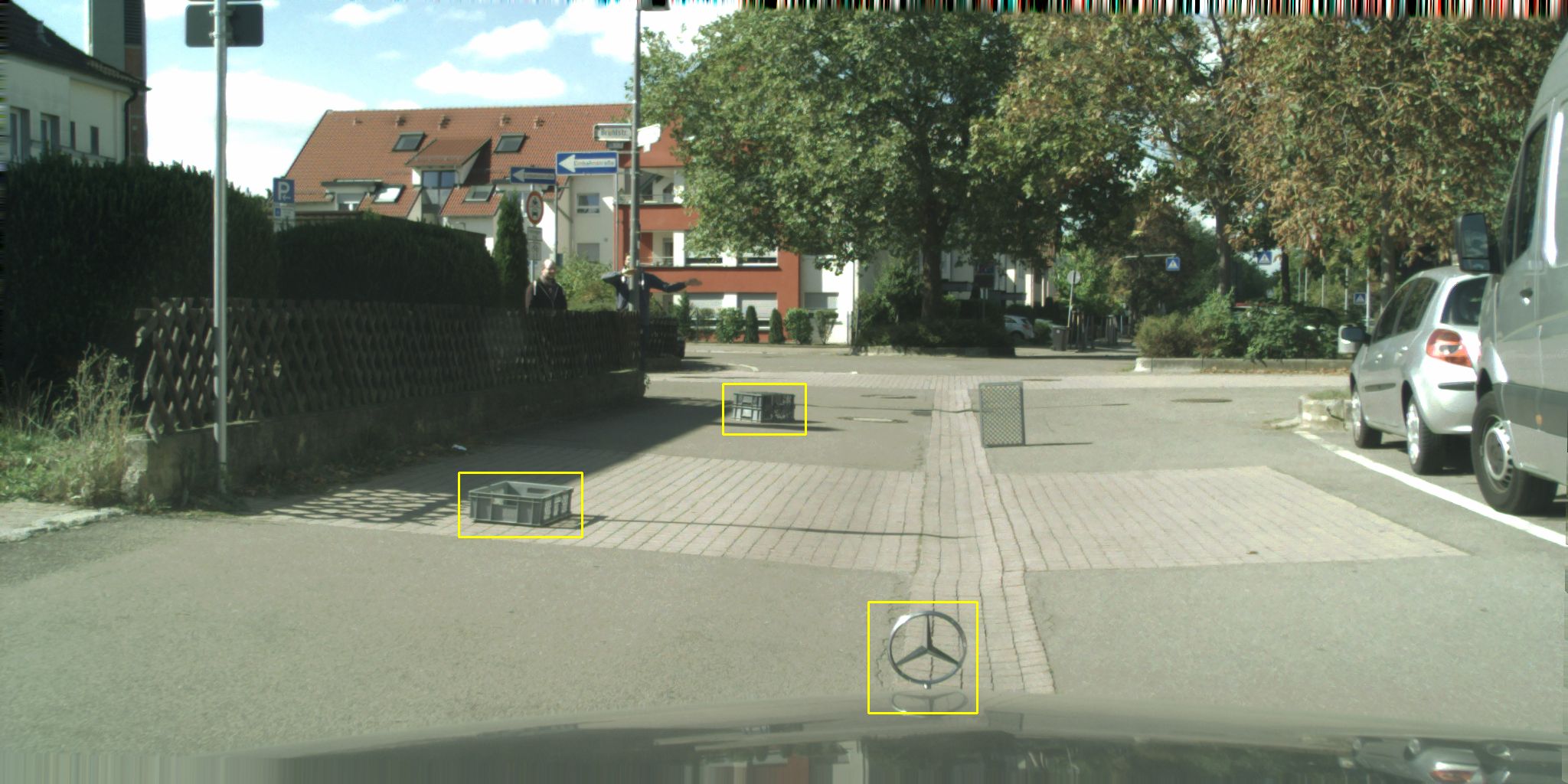}
  \end{subfigure}%
  \begin{subfigure}{0.25\textwidth}
    \includegraphics[width=\linewidth]{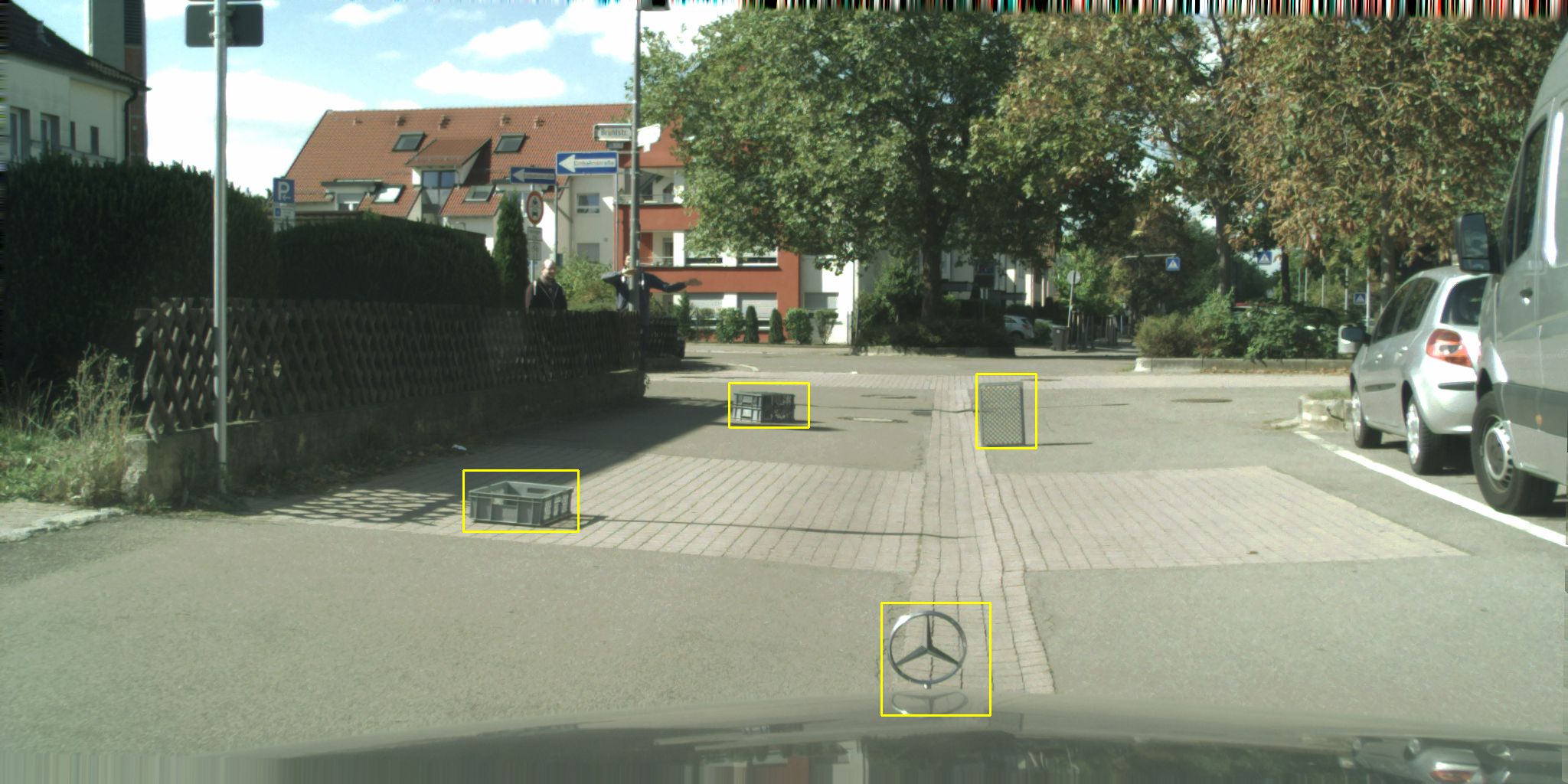}
  \end{subfigure}%
  \begin{subfigure}{0.25\textwidth}
    \includegraphics[width=\linewidth]{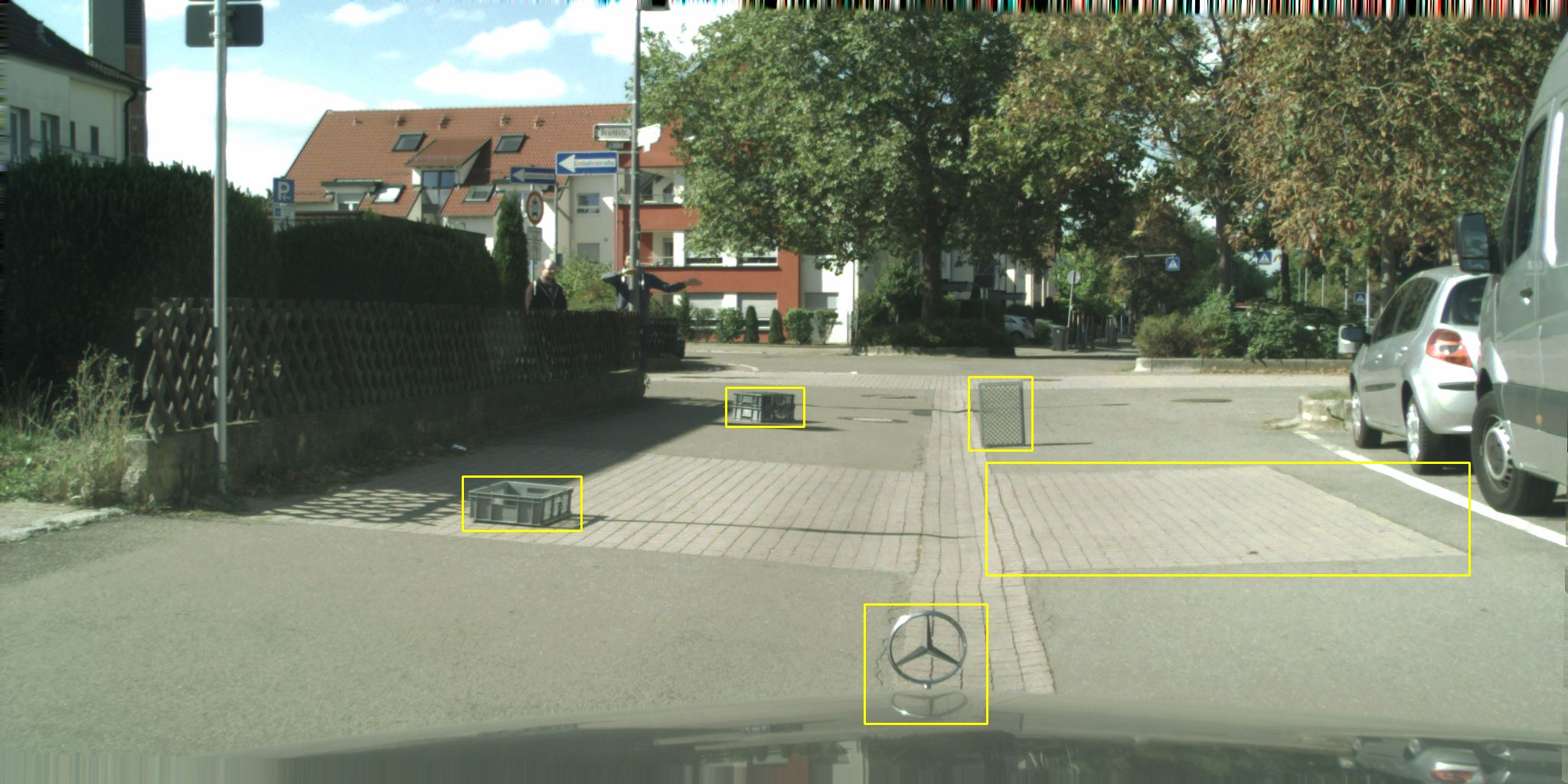}
  \end{subfigure}
  
    \begin{subfigure}{0.25\textwidth}
    \includegraphics[width=\linewidth]{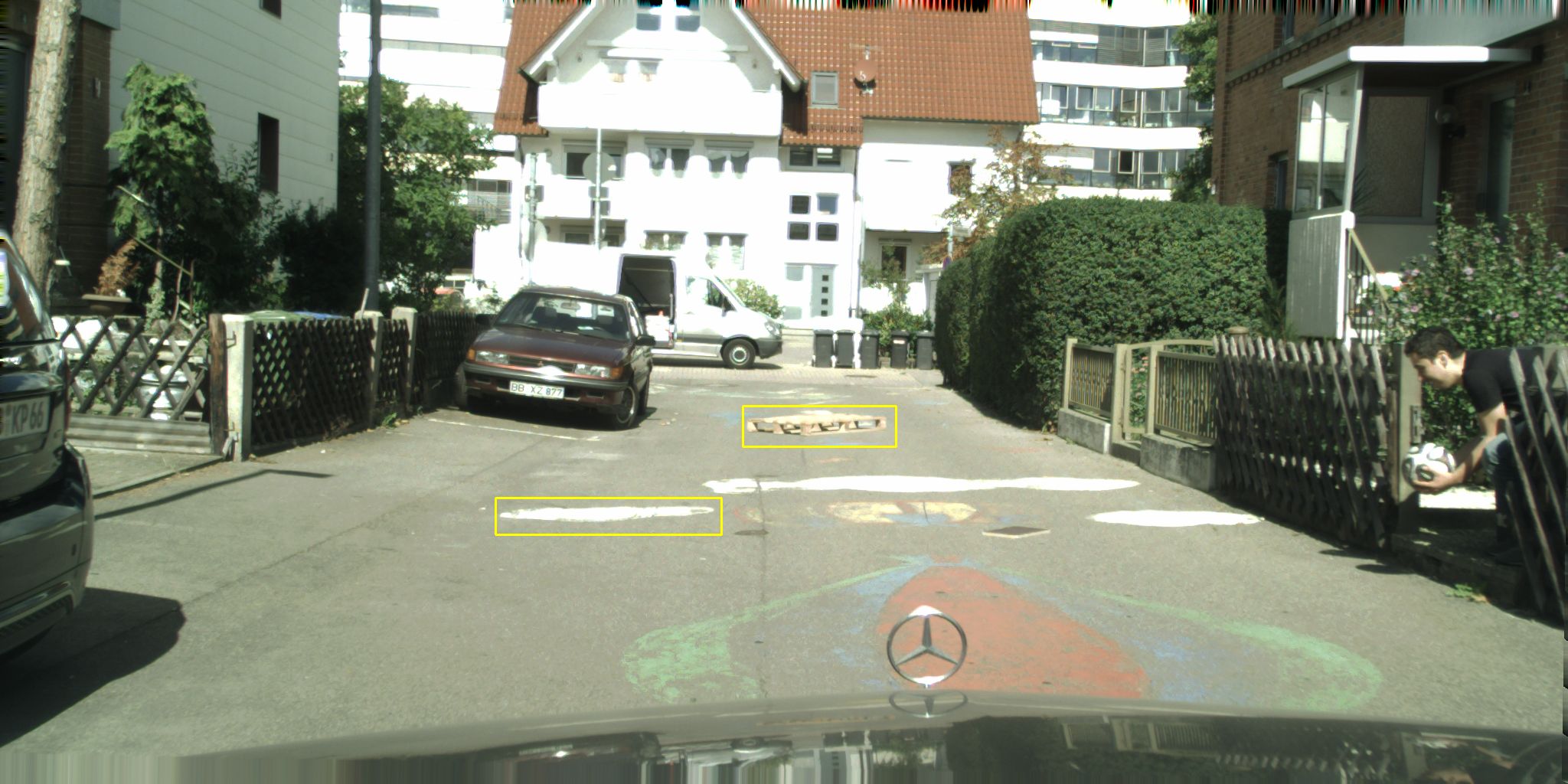}
  \end{subfigure}%
      \begin{subfigure}{0.25\textwidth}
    \includegraphics[width=\linewidth]{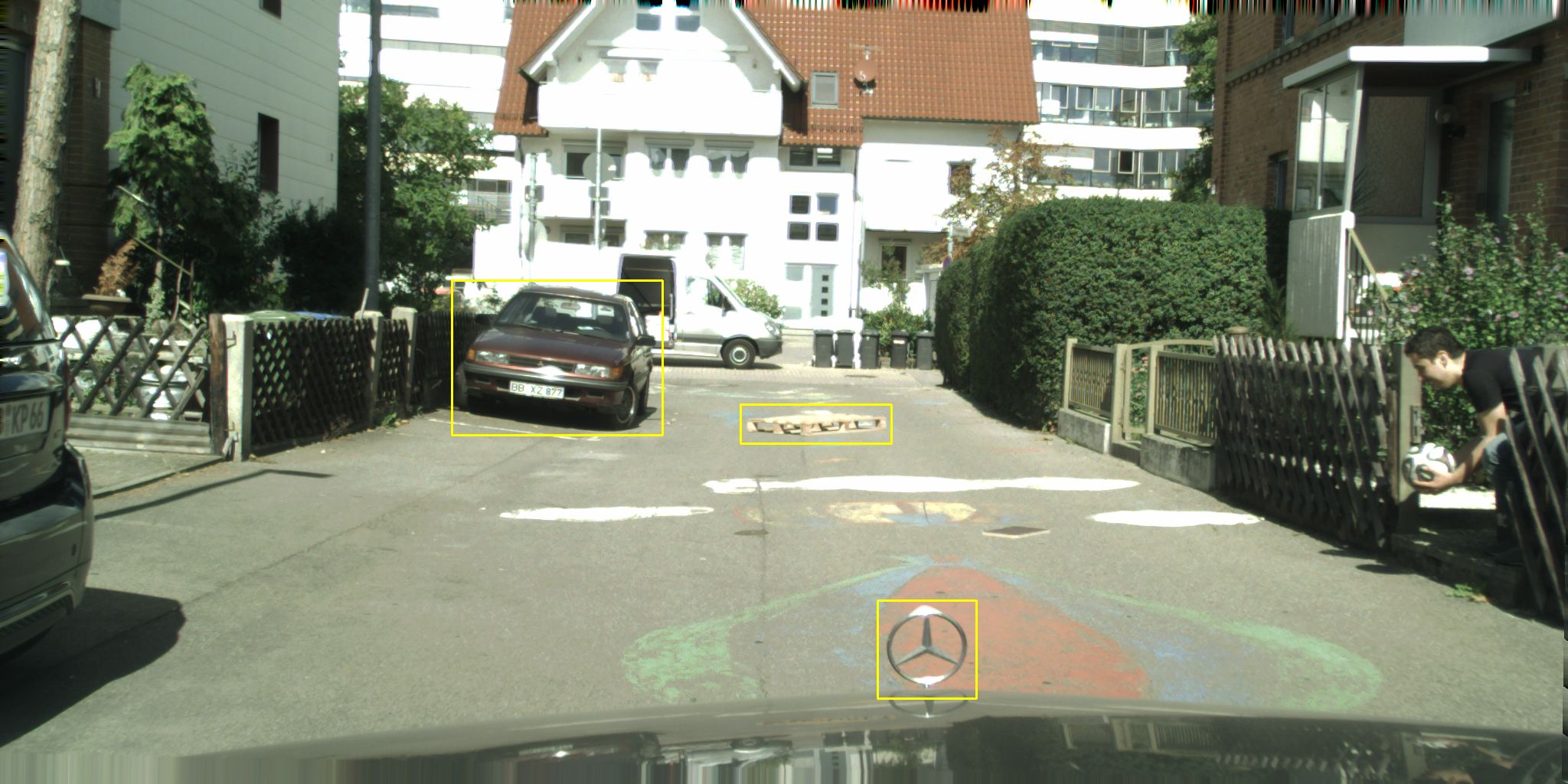}
  \end{subfigure}%
  \begin{subfigure}{0.25\textwidth}
    \includegraphics[width=\linewidth]{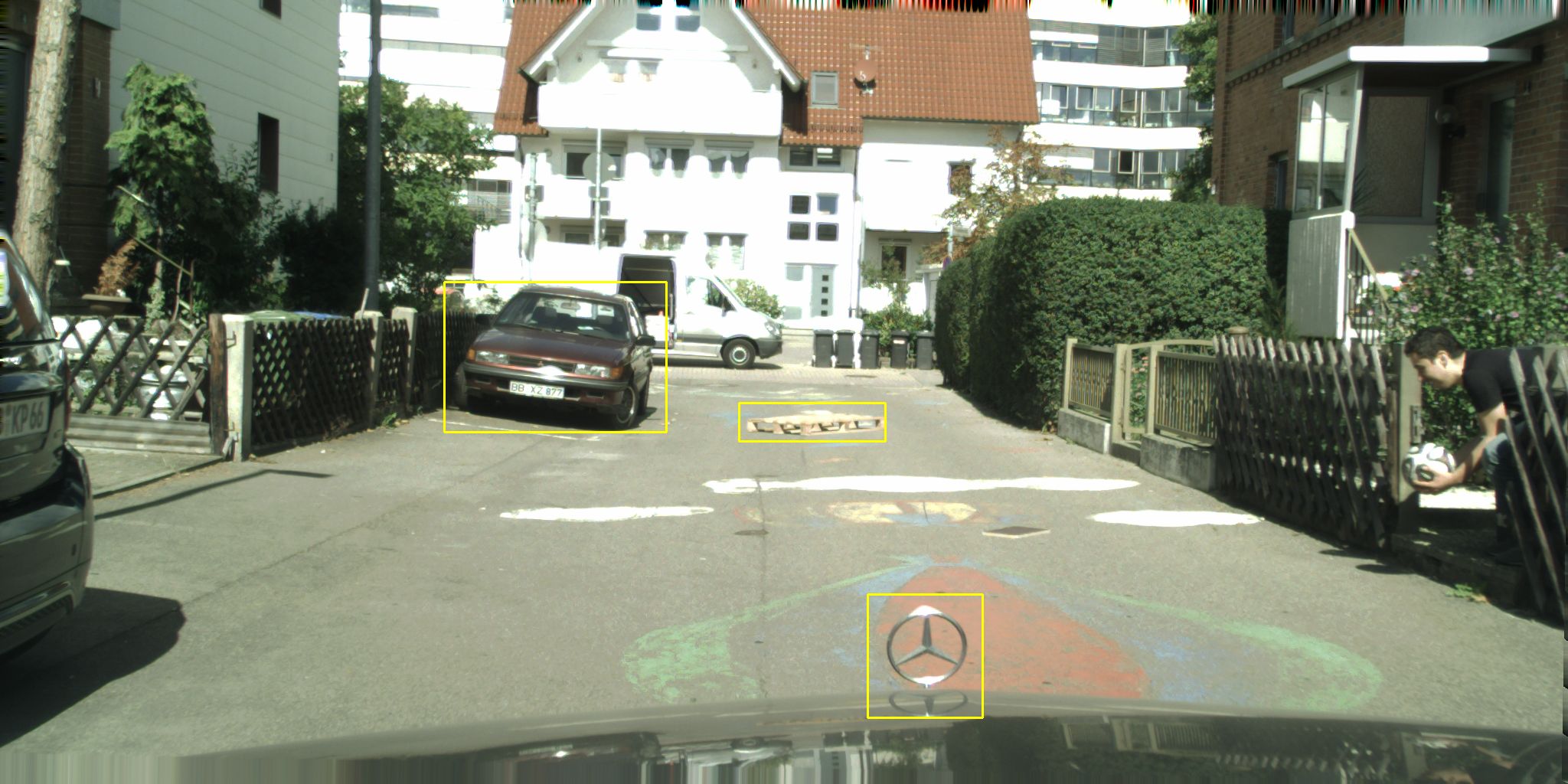}
  \end{subfigure}%
  \begin{subfigure}{0.25\textwidth}
    \includegraphics[width=\linewidth]{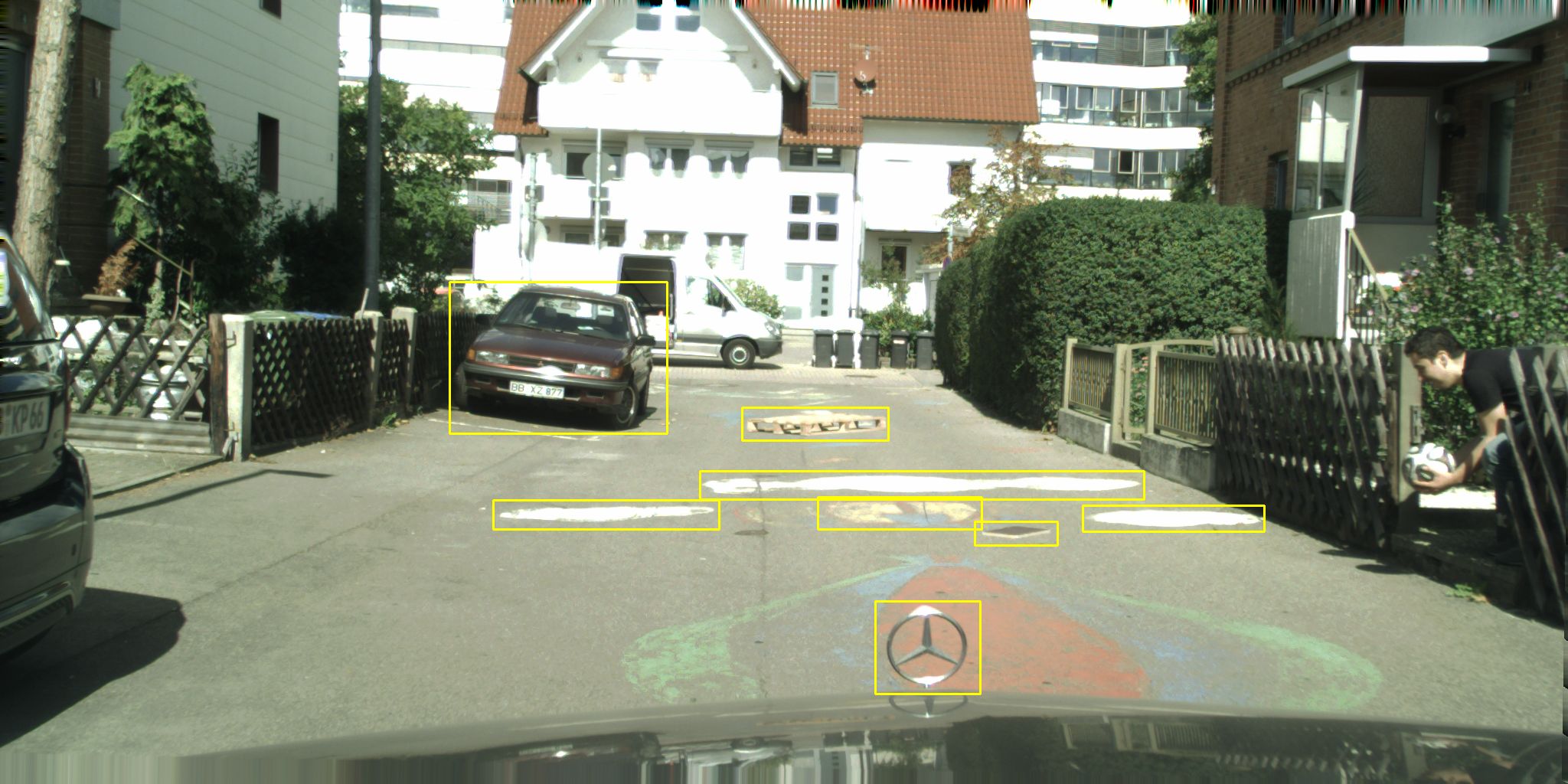}
  \end{subfigure}

      \begin{subfigure}{0.25\textwidth}
    \includegraphics[width=\linewidth]{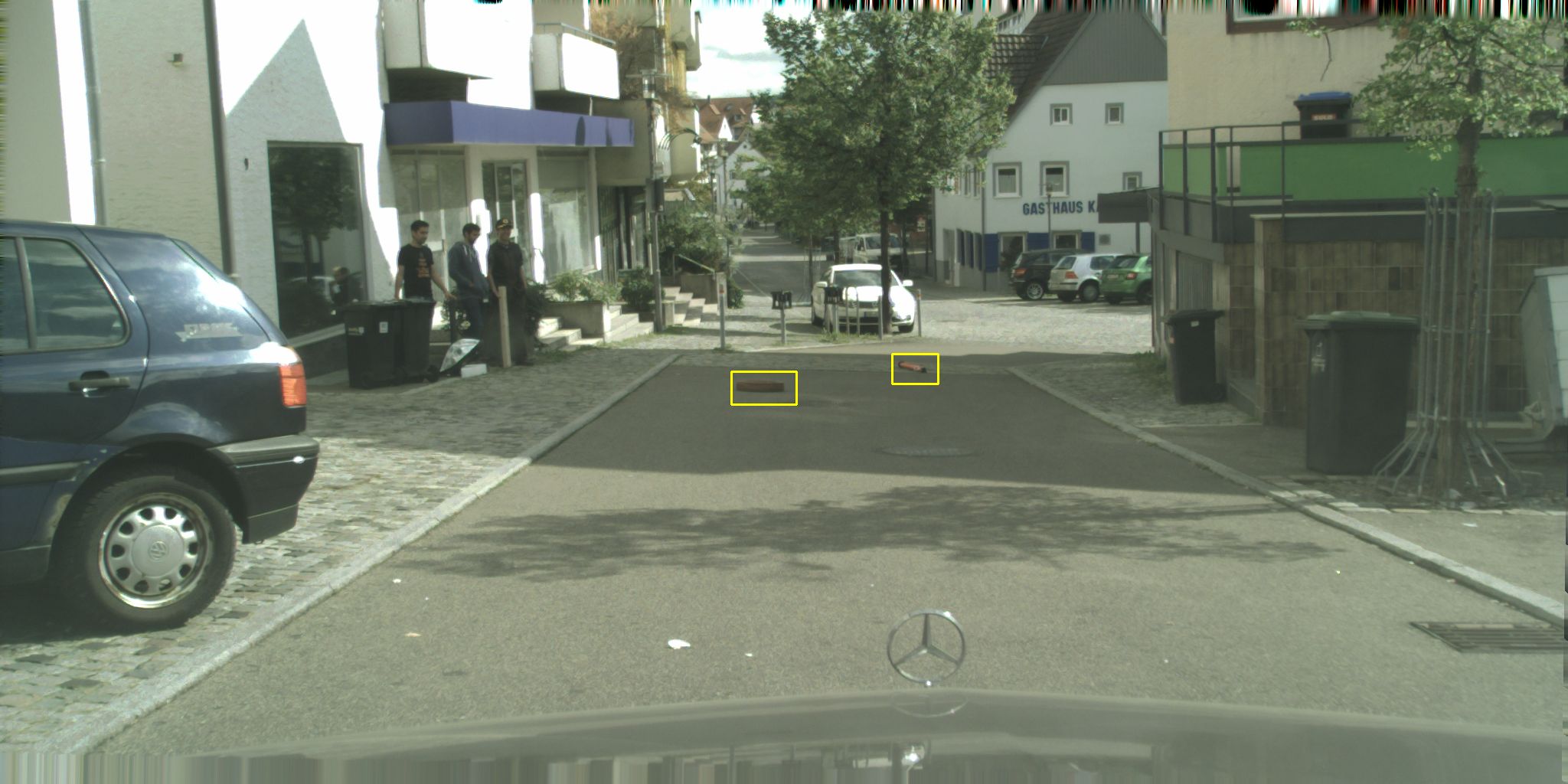}
    \caption{Grounding DINO}
  \end{subfigure}%
      \begin{subfigure}{0.25\textwidth}
    \includegraphics[width=\linewidth]{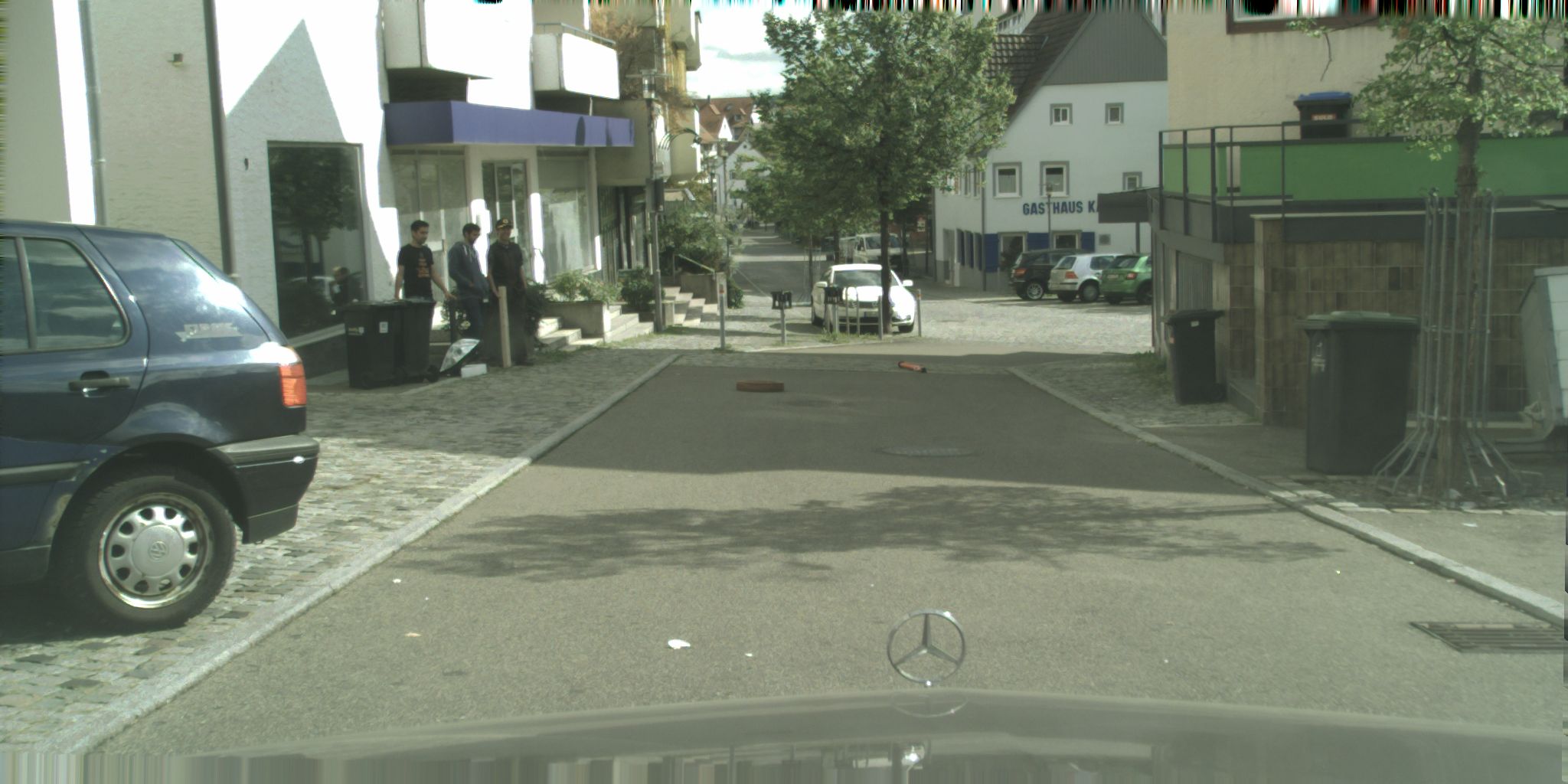}
    \caption{YOLO-World}
  \end{subfigure}%
  \begin{subfigure}{0.25\textwidth}
    \includegraphics[width=\linewidth]{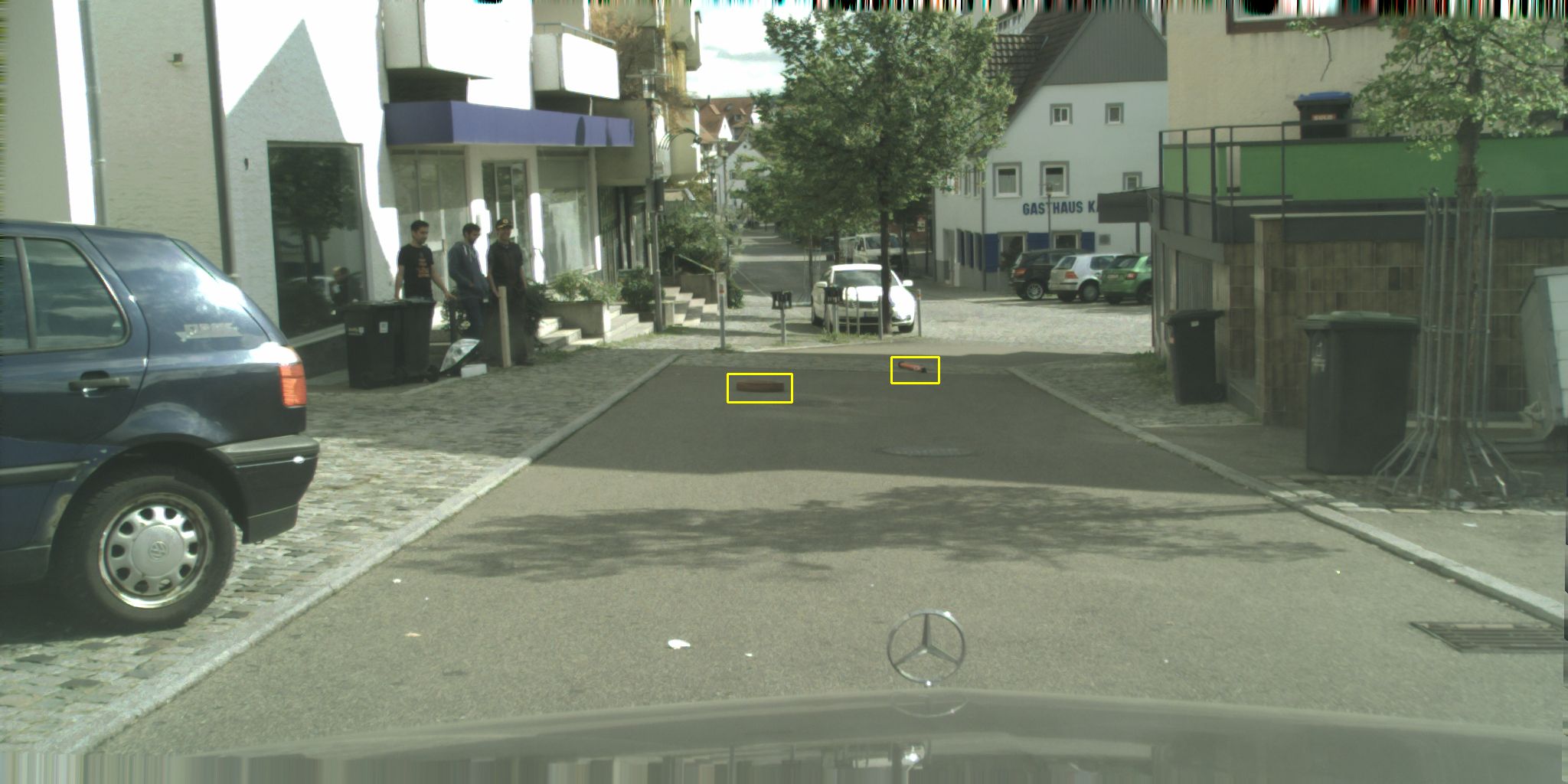}
    \caption{MDETR}
  \end{subfigure}%
  \begin{subfigure}{0.25\textwidth}
    \includegraphics[width=\linewidth]{media/media/05_Schafgasse_1_000007_000110_leftImg8bit.png}
      \caption{OmDet}
  \end{subfigure}
    \caption{Qualitative results on LostAndFound}
    \label{fig:lostandfound}
\end{figure}

\begin{figure*}[!h]
\centering
  \captionsetup{justification=centering,margin=0.5cm}
  \begin{subfigure}{0.25\textwidth}
    \includegraphics[width=\linewidth]{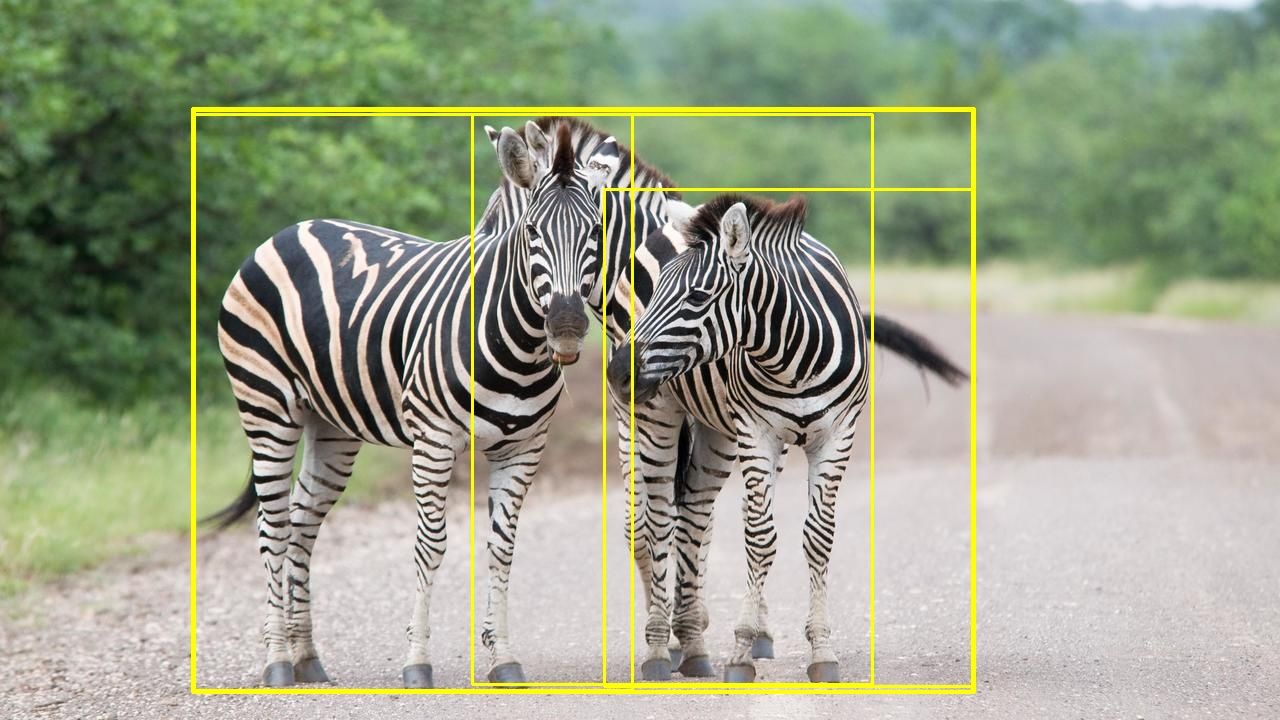}
  \end{subfigure}%
  \begin{subfigure}{0.25\textwidth}
    \includegraphics[width=\linewidth]{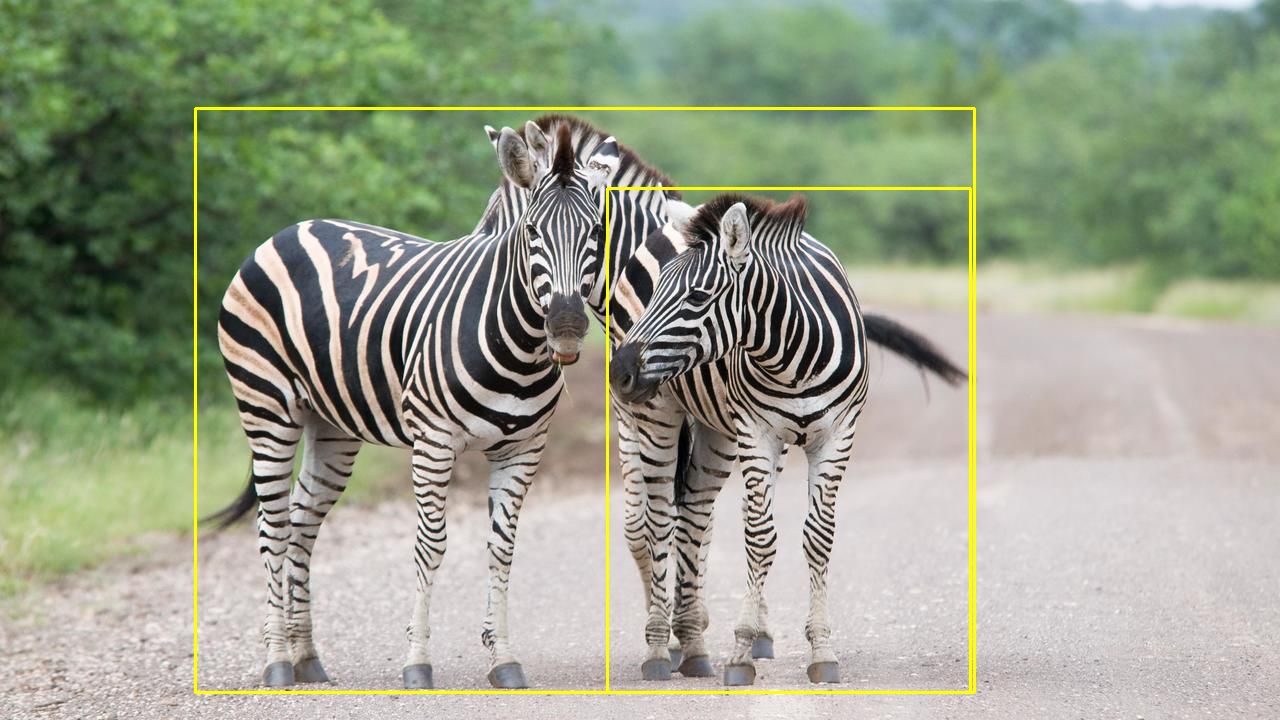}
  \end{subfigure}%
  \begin{subfigure}{0.25\textwidth}
    \includegraphics[width=\linewidth]{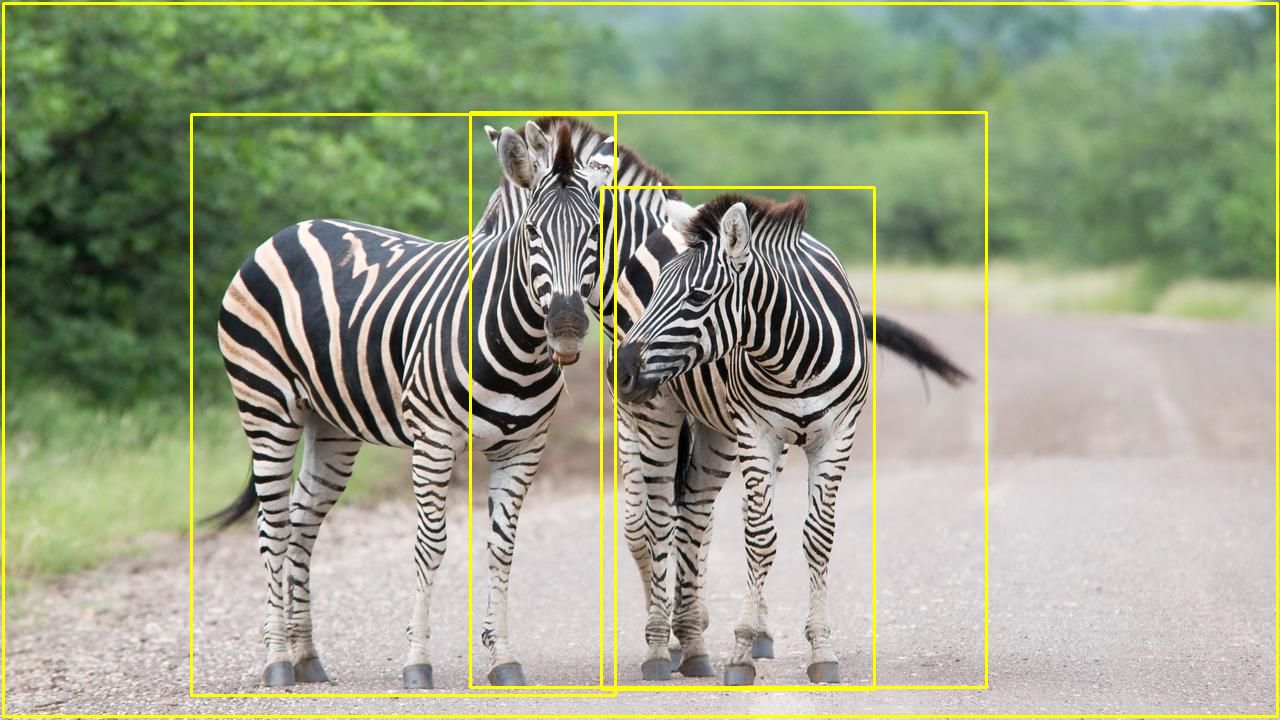}
  \end{subfigure}%
    \begin{subfigure}{0.25\textwidth}
    \includegraphics[width=\linewidth]{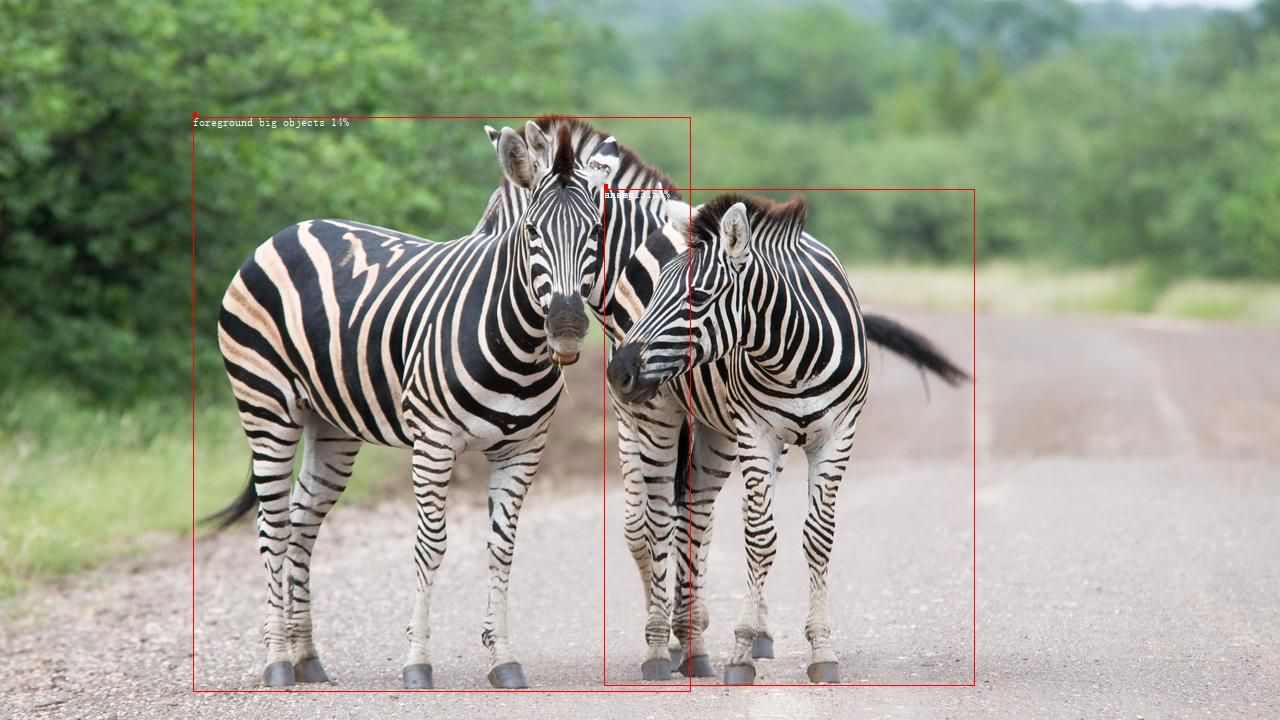}
  \end{subfigure}
  
  \begin{subfigure}{0.25\textwidth}
    \includegraphics[width=\linewidth]{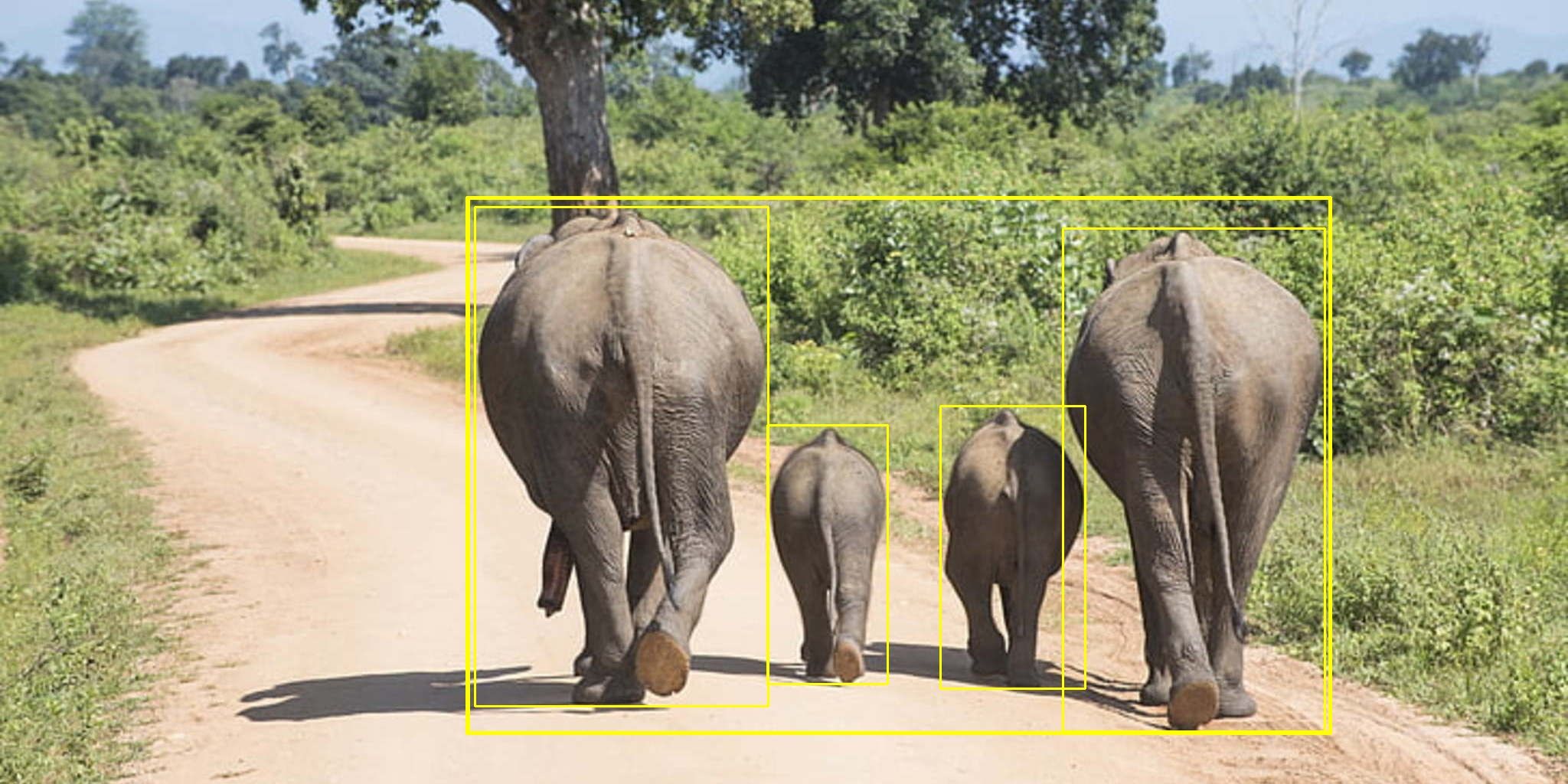}
  \end{subfigure}%
  \begin{subfigure}{0.25\textwidth}
    \includegraphics[width=\linewidth]{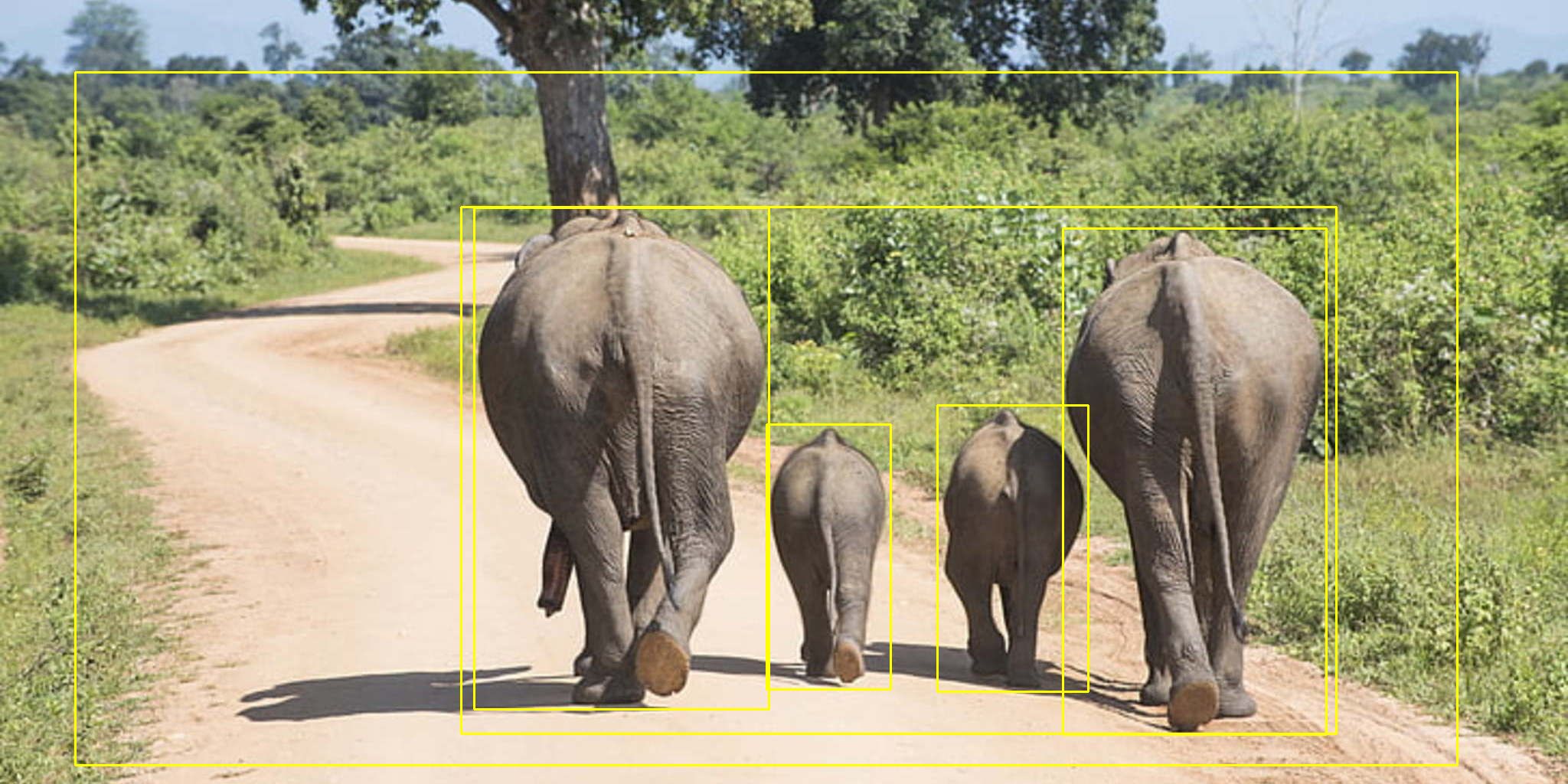}
  \end{subfigure}%
    \begin{subfigure}{0.25\textwidth}
    \includegraphics[width=\linewidth]{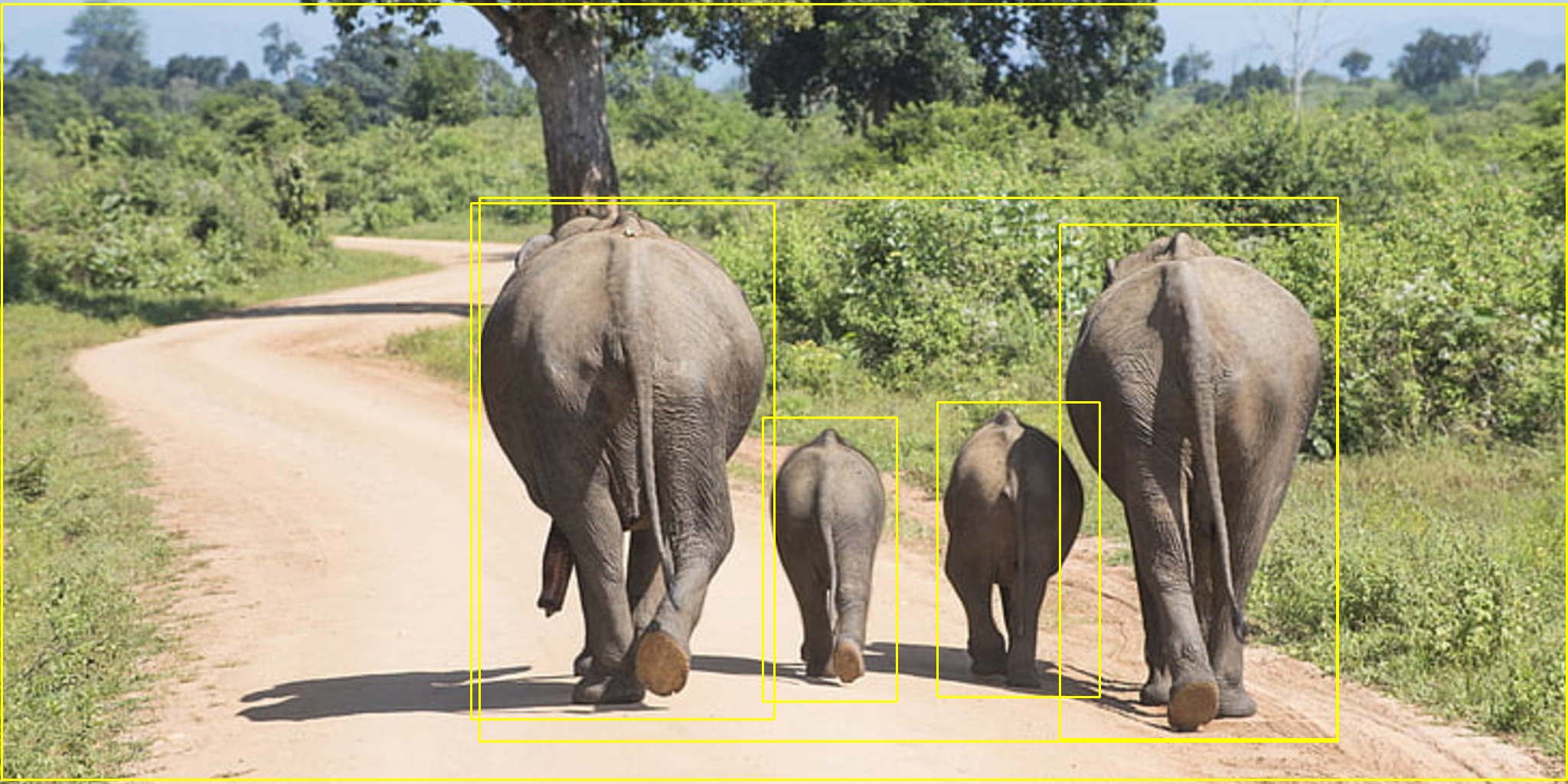}
  \end{subfigure}%
  \begin{subfigure}{0.25\textwidth}
    \includegraphics[width=\linewidth]{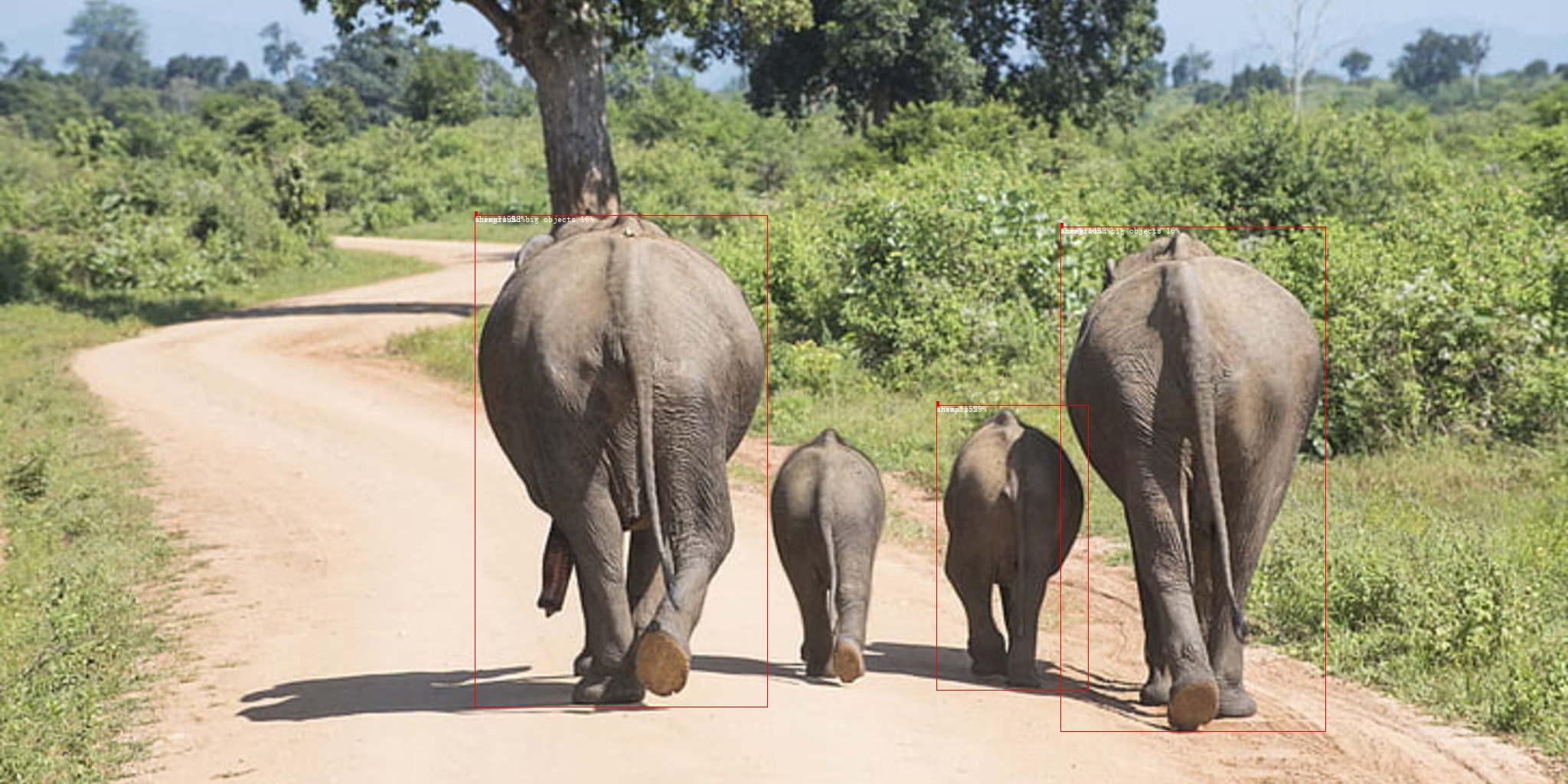}
  \end{subfigure}
  
  \begin{subfigure}{0.25\textwidth}
    \includegraphics[width=\linewidth]{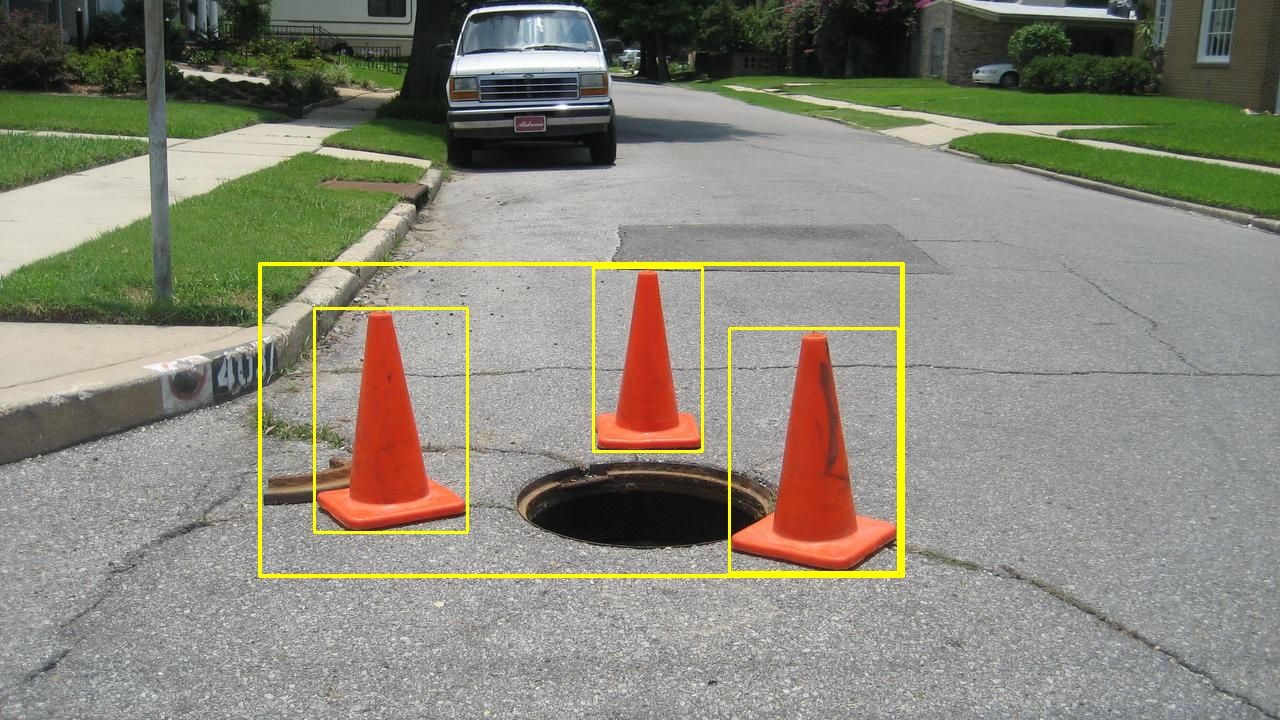}
  \end{subfigure}%
    \begin{subfigure}{0.25\textwidth}
    \includegraphics[width=\linewidth]{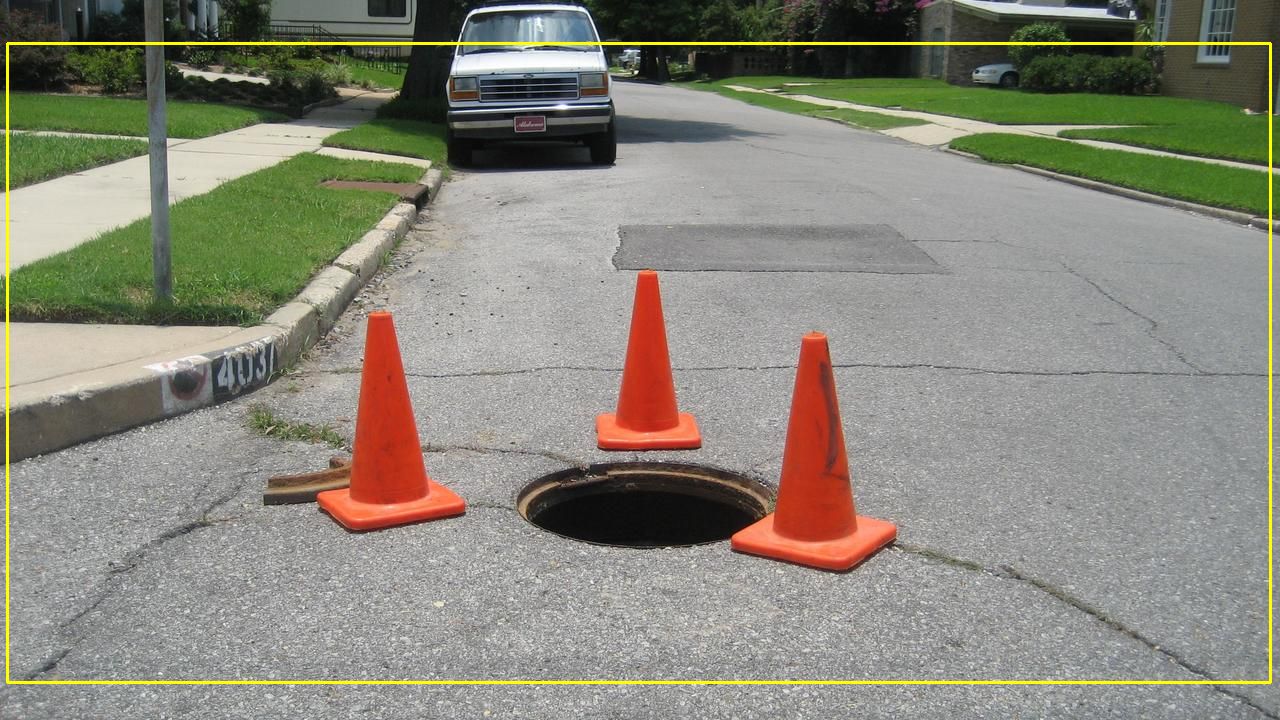}
  \end{subfigure}%
  \begin{subfigure}{0.25\textwidth}
    \includegraphics[width=\linewidth]{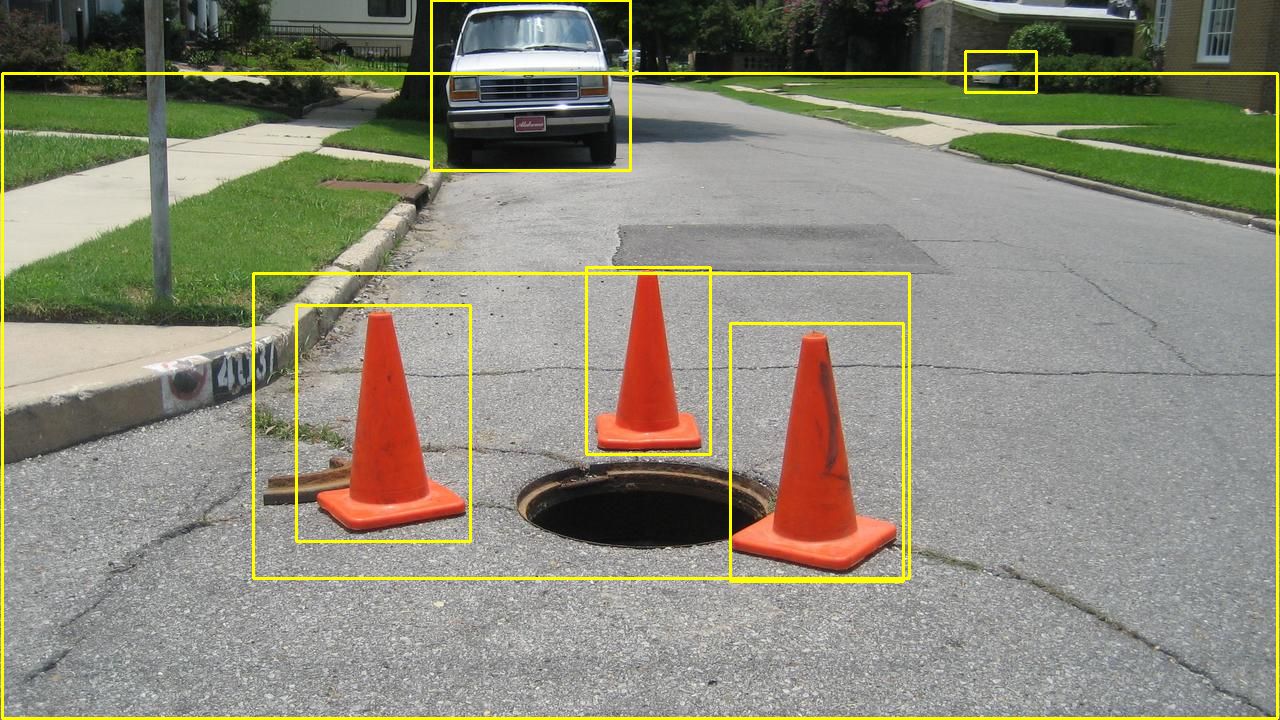}
  \end{subfigure}%
  \begin{subfigure}{0.25\textwidth}
    \includegraphics[width=\linewidth]{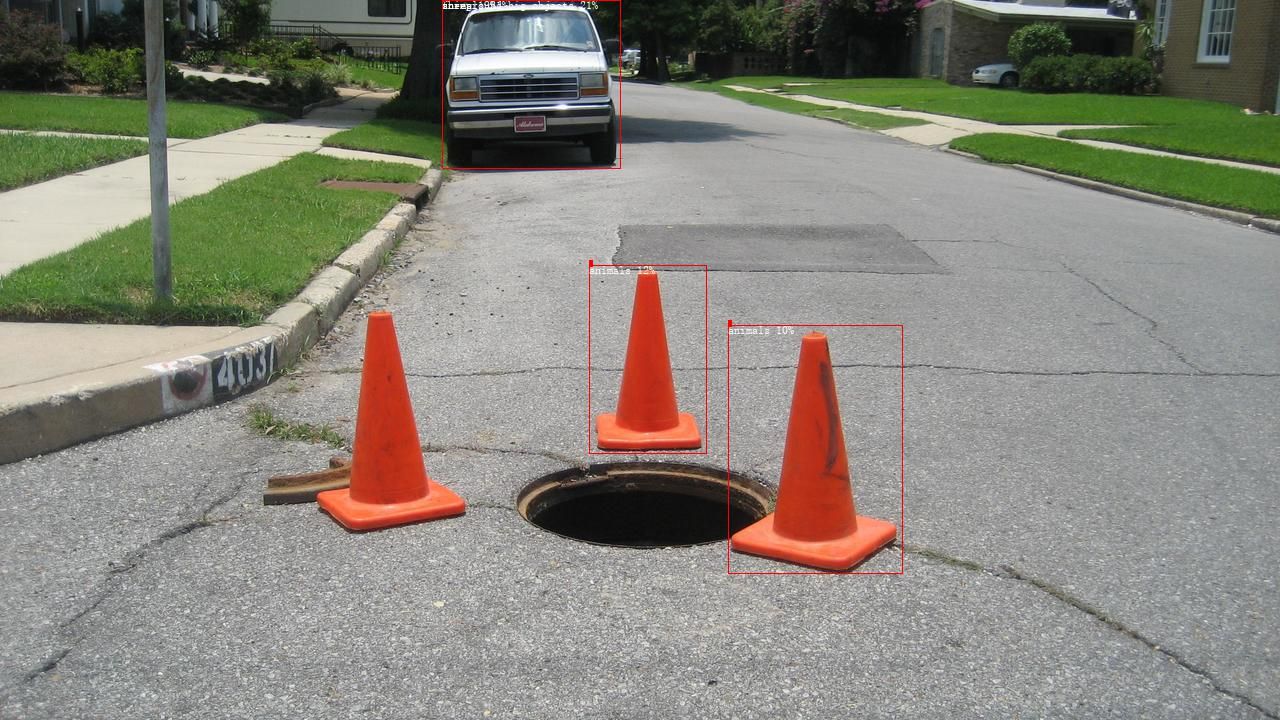}
  \end{subfigure}

    \begin{subfigure}{0.25\textwidth}
    \includegraphics[width=\linewidth]{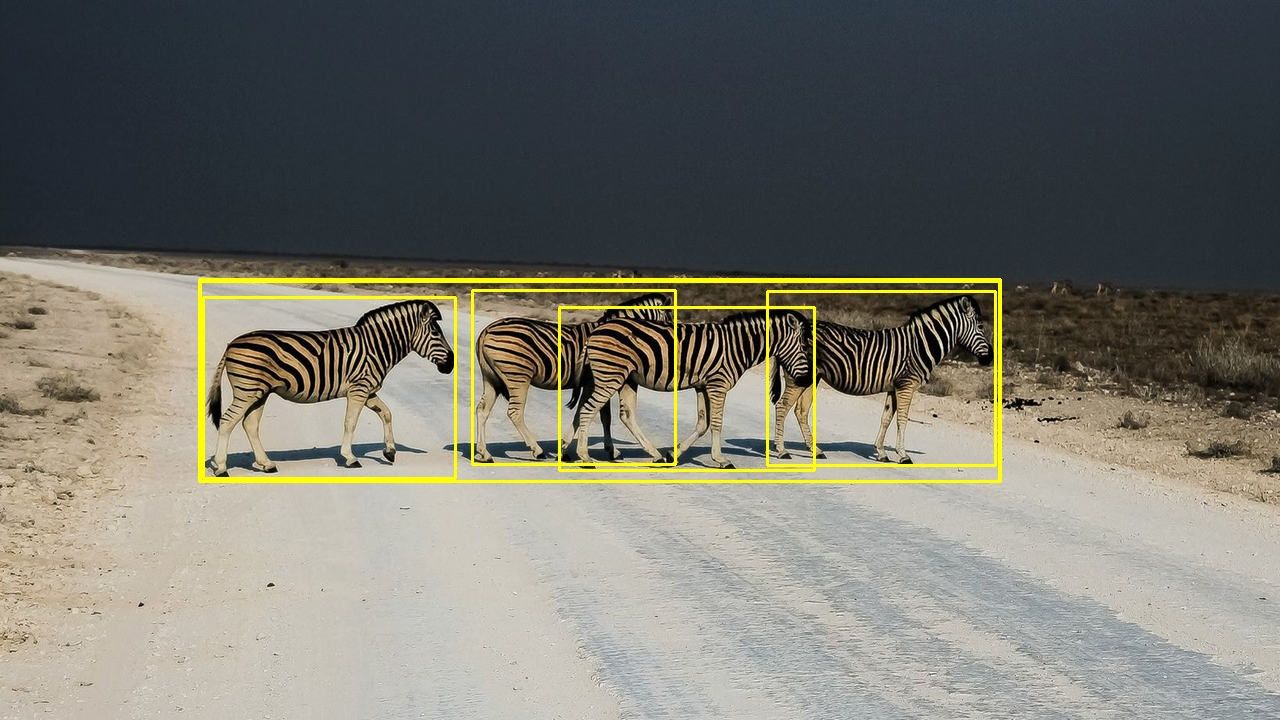}
    \caption{Grounding DINO}
  \end{subfigure}%
    \begin{subfigure}{0.25\textwidth}
    \includegraphics[width=\linewidth]{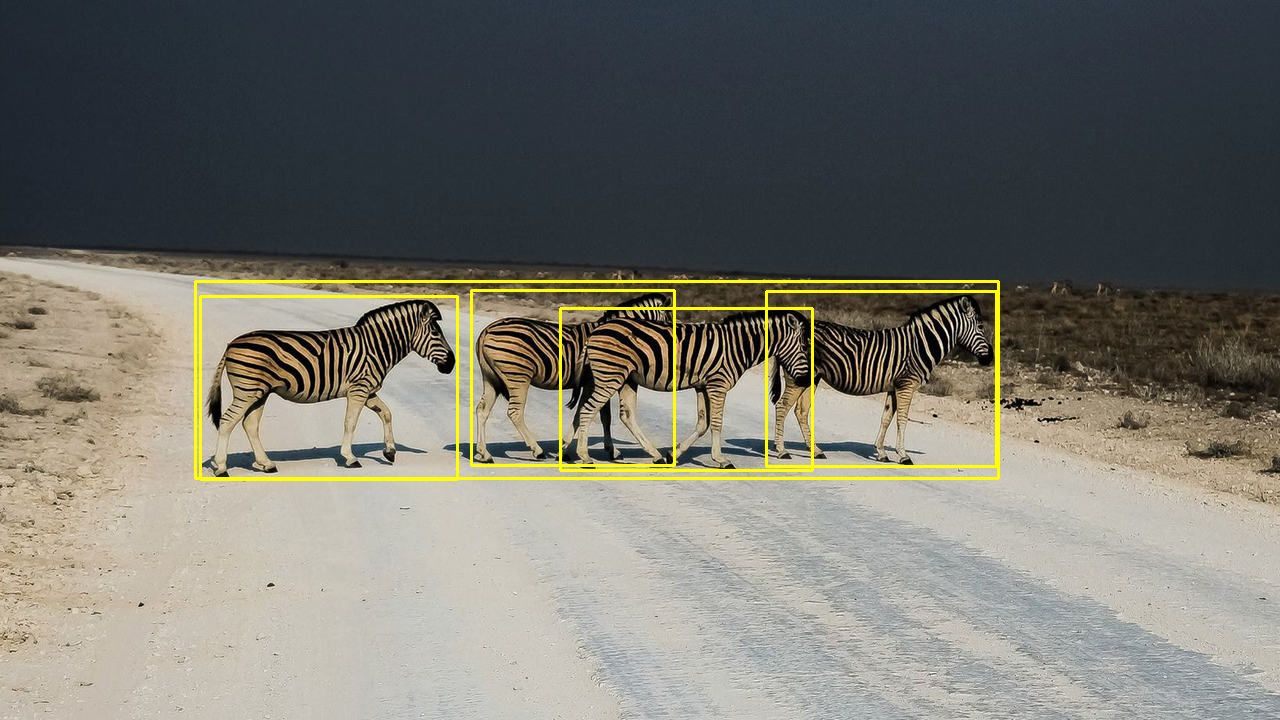}
    \caption{YOLO-World}
  \end{subfigure}%
  \begin{subfigure}{0.25\textwidth}
    \includegraphics[width=\linewidth]{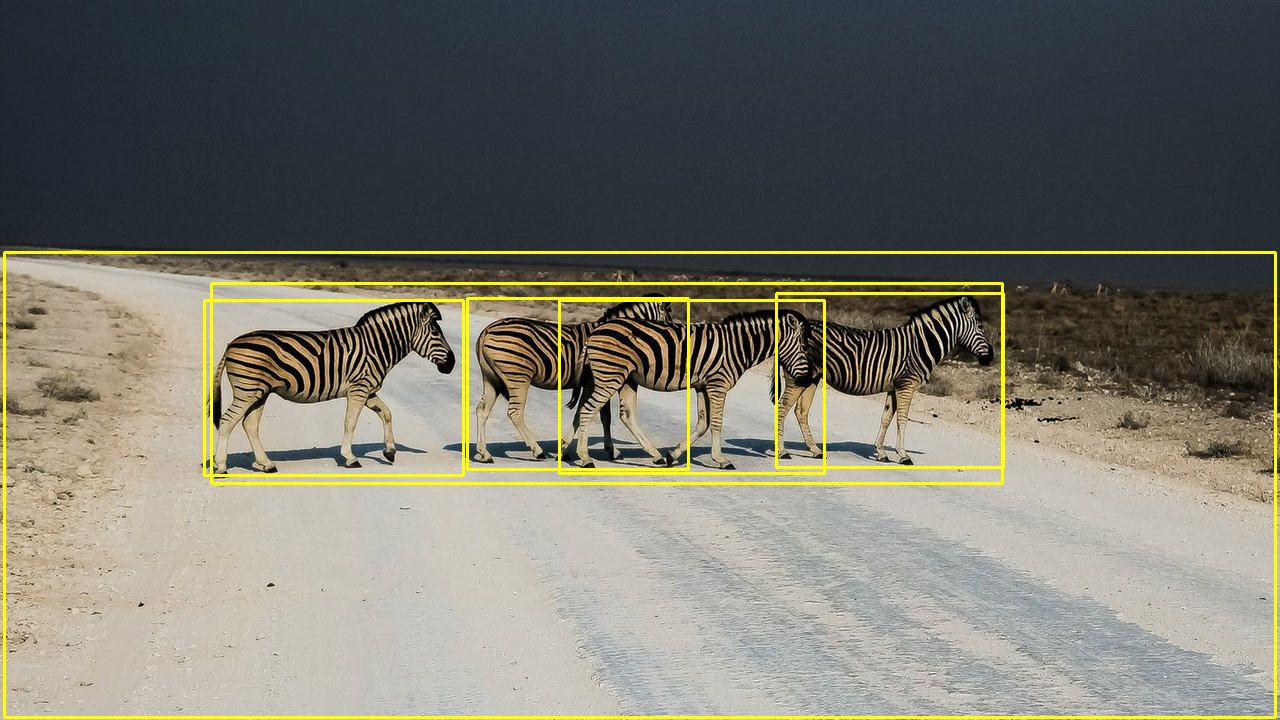}
    \caption{MDETR}
  \end{subfigure}%
  \begin{subfigure}{0.25\textwidth}
    \includegraphics[width=\linewidth]{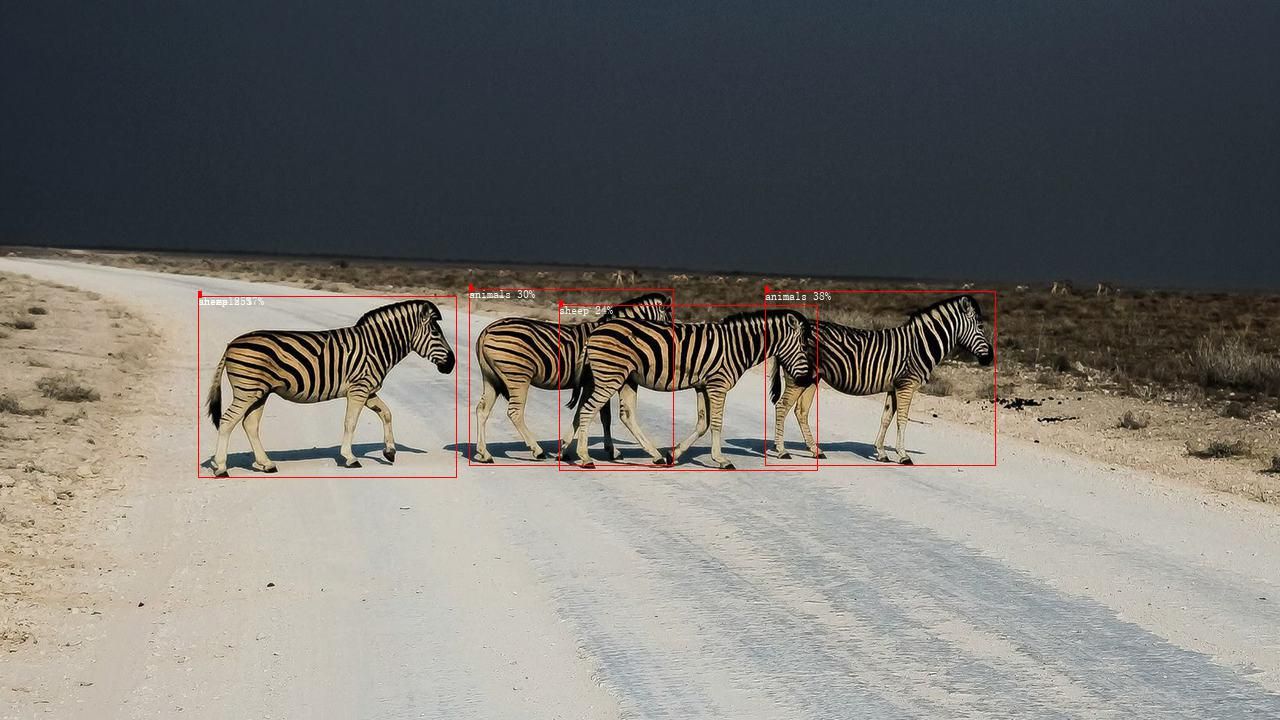}
    \caption{OmDet}
  \end{subfigure}
  \caption{Qualitative results on RoadAnomaly21}
  \label{fig:roadanomaly}
\end{figure*}

In Figure \ref{fig:lostandfound}, Grounding DINO and MDETR successfully detect all OOD objects, while YOLO-World and OmDet slightly lag behind in this task. Grounding DINO and OmDet, however, mistakenly identify road paintings as OOD objects, leading to increased false positives. In Figure \ref{fig:roadanomaly}, OmDet achieves the least false positive rate, while the other models consistently provide multiple predictions for the identical objects. Although OmDet misses a few OOD objects, such as one of the cones and an elephant, it does not merge multiple objects into a single detection.

\subsection{Additional visualization results for Grounding DINO using Test Time Augmentation}

\begin{figure*}[!h]
\centering
  \captionsetup{justification=centering,margin=0.5cm}
  \begin{subfigure}{0.4\textwidth}
    \includegraphics[width=\linewidth]{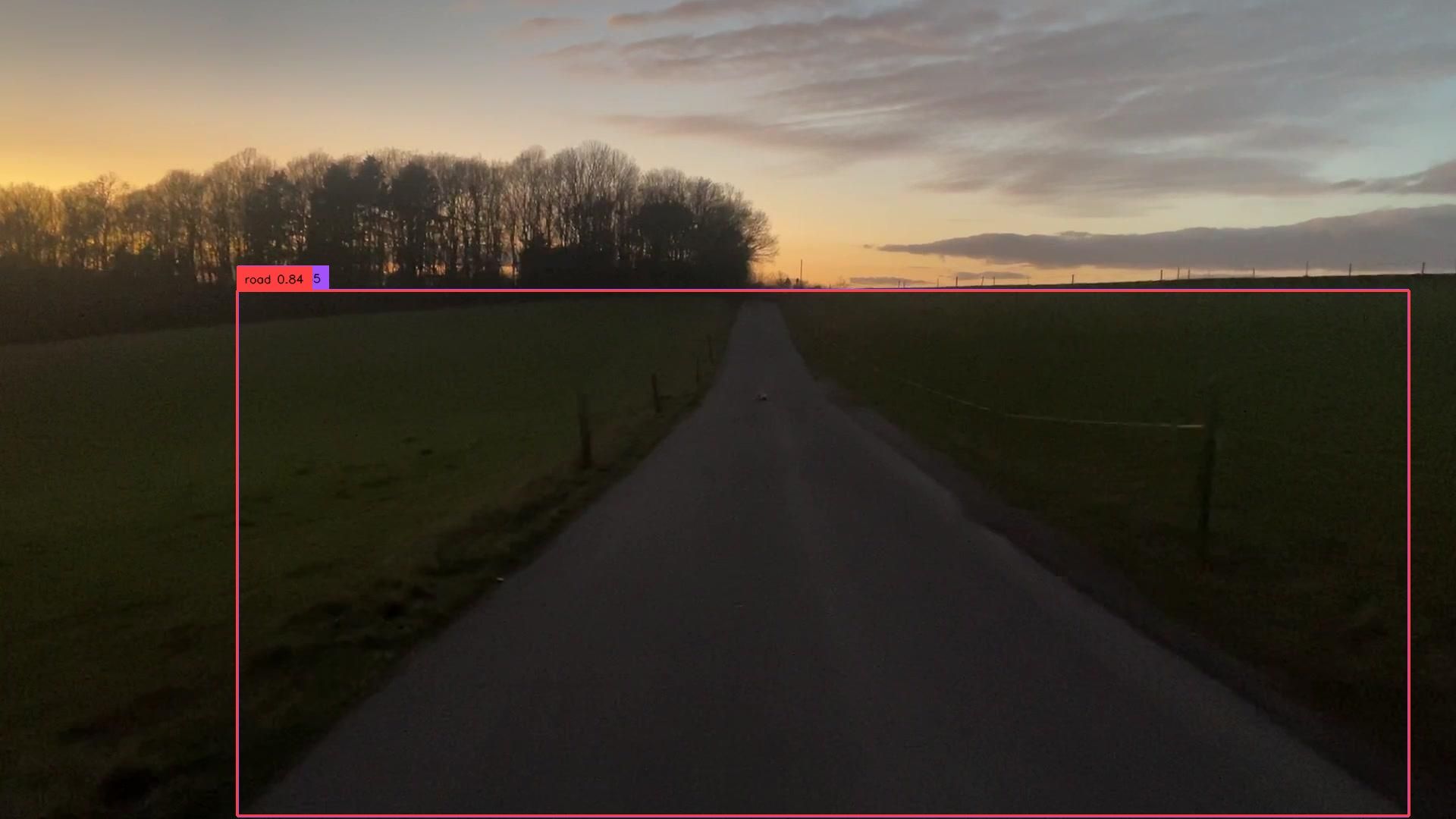}
  \end{subfigure}
  \begin{subfigure}{0.4\textwidth}
    \includegraphics[width=\linewidth]{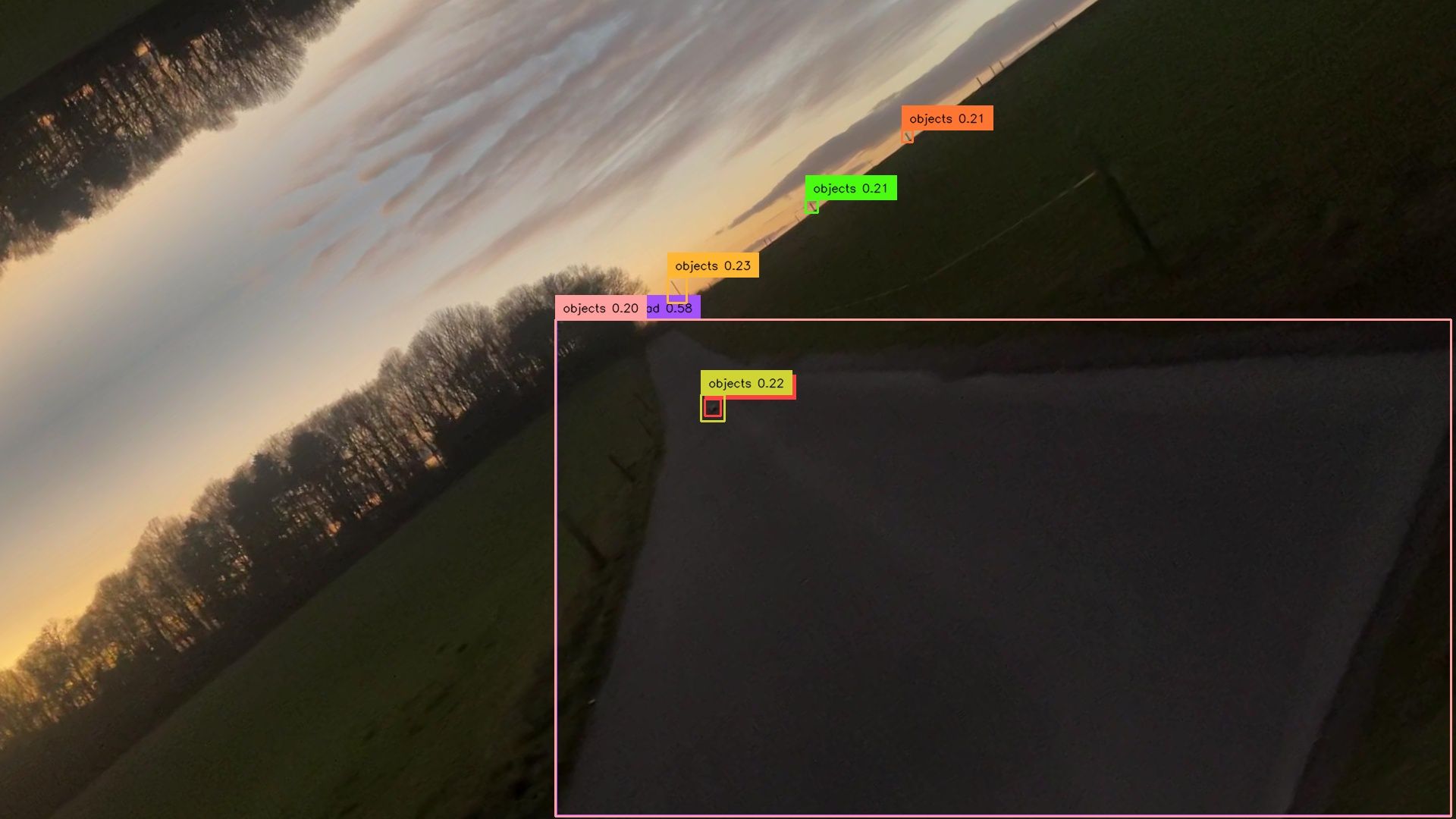}
  \end{subfigure}

  \begin{subfigure}{0.4\textwidth}
    \includegraphics[width=\linewidth]{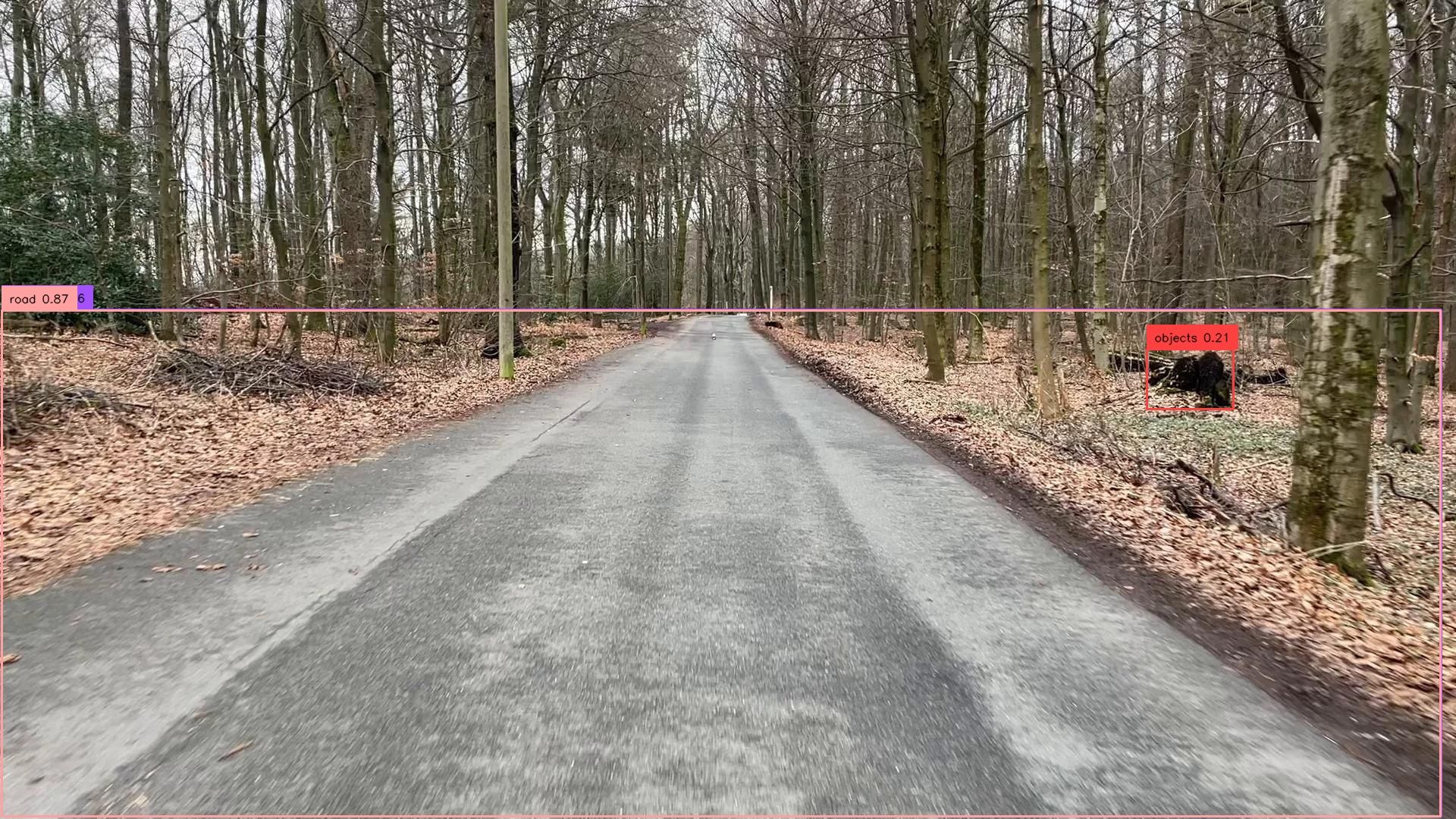}
  \end{subfigure}
  \begin{subfigure}{0.4\textwidth}
    \includegraphics[width=\linewidth]{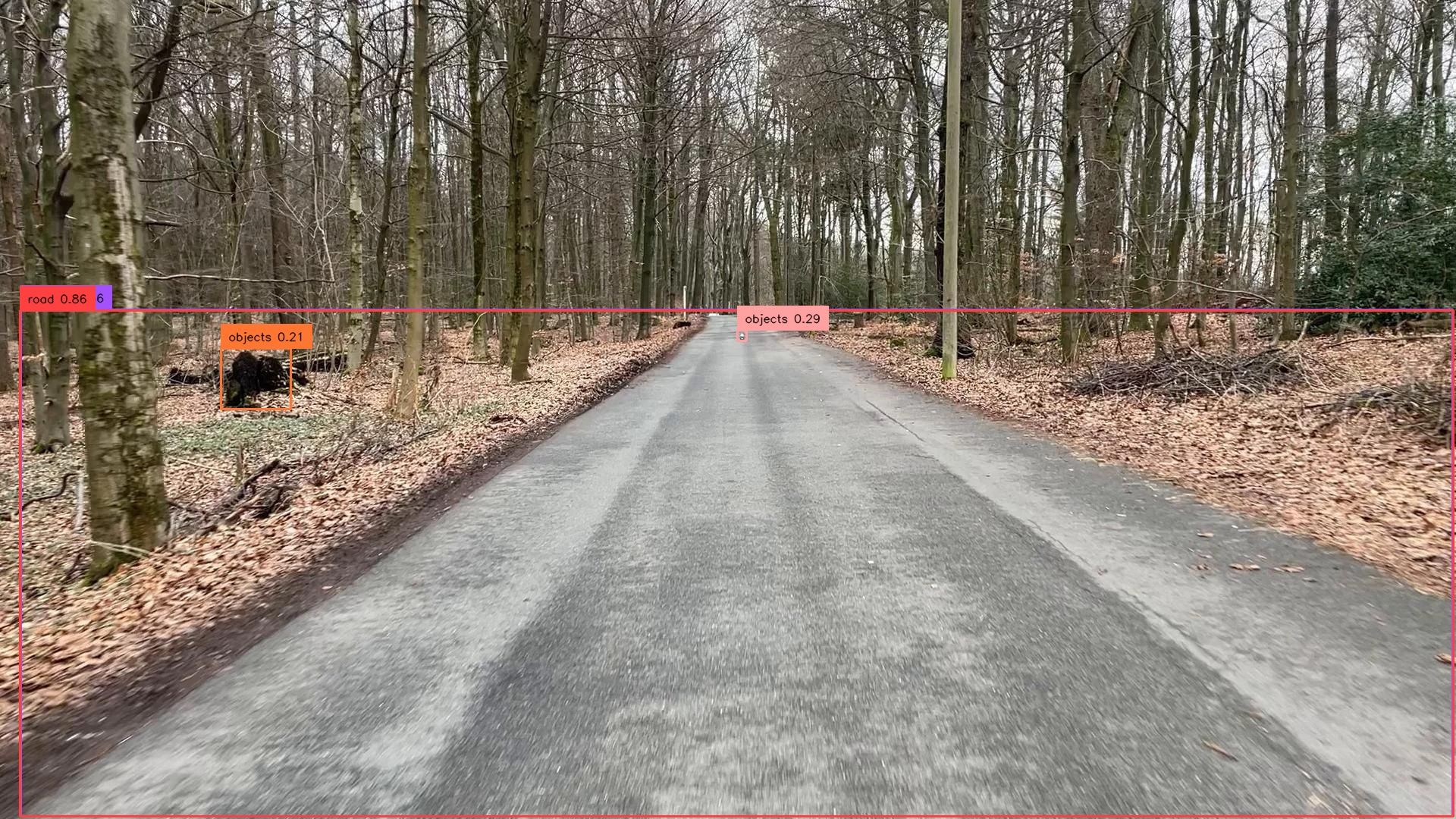}
  \end{subfigure}
  
  \begin{subfigure}{0.4\textwidth}
    \includegraphics[width=\linewidth]{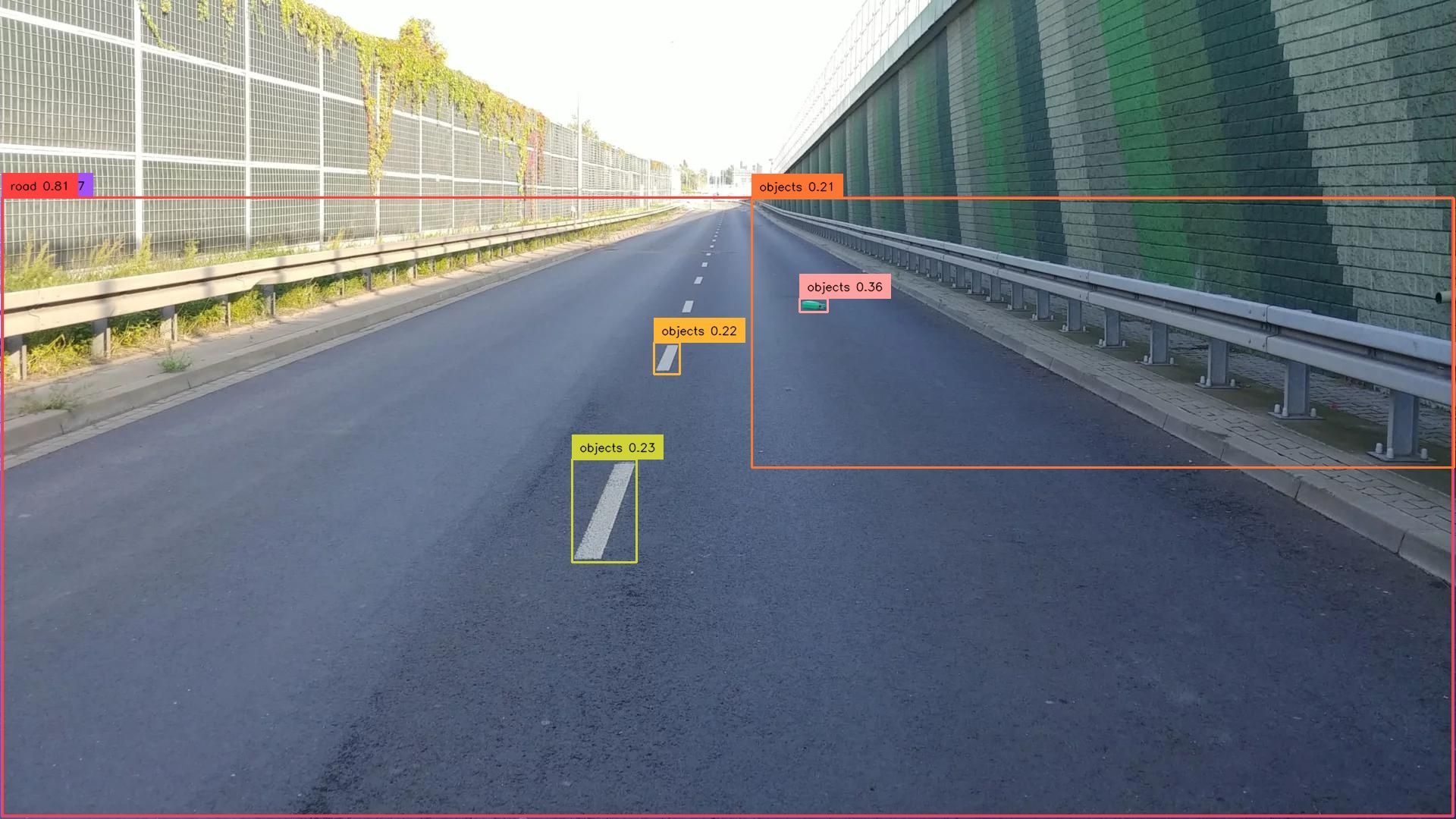}
  \end{subfigure}
  \begin{subfigure}{0.4\textwidth}
    \includegraphics[width=\linewidth]{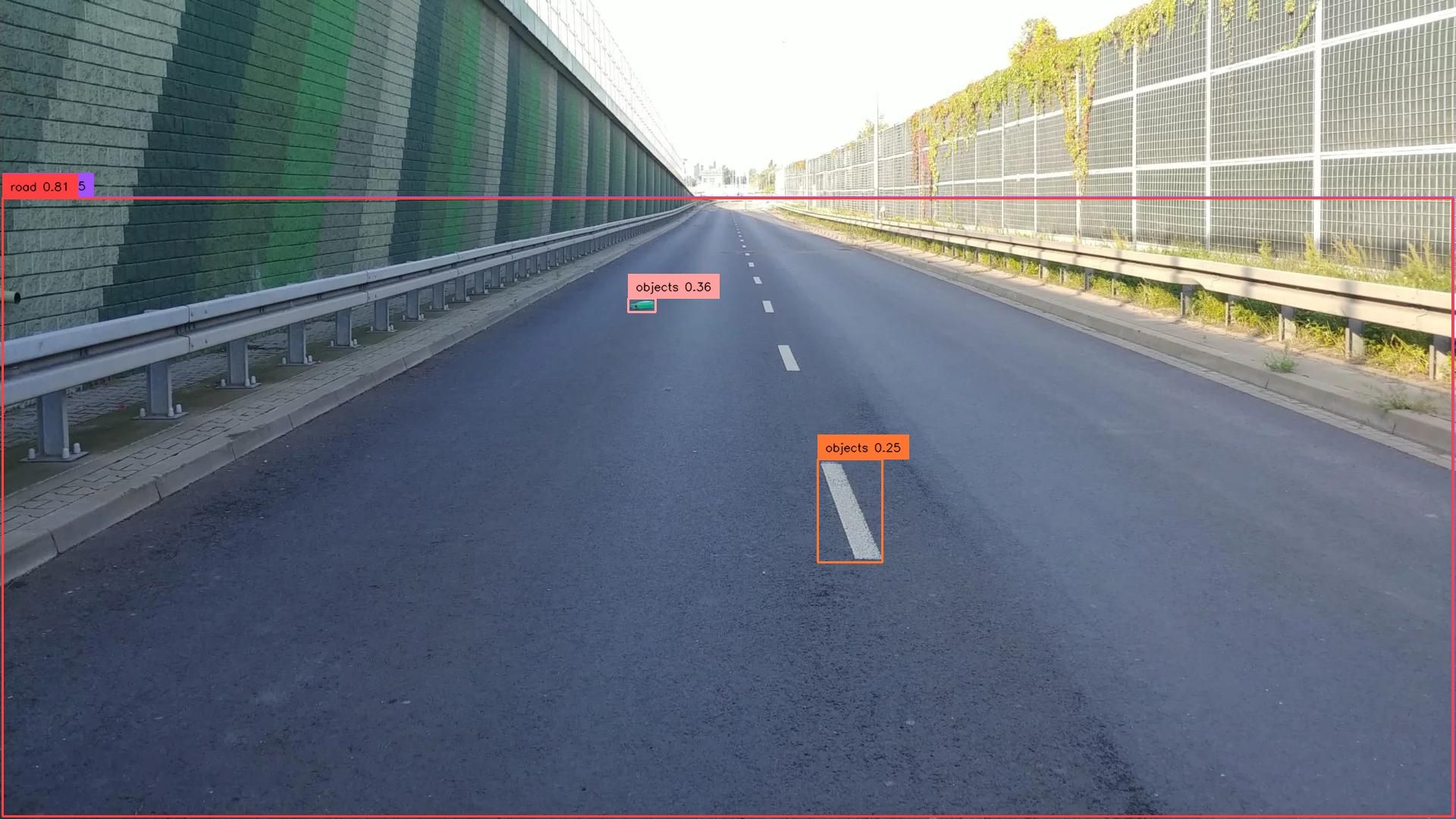}
  \end{subfigure}

  \begin{subfigure}{0.4\textwidth}
    \includegraphics[width=\linewidth]{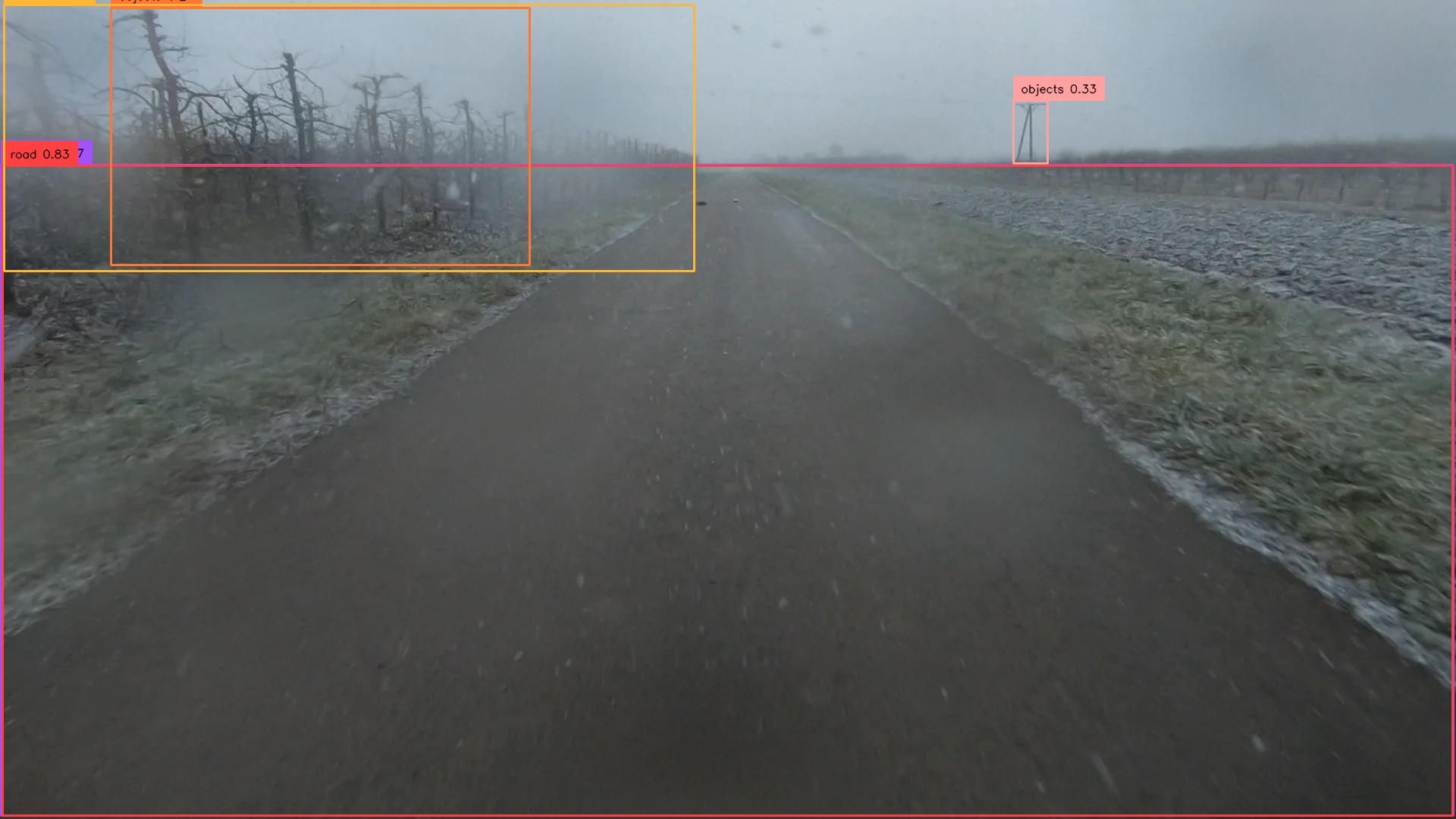}
    \caption{Original}
  \end{subfigure}
  \begin{subfigure}{0.4\textwidth}
    \includegraphics[width=\linewidth]{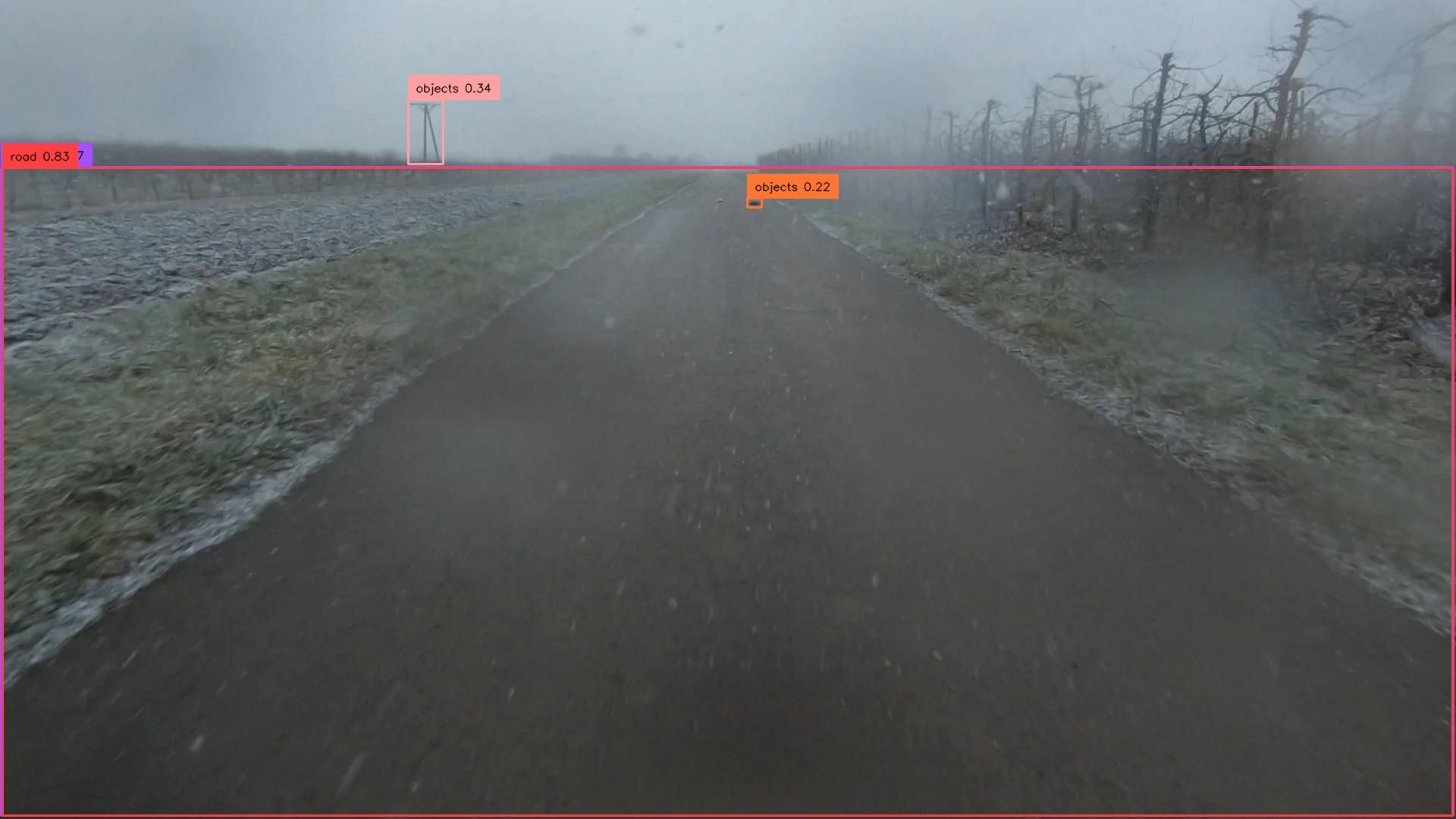}
    \caption{Augmented}
  \end{subfigure}
  
    \caption{Qualitative results on RoadObstacle21 using Test time augmentation}
    \label{fig:tta}
\end{figure*}

In this section, we compare the results of Grounding DINO when applied to original images versus augmented images, shown in Figure \ref{fig:tta}. The model is able to provide additional detections when augmenting images. Specifically, we obtain additional correct detections in many images with horizontal flips. In some cases, rotations also help the model to detect OOD objects.

\end{document}